\documentclass[10pt,twocolumn,letterpaper]{article}

\usepackage{3dv}
\usepackage{times}
\usepackage{epsfig}
\usepackage{graphicx}
\usepackage{amsmath}
\usepackage{amssymb}

\usepackage{booktabs}
\usepackage{multirow}
\usepackage{pifont}
\usepackage{subcaption}
\usepackage{graphicx}
\usepackage{tabularx}
\usepackage{gensymb}
\usepackage{rotating}

\usepackage[pagebackref=true,breaklinks=true,colorlinks,bookmarks=false]{hyperref}

\threedvfinalcopy 

\def\threedvPaperID{123} 

\ifthreedvfinal\fi

\usepackage{xstring,xifthen}
\newcommand{\subsubsec}[1]{\vspace{0.2em}\noindent\textbf{\IfEndWith{#1}{.}{#1}{#1.}}}

\newcommand{\x}{\ding{54}}
\newcommand{\V}{\ding{52}}

\begin{document}

\title{Multi-Category Mesh Reconstruction From Image Collections}

\author{Alessandro Simoni$^*$, Stefano Pini$^*$, Roberto Vezzani, Rita Cucchiara \\
Department of Engineering ``Enzo Ferrari'', University of Modena and Reggio Emilia (Italy) \\
{\tt\small \{alessandro.simoni, s.pini, roberto.vezzani, rita.cucchiara\}@unimore.it}
}

\maketitle
\thispagestyle{empty}

\begin{abstract}
Recently, learning frameworks have shown the capability of inferring the accurate shape, pose, and texture of an object from a single RGB image.
However, current methods are trained on image collections of a single category in order to exploit specific priors, and they often make use of category-specific 3D templates.
In this paper, we present an alternative approach that infers the textured mesh of objects combining a series of deformable 3D models and a set of instance-specific deformation, pose, and texture.
Differently from previous works, our method is trained with images of multiple object categories using only foreground masks and rough camera poses as supervision.
Without specific 3D templates, the framework learns category-level models which are deformed to recover the 3D shape of the depicted object.
The instance-specific deformations are predicted independently for each vertex of the learned 3D mesh, enabling the dynamic subdivision of the mesh during the training process. 
Experiments show that the proposed framework can distinguish between different object categories and learn category-specific shape priors in an unsupervised manner.
Predicted shapes are smooth and can leverage from multiple steps of subdivision during the training process, obtaining comparable or state-of-the-art results on two public datasets.
Models and code are publicly released~\footnote{~\url{https://github.com/aimagelab/mcmr}}.
\end{abstract}


\renewcommand{\thefootnote}{\fnsymbol{footnote}}
\footnotetext{~Equal contribution.}
\renewcommand{\thefootnote}{\fnsymbol{arabic}}

\section{Introduction}
In recent years, the inference of 3D object shapes from 2D images has shown astonishing progress in the computer vision community.
By addressing the task as an inverse graphics problem, \ie considering the 2D image as the rendering of a 3D model, several methods~\cite{kanazawa2018learning,goel2020shape,tulsiani2020implicit} have shown that deep models are capable of restoring the shape, pose, and texture of the portrayed object.
While previous methods rely on direct 3D supervision~\cite{choy20163d,girdhar2016learning,wang2018pixel2mesh,xie2019pix2vox} or multiple views~\cite{tatarchenko2016multi,gwak2017weakly,tulsiani2017multi,lin2019photometric}, recent approaches only require segmentation masks, object keypoints, and coarse camera poses~\cite{kanazawa2018learning,goel2020shape,tulsiani2020implicit}. In the last couple of years, some methods have lessened the dependency on keypoints~\cite{tulsiani2020implicit} and even on the camera viewpoint~\cite{goel2020shape}.
All these methods share the same underlying approach: a deep model learns a mean 3D shape, called \textit{meanshape}, for the object category during training; then, instance-specific deformation, texture and camera pose are predicted and applied to the learned meanshape to regress the 3D model of the object.

A major limitation of existing methods is that they are category-specific: they must be trained and evaluated on image collections of a single object category.
This choice has been motivated by the need of category-specific priors in order to recover the 3D shape from 2D images, which is indeed an ill-posed problem
unless additional constraints are taken into account.
Moreover, most of the approaches~\cite{kanazawa2018learning,goel2020shape,tulsiani2020implicit} initialize the learnable meanshape with a category-specific representative 3D model.
To the best of our knowledge, there have been no attempts to extend these methods to scenarios where image collections of multiple categories are available both in training and at inference time.

\begin{figure}
    \begin{center}
        \includegraphics[width=0.95\linewidth]{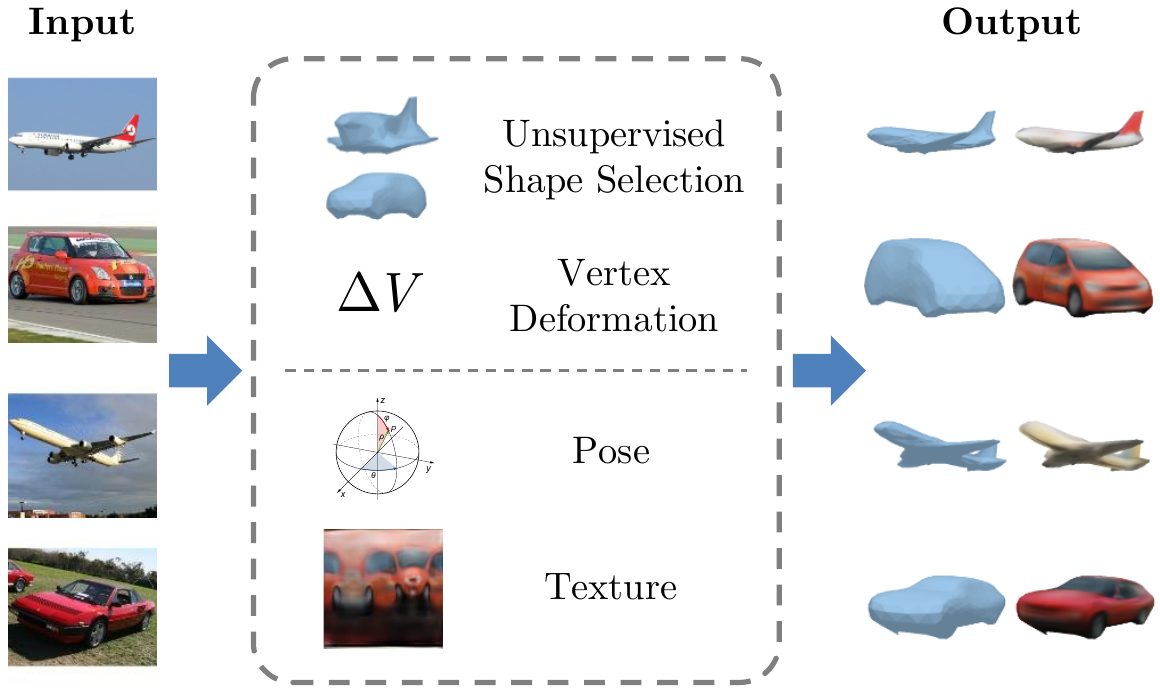}
    \end{center}
    \vspace{-5pt}
    \caption{Overview of the proposed approach. The method predicts realistic 3D textured shapes of objects of different categories and their 3D pose from a single RGB image.}
    \label{fig:method_overview}
    \vspace{-1em}
\end{figure}

In this paper, we present a multi-category approach 
that learns to infer the 3D mesh of an object from a single RGB image.
As illustrated in Figure~\ref{fig:method_overview}, the method learns a series of deformable 3D models and predicts a set of instance-specific deformation, pose, and texture based on the input image.
Differently from previous approaches, the proposed framework is trained with images of multiple object categories using only foreground masks and rough camera poses as supervision.
While rough camera poses could depend on the object category, this is not strictly needed for classes that share semantic keypoints.
The method learns several 3D models in an unsupervised manner, \ie without explicit category supervision, starting from a set of spheres and automatically selects the proper one during inference.
Moreover, the instance-specific deformation is inferred by a network that independently predicts the displacement of each vertex of the learned 3D mesh, given the 3D position of the vertex and conditioned on the selected shape
and the visual features extracted from the input image.
The predicted deformation is naturally smooth and the number of vertices and triangles of the 3D mesh can be dynamically changed during training, with either a global or a local subdivision.

To showcase the quality of the proposed method, we present a variety of experiments in different settings on two datasets, namely Pascal3D+~\cite{xiang2014beyond} and CUB~\cite{WahCUB_200_2011}, and run several ablation studies.  
For instance, we test the method on multiple object categories related to the automotive environment of the Pascal3D+ dataset (\ie bicycle, bus, car, and motorbike) and on the entire set of Pascal3D+ categories.
Qualitative and quantitative results confirm the quality of the proposed approach and show that the model is capable of learning category-specific shape priors without direct supervision.

To sum up, our main contributions are as follows:
\begin{itemize}
    \item We present an approach that recovers the 3D shape, pose, and texture of an object from a 2D image. The method is trained using image collections with foreground masks and coarse camera poses, but no explicit category nor 3D supervision.
    \item Our multi-category framework learns to distinguish between different object categories and produces meaningful meanshapes starting from a set of 3D spheres.
    \item Our approach predicts single vertex deformations, resulting in smooth 3D surfaces and enabling the dynamic subdivision of the learned meshes.
\end{itemize}

\section{Related Work}
In the last decade, many methods have been proposed to tackle the task of 3D reconstruction from a single image.
However, the majority of these methods require supervisory signals which are hard to obtain in the real world and in the wild, such as 3D 
models~\cite{choy20163d,girdhar2016learning,zhu2017rethinking,mandikal20183d,wang2018pixel2mesh,richter2018matryoshka,xie2019pix2vox,aumentado2020cycle,li2020saliency} 
or multi-view image 
collections~\cite{tatarchenko2016multi,yan2016perspective,gwak2017weakly,wiles2017silnet,tulsiani2017multi,tulsiani2018multi,insafutdinov2018unsupervised,lin2019photometric}.

\begin{table}[t]
    \begin{center}
    \renewcommand{\arraystretch}{1.1}
    \resizebox{1.0\linewidth}{!}{
    \begin{tabular}{l|ccc|c|c|c}
        \multirow{2}{*}{\textbf{Approach}} & \multicolumn{3}{c|}{\textbf{Supervision}} & W/o 3D & Multi & Dynamic \\
        & Keypoint & Camera & Mask & Template & category & subdiv.\\
        \midrule
        CSDM~\cite{kar2015category}             &\x &\x &\x &   &   &   \\
        CMR~\cite{kanazawa2018learning}         &\x &\x &\x &   &   &   \\
        VPL~\cite{kato2019learning}           &   &\x &\x &   &   &   \\
        CSM~\cite{kulkarni2019canonical}        &   &   &\x &   &   &   \\
        A-CSM~\cite{kulkarni2020articulation}   &   &   &\x &   &   &   \\
        IMR~\cite{tulsiani2020implicit}         &   &   &\x &   &   &   \\
        U-CMR~\cite{goel2020shape}              &   &   &\x &   &   &   \\
        UMR~\cite{li2020self}                   &   &   &\x &\V &   &   \\
        \textbf{Ours}                           &   &\x &\x &\V &\V &\V \\
        \hline
    \end{tabular}
    }
    \end{center}
    \vspace{-5pt}
    \caption{
    Comparison between available approaches based on training supervision, independence from offline-computed 3D templates, multi-category and dynamic subdivision support. 
    }
    \label{tab:related}
    \vspace{-1em}
\end{table}

Recently, thanks to the development of several differentiable renderers~\cite{loper2014opendr,kato2018neural,palazzi2018end,liu2019soft,chen2019learning}, a handful of methods~\cite{kar2015category,henderson2018learning,kanazawa2018learning} have shown that the task can be addressed as an inverse graphics problem using fewer supervisory signals, such as 2D segmentation masks and object keypoints.
Following methods have even relaxed these constraints, training without keypoint supervision~\cite{chen2019learning,kato2019self,kato2019learning} or known camera poses~\cite{tulsiani2020implicit,goel2020shape,li2020self}.
However, these methods require image collections of a single object category and some of them need a meaningful initialization of a category-specific shape.
Differently, our method is capable of jointly learning shapes of several object categories using only foreground masks and coarse camera poses as supervision.

Another group of works that exploit differentiable renderers address the reconstruction task as a canonical surface mapping~\cite{kulkarni2019canonical,kulkarni2020articulation} or a surface estimation task~\cite{lei2020pix2surf}.
These methods usually require 3D supervision~\cite{lei2020pix2surf} or category-specific shape templates~\cite{kulkarni2019canonical,kulkarni2020articulation}.
In this paper, we focus on the 3D mesh reconstruction from single-view images without any category-specific template.

Recently, Li \etal~\cite{li2020online} proposed a video-based method and the use of multiple meanshapes 
(referred as ``base shapes'')
that are combined to produce a single deformable shape. 
This is the most similar work to our approach, but it has some key differences. 
Firstly, the meanshapes are defined offline and set before training, thus they are not learned. Then, they are introduced for one single dataset to exclusively cover the intra-class variation.
On the contrary, our meanshapes are learned during training without category supervision and our approach can deal with several object categories and their intra- and inter-class variations.

\begin{figure*}
    \begin{center}
        \includegraphics[width=0.98\linewidth]{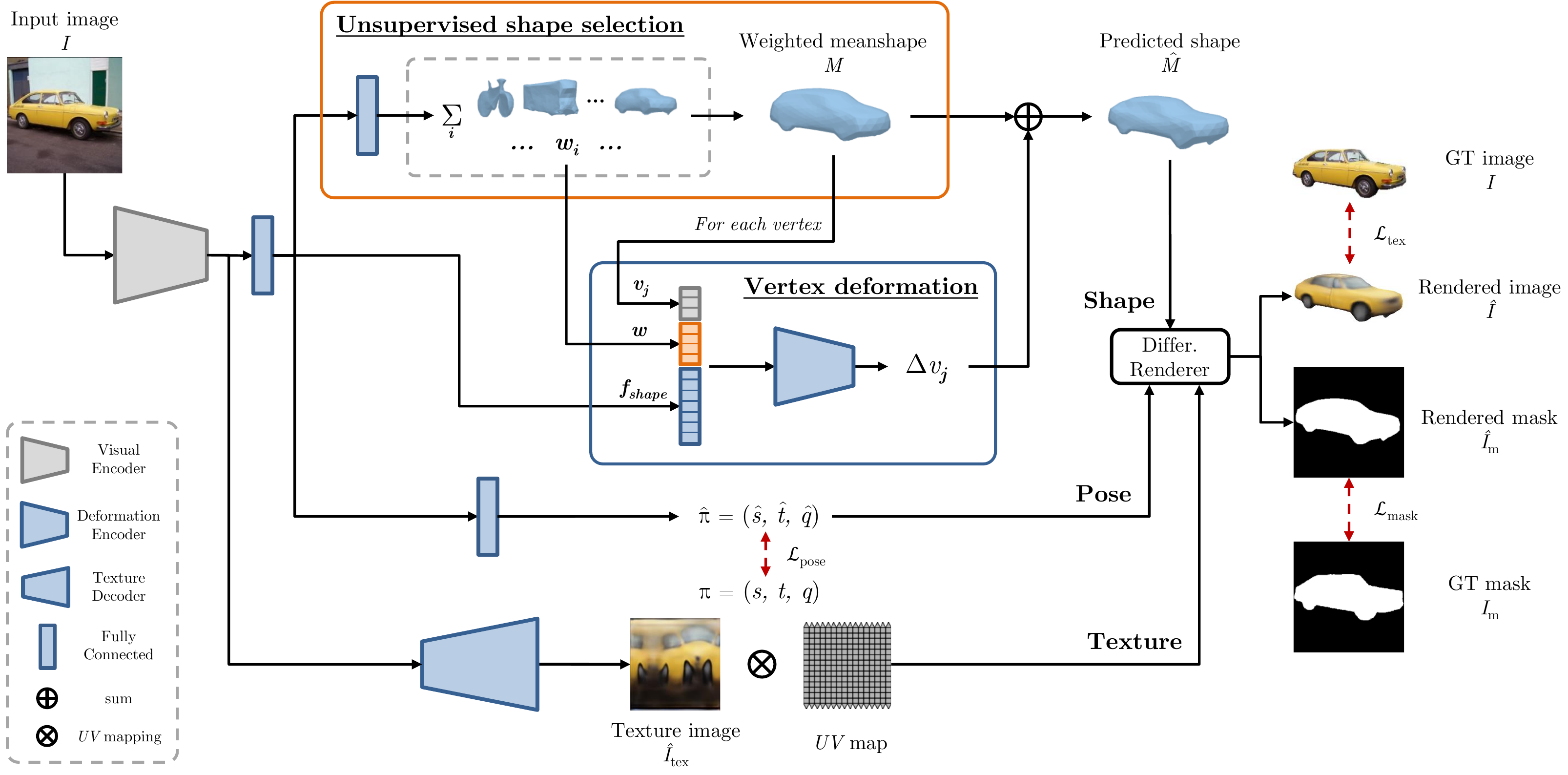}
    \end{center}
    \vspace{-5pt}
    \caption{Overview of the proposed method.
    The \textit{unsupervised shape selection} module predicts the category meanshape while the \textit{vertex deformation} module infers the instance-specific deformation, obtaining the predicted shape.
    In parallel, pose and texture are estimated and then provided, along with the shape, to a differentiable renderer that renders the textured image.
    }
    \label{fig:model_main}
    \vspace{-1em}
\end{figure*}

A comparative study of literature methods is proposed in Table~\ref{tab:related}, highlighting the differences in terms of training supervision, independence from offline-computed 3D templates, 
multi-category and dynamic subdivision
support. As shown, the proposed method still relies on camera supervision, but introduces some unique features.
Indeed, it learns category-specific shape priors in an unsupervised manner and instance-specific deformations from multi-category image collections.
Moreover, the method exploits multiple steps of subdivision during the training process.

\section{Method}
In this section, we present the components of our method, from the input image to the reconstructed 3D textured mesh.
The architecture is illustrated in Figure~\ref{fig:model_main}.

\subsection{Preliminary definitions}
\label{preliminaty_def}

\subsubsec{Shape.} As other approaches in the literature~\cite{kanazawa2018learning,kato2019learning,goel2020shape,li2020self}, we use the triangle mesh as 3D shape representation, which is defined by a set of vertices $V = \{v_j=[x, y, z], \; j = [1, \ldots , k]\}$ and a set of triangle faces $F$.
The faces determine the connectivity between vertices, but are also related to the texture mapping. In our approach, we leverage this connectivity property and dynamically change, during training, the number of vertices and faces of the 3D shape aiming for smoothness and better textures. We refer to this technique as \textit{dynamic mesh subdivision}.

\subsubsec{Texture.}
The triangle mesh texture is represented by a texture image $I_{\text{tex}}$ and a color map $UV$ which maps between the 2D coordinate space of $I_{\text{tex}}$ and the 3D coordinate space of the mesh surface of a sphere.
Thus, the $UV$ mapping is defined by spherical coordinates.

\subsubsec{Pose.} We use a weak-perspective camera projection to define the 3D object pose, as commonly done in literature. This geometric projection is a simplified version of the standard perspective projection. Thus, the object pose is parametrized by a scale factor $s \in \mathbb{R}$, a translation $t=(x,y)$ in image coordinates and a quaternion rotation $q$ obtained by a rotation matrix computed from Euler angles (\ie azimuth, elevation and roll). 
We define $\pi=(s, t, q)$ as the weak-perspective camera projection.

\subsubsec{Rendering.} In order to render a 3D shape with its texture, we rely on the differentiable renderer Soft Rasterizer~\cite{liu2019soft}.
It takes a triangle mesh, a texture image $I_{\text{tex}}$ and an object pose $\pi$ as input and outputs the rendering of the textured object as the RGB image $\hat{I}$ and the foreground mask $\hat{I}_\text{m}$.

\subsection{Multi-category mesh reconstruction}
In this paper, we aim to recover the 3D shape of an object from a single image.
In the literature, this task has been often addressed by splitting it in two parts: on the one hand, the definition or learning of a category-specific base shape, named \textit{meanshape}; on the other hand, the prediction of an instance-specific deformation of the learned shape.
Differently from the majority of previous works (see Table~\ref{tab:related}), we do not need a category-specific initialization of these shapes and propose the joint and unsupervised training of shapes for multiple object categories.
In the following, we provide the details of our approach.

\subsubsec{Feature extraction}
Given an RGB image $I \in \mathbb{R}^{3 \times w \times h}$ as input, the first step of our framework is the extraction of visual features 
with a convolutional encoder (\eg ResNet-18~\cite{he2016deep} in our experiments).
These features are defined as $f_\text{tex}$ and used to estimate the 3D object texture with a specific decoder.
The same features are flattened and mapped into a compact version $f_\text{shape}$, used to recover the shape and its viewpoint.

\subsubsec{Unsupervised shape selection.} 
In contrast to current literature approaches, which are category specific, we propose an unsupervised technique 
that automatically learns to distinguish between different object categories.
Instead of a single meanshape, we define a set of $N$ deformable spheres and use a network to select the instance-specific meanshape according to the input image.
The features $f_\text{shape}$ are passed through a set of fully connected layers and a softmax function.
Then, the resulting scores are used to compute a weighted sum of the mesh vertices and obtain a single mesh,
approximating the argmax function over the $N$ meanshapes.
While the meanshapes are initially defined as spheres, they are updated during the training process and progressively specialize in different object categories.
Formally, let $M_i = (V_i, F)$ be one of the $N$ meanshapes 
and $\mathbf{w} = [w_1,\ldots,w_N]$ be the output of
the network.
The weighted meanshape $M$ is computed as:
\begin{equation}
    M = (V, F) = (\sum_{i=1}^N w_i V_i \,,\, F)
\end{equation}
This mesh $M$ will be deformed according to the object depicted in the input image $I$, as explained in the following.

\subsubsec{Vertex deformation.} Inspired by previous works~\cite{groueix2018papier,park2019deepsdf}, we develop a lightweight network
which deforms the meanshape $M$ taking as input the features $f_\text{shape}$ and the 3D coordinates of a single meanshape vertex $v_j$ at a time. We further condition the output on the selected meanshape giving the weighting scores $\mathbf{w}$ produced by the previous module as additional input. In this way, we enforce the connection between the weighted meanshape $M$ and the predicted deformation.
The module outputs a 3D displacement or deformation $\Delta v_j$ of the vertex $v_j$ in the 3D space.
This approach makes the architecture independent of the number of vertices of the mesh, enabling us to predict the deformation of meshes of variable sizes.
Given a set of deformations $\Delta V$ for each vertex of a meanshape $M$, the predicted shape can be defined as $\hat{M} = M + \Delta V = (V + \Delta V, F)$.

\subsubsec{Dynamic mesh subdivision.} In order to improve the smoothness of the predicted deformed shape, we apply during training a \textit{dynamic subdivision} of the triangle mesh. In particular, we use a global subdivision that divides each triangle of a mesh $M$ in 4 equal parts. 
Other methods that make use of mesh subdivision (\eg~\cite{wang2018pixel2mesh,li2020saliency}) need architectural changes that drastically increase the required memory and the inference time.
On the contrary, our method is not heavily affected by the mesh subdivision operation and does not require any architectural changes, thanks to the per-vertex prediction of the deformation network.

\subsubsec{3D pose regression.} We further predict the object viewpoint with a supervised regression technique using two fully connected layers which take as input the features $f_\text{shape}$ and output a 3D weak-perspective pose $\hat{\pi}=(\hat{s}, \hat{t}, \hat{q})$.

\subsubsec{Texture prediction}
In order to produce a realistic 3D shape, we finally predict the texture that the differentiable renderer applies to the predicted deformed mesh $\hat{M}$.
Similar to the work of Goel \etal~\cite{goel2020shape},
we use a convolutional decoder
that takes as input the visual features $f_\text{tex}$, which preserve the spatiality, and directly outputs an RGB image $\hat{I}_\text{tex}$.
The texture is mapped onto the $UV$ space of the shape, which is homeomorphic to a sphere, so that it can be exploited by the renderer to produce the final image $\hat{I}$.

\subsection{Losses and priors}
The shape prediction is supervised only by two annotated information that are the binary object mask $I_\text{m}$ and the 3D camera pose $\pi$.

We first handle the shape deformation applying a mask loss $\mathcal{L}_\text{mask}=||I_\text{m} - \hat{I}_\text{m}||^2_2$ where $\hat{I}_\text{m}$ is the binary object mask produced by the renderer using the ground truth pose $\pi$.
In addition to this loss, we also use some priors in order to maintain a certain smoothness of the object surface. The first prior is a laplacian smoothing loss $\mathcal{L}_\text{smooth}=||LV||_2$ where the Laplace-Beltrami operator~\cite{sorkine2006differential} minimizes the mean curvature; we apply this smoothing prior both to the predicted deformations $\Delta V$ and the vertices of the deformed shape $\hat{M}$. The second prior is a regularization term $\mathcal{L}_\text{def}=||\Delta V||_2$ which prevents the network from learning large deformations and helps to produce more realistic meanshapes.
Our final shape loss is represented by:
\begin{equation}
    \mathcal{L}_\text{shape} = \mathcal{L}_\text{mask} + \mathcal{L}_\text{smooth} + \mathcal{L}_\text{def} 
\end{equation}

For the pose regression module we use a loss defined as:
\begin{equation}
    \mathcal{L}_\text{pose} = ||\hat{s} - s||_2^2 + ||\hat{t} - t||_2^2 + 
                              \left( 1 - |q \ast (\hat{q} \odot -\hat{q})| \right)
\label{pose_loss}
\end{equation}
where the first two terms consist of the mean squared error for scale and translation and the last term is the geodesic quaternion loss.
The operator $\ast$ is the Hamilton product and $\odot$ the concatenation between the original quaternion and its version rotated by $360$ degrees, representing the same rotation.
Moreover, following the approach proposed by Pavllo \etal~\cite{pavllo2018quaternet}, we further regularize the quaternion prediction with the penalty term $\mathcal{L}_\text{pose\_reg}=w^2 + x^2 + y^2 + z^2 -1^2$
that forces the quaternion to have unit length 
and thus representing a valid rotation.
The overall camera loss is set as:
\begin{equation}
    \mathcal{L}_\text{cam} = \mathcal{L}_\text{pose} + \mathcal{L}_\text{pose\_reg}
\end{equation}

In order to produce realistic colors and details for the object texture, we convert the rendered RGB image and the masked input image to the LAB color space and apply the following losses: 
a color loss $\mathcal{L}_\text{color}=||\hat{I}_{ab} - (I \cdot I_\text{m})_{ab}||^2_2$ on the AB channels for more faithful texture details and a style loss $\mathcal{L}_\text{style}=||\hat{I}_{L} - (I \cdot I_\text{m})_{L}||^2_2$ on the L channel for sharper high-frequency details. 
Moreover, we apply a perceptual loss $\mathcal{L}_\text{percept}=F_\text{dist}(\hat{I}, I \cdot I_\text{m})$ where $F_\text{dist}$ is the metric defined by Zhang \etal~\cite{zhang2018unreasonable} using a VGG16 backbone as feature extractor. The final texture loss is defined by:
\begin{equation}
    \mathcal{L}_\text{tex} = \mathcal{L}_\text{color} + \mathcal{L}_\text{style} + \mathcal{L}_\text{percept}
\end{equation}

The overall objective applied during training is a weighted sum of the shape, camera, and texture losses,
obtaining a balanced learning of the different network modules.
For more details about the loss weights, please refer to the supplementary material.

\section{Experiments}
In this section, we firstly present the employed datasets and the experimental setting.
Then, we present quantitative and qualitative evaluations of our approach in comparison with literature methods.
Finally, we report an ablation study on the key elements of the proposed approach.

\subsection{Datasets and Experimental Setting}
Two common datasets, namely Pascal3D+~\cite{xiang2014beyond} and CUB-200-2011~\cite{WahCUB_200_2011}, have been used  
to evaluate the proposed approach on a diverse set of object categories and, at the same time, to obtain a comparison with the current state-of-the-art methods.
As done in previous works~\cite{kanazawa2018learning,goel2020shape}, 2D image collections, foreground masks and coarse camera/object poses -- manually or automatically annotated -- are used for training.
We do not take advantage of annotated keypoint positions nor coarse 3D model correspondences.

\subsubsec{Pascal3D+}
The Pascal3D+ dataset~\cite{xiang2014beyond} contains images of $12$ object classes, from both PASCAL VOC~\cite{everingham2015pascal,hariharan2011semantic} and ImageNet~\cite{deng2009imagenet}, associated with 3D category-level models and coarse viewpoints~\cite{sun2018pix3d,palazzi2020warp,simoni2021future,simoni2021improving}.
Manually-annotated foreground masks are available for the PASCAL VOC subset, while an off-the-shelf segmentation algorithm~\cite{he2017mask} is used for the other subset, as done in previous works~\cite{kanazawa2018learning,goel2020shape,tulsiani2020implicit}.
We evaluate the system using the same train/test split and categories, \ie \textit{aeroplane} and \textit{car}, of the competitors.
In addition, we use the segmentation masks obtained by the novel PointRend architecture~\cite{kirillov2020pointrend} and evaluate our model on a set of automotive classes, \ie \textit{bicycle, bus, car, motorbike},
and on the entire set of $12$ classes in the ablation study.

\subsubsec{CUB}  
We also use the images of 200 bird species and their
foreground masks provided in CUB-200-2011~\cite{WahCUB_200_2011} and the camera poses computed by Kanazawa \etal~\cite{kanazawa2018learning}, as done in previous works~\cite{kanazawa2018learning,goel2020shape,tulsiani2020implicit}.
The dataset also contains 312 binary attribute labels divided in several categories.

\subsubsec{Network architecture} Our model is composed of 5 modules: (i) a visual encoder, defined as a pre-trained ResNet-18, with an additional convolutional layer,
(ii) an unsupervised shape selection module composed of two fully connected layers and a softmax activation function, (iii) a vertex deformation network with four 512-dimensional fully connected layers with random dropout and a tanh activation function, (iv) a camera pose regressor with two fully connected layers and random dropout, and (v) a texture decoder that follows the 
implementation of the SPADE architecture~\cite{park2019semantic} with 6 upsampling steps.
Additional details are available in the supplementary material.

\subsubsec{Training procedure} We train our network on both datasets for $500$ epochs with an initial learning rate of $1e^{-4}$. The meanshapes are initialized as icospheres with 162 vertices and 320 faces (corresponding to the subdivision level $3$). After $350$ epochs, we apply the dynamic subdivision to the 3D shapes (roughly obtaining the subdivision level $4$) and reduce the learning rate to $1e^{-5}$. Our final 3D shape has
roughly
the same number of vertices and faces as the competitor approaches~\cite{kanazawa2018learning, goel2020shape} which use a deformable template with subdivision level fixed to $4$.

All input images are cropped using the object bounding box and resized to a dimension of $256 \times 256$ and the model predicts a texture image of the same size. As data augmentation, we apply standard random jittering on the bounding box size and location and random horizontal image flipping.
In addition, instead of forcing the shape to be symmetric with post-processing steps (as done in other works, \eg \cite{kanazawa2018learning,goel2020shape,li2020self}),
we force the network to predict symmetric shapes with the following approach, similar to what is done 
in the work of Wu \etal~\cite{wu2020unsupervised}.
During training, the predicted shape (\ie its pose) is randomly rotated by 180 degrees around the vertical axis and compared with the flipped versions of the ground truth image and mask.
In this way, the network is forced to predict symmetric shapes (along the vertical axis) and thus to consistently minimize the losses without computational overhead.

We use a batch size of 16 and Adam~\cite{kingma2014adam} as optimizer with a momentum of 0.9. The code is developed using the PyTorch~\cite{paszke2017automatic} framework.

\subsection{Results}
In this section, we provide a thorough comparison between the proposed method and the competitors on the two previously presented datasets, Pascal3D+ and CUB.

\begin{table}[t]
    \begin{center}
    \renewcommand{\arraystretch}{1.1}
    \resizebox{1\linewidth}{!}{
    \begin{tabular}{lc|cc|c}
        \textbf{Approach} & \textbf{Training} & \textbf{Aeroplane} & \textbf{Car} & \textbf{Avg} \\
        \midrule
        CSDM~\cite{kar2015category} & indep. & $0.400$ & $0.600$ & $0.500$ \\
        DRC~\cite{tulsiani2017multi} & indep. & $0.420$ & $0.670$ & $0.545$ \\
        CMR~\cite{kanazawa2018learning} & indep. & $\mathbf{0.460}$ & $0.640$ & $0.550$ \\
        IMR~\cite{tulsiani2020implicit} & indep. & $0.440$ & $0.660$ & $0.550$ \\
        U-CMR~\cite{goel2020shape} & indep. & - & $0.646$ & - \\
        \textbf{Ours} ($N$ meanshapes) & indep. & $\mathbf{0.460}$ & $\mathbf{0.684}$ & $\mathbf{0.572}$ \\
        \midrule
        \textbf{Ours} ($2$ meanshapes) & \textbf{joint} & $\mathbf{0.448}$ & $\mathbf{0.686}$ & $\mathbf{0.567}$ \\
        \bottomrule
    \end{tabular}
    }
    \end{center}
    \vspace{-5pt}
    \caption{3D IoU on Pascal3D+ dataset~\cite{xiang2014beyond}. 
    Our method is trained on aeroplanes and cars independently using $N$ meanshapes (one for each subclass) or on aeroplanes and cars jointly with 2 meanshapes.
    }
    \label{tab:pascal_single_col}
    \vspace{-0.5em}
\end{table}

\begin{figure}[t]
    \centering
    \setlength{\tabcolsep}{1.5pt}
    \renewcommand{\arraystretch}{1}
    \begin{tabular}{cccccc}
        & \footnotesize{$+60\degree$} & \footnotesize{$+120\degree$} & \footnotesize{$+180\degree$} & \footnotesize{$+240\degree$} & \footnotesize{$+300\degree$} \\
        \includegraphics[width=0.154\linewidth,trim=10pt 60pt 10pt 40pt,clip]{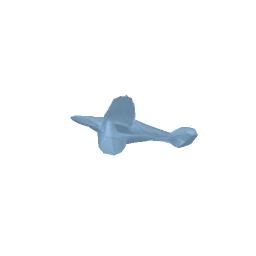} &
        \includegraphics[width=0.154\linewidth,trim=10pt 60pt 10pt 40pt,clip]{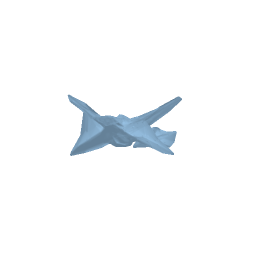} &
        \includegraphics[width=0.154\linewidth,trim=10pt 60pt 10pt 40pt,clip]{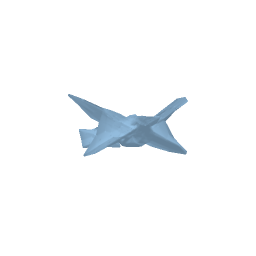} &
        \includegraphics[width=0.154\linewidth,trim=10pt 60pt 10pt 40pt,clip]{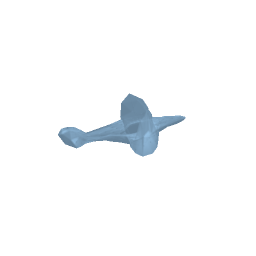} &
        \includegraphics[width=0.154\linewidth,trim=10pt 60pt 10pt 40pt,clip]{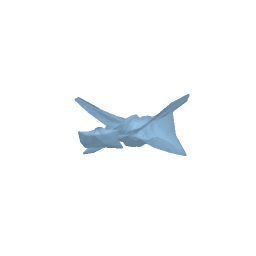} &
        \includegraphics[width=0.154\linewidth,trim=10pt 60pt 10pt 40pt,clip]{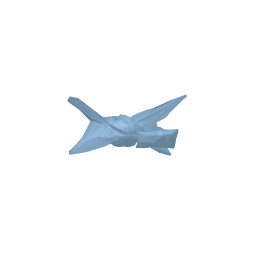} \\
        \includegraphics[width=0.154\linewidth,trim=10pt 60pt 10pt 40pt,clip]{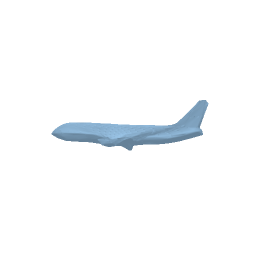} &
        \includegraphics[width=0.154\linewidth,trim=10pt 60pt 10pt 40pt,clip]{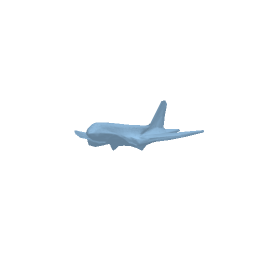} &
        \includegraphics[width=0.154\linewidth,trim=10pt 60pt 10pt 40pt,clip]{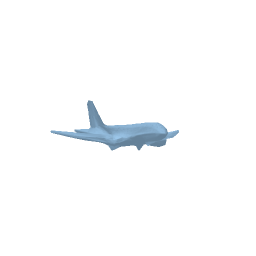} &
        \includegraphics[width=0.154\linewidth,trim=10pt 60pt 10pt 40pt,clip]{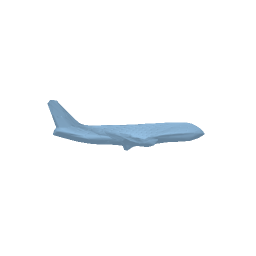} &
        \includegraphics[width=0.154\linewidth,trim=10pt 60pt 10pt 40pt,clip]{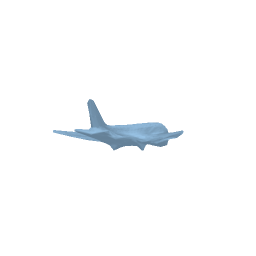} &
        \includegraphics[width=0.154\linewidth,trim=10pt 60pt 10pt 40pt,clip]{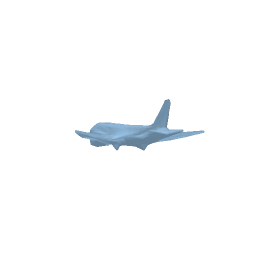} \\
        \includegraphics[width=0.154\linewidth,trim=10pt 40pt 10pt 40pt,clip]{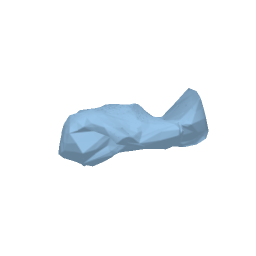} &
        \includegraphics[width=0.154\linewidth,trim=10pt 40pt 10pt 40pt,clip]{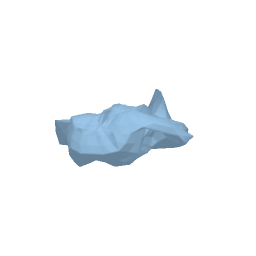} &
        \includegraphics[width=0.154\linewidth,trim=10pt 40pt 10pt 40pt,clip]{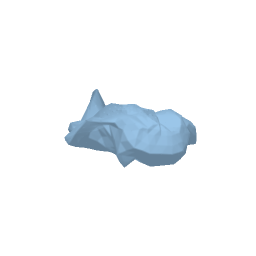} &
        \includegraphics[width=0.154\linewidth,trim=10pt 40pt 10pt 40pt,clip]{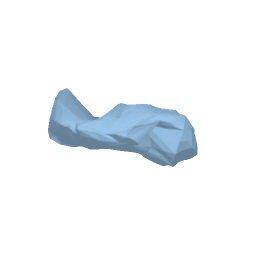} &
        \includegraphics[width=0.154\linewidth,trim=10pt 40pt 10pt 40pt,clip]{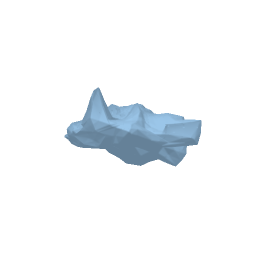} &
        \includegraphics[width=0.154\linewidth,trim=10pt 40pt 10pt 40pt,clip]{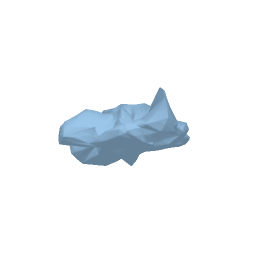} \\
        \midrule
        \includegraphics[width=0.154\linewidth,trim=10pt 40pt 10pt 40pt,clip]{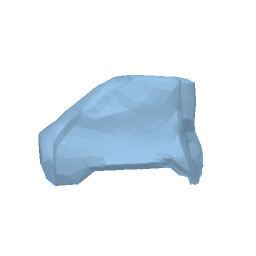} &
        \includegraphics[width=0.154\linewidth,trim=10pt 40pt 10pt 40pt,clip]{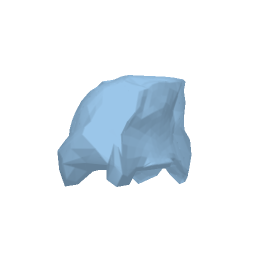} &
        \includegraphics[width=0.154\linewidth,trim=10pt 40pt 10pt 40pt,clip]{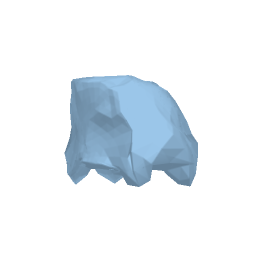} &
        \includegraphics[width=0.154\linewidth,trim=10pt 40pt 10pt 40pt,clip]{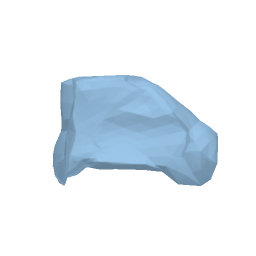} &
        \includegraphics[width=0.154\linewidth,trim=10pt 40pt 10pt 40pt,clip]{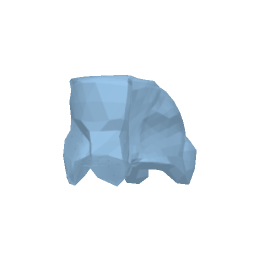} &
        \includegraphics[width=0.154\linewidth,trim=10pt 40pt 10pt 40pt,clip]{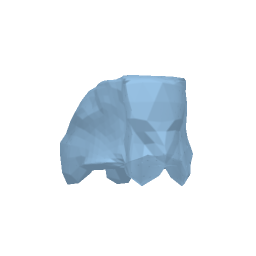} \\
        \includegraphics[width=0.154\linewidth,trim=10pt 60pt 10pt 40pt,clip]{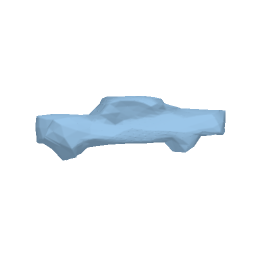} &
        \includegraphics[width=0.154\linewidth,trim=10pt 60pt 10pt 40pt,clip]{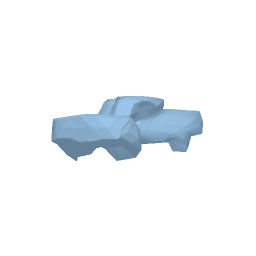} &
        \includegraphics[width=0.154\linewidth,trim=10pt 60pt 10pt 40pt,clip]{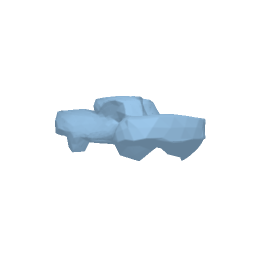} &
        \includegraphics[width=0.154\linewidth,trim=10pt 60pt 10pt 40pt,clip]{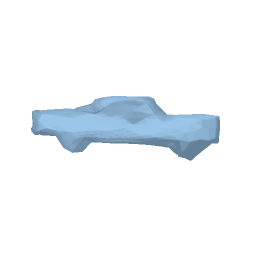} &
        \includegraphics[width=0.154\linewidth,trim=10pt 60pt 10pt 40pt,clip]{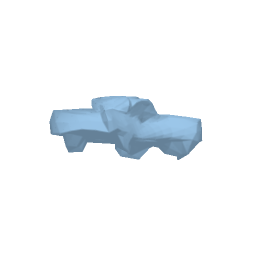} &
        \includegraphics[width=0.154\linewidth,trim=10pt 60pt 10pt 40pt,clip]{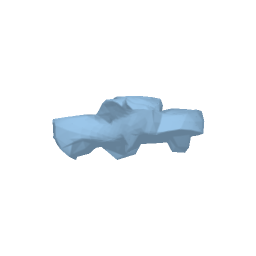} \\
        \includegraphics[width=0.154\linewidth,trim=10pt 40pt 10pt 40pt,clip]{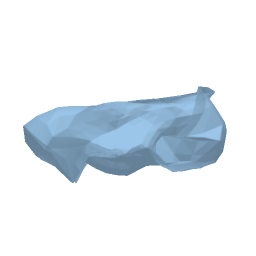} &
        \includegraphics[width=0.154\linewidth,trim=10pt 40pt 10pt 40pt,clip]{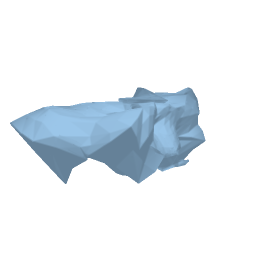} &
        \includegraphics[width=0.154\linewidth,trim=10pt 40pt 10pt 40pt,clip]{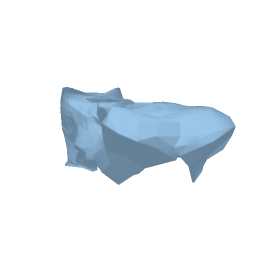} &
        \includegraphics[width=0.154\linewidth,trim=10pt 40pt 10pt 40pt,clip]{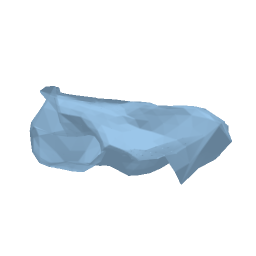} &
        \includegraphics[width=0.154\linewidth,trim=10pt 40pt 10pt 40pt,clip]{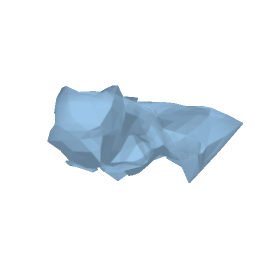} &
        \includegraphics[width=0.154\linewidth,trim=10pt 40pt 10pt 40pt,clip]{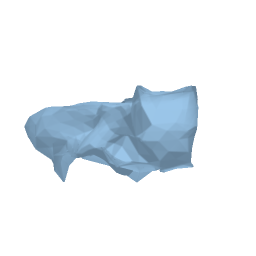} \\
        \midrule
        \includegraphics[width=0.154\linewidth,trim=10pt 60pt 10pt 40pt,clip]{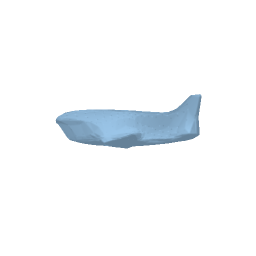} &
        \includegraphics[width=0.154\linewidth,trim=10pt 60pt 10pt 40pt,clip]{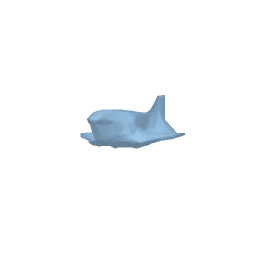} &
        \includegraphics[width=0.154\linewidth,trim=10pt 60pt 10pt 40pt,clip]{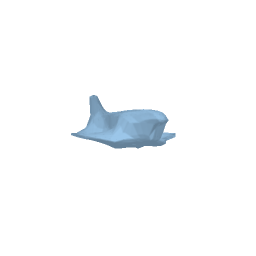} &
        \includegraphics[width=0.154\linewidth,trim=10pt 60pt 10pt 40pt,clip]{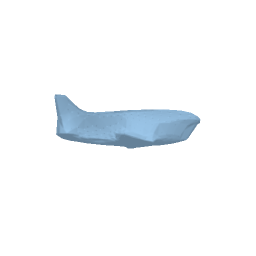} &
        \includegraphics[width=0.154\linewidth,trim=10pt 60pt 10pt 40pt,clip]{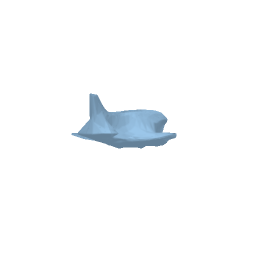} &
        \includegraphics[width=0.154\linewidth,trim=10pt 60pt 10pt 40pt,clip]{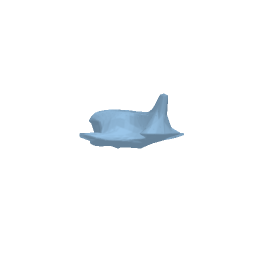} \\
        \includegraphics[width=0.154\linewidth,trim=10pt 60pt 10pt 40pt,clip]{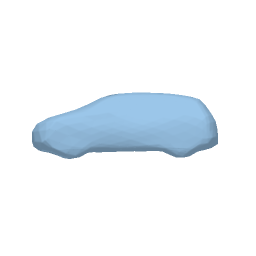} &
        \includegraphics[width=0.154\linewidth,trim=10pt 60pt 10pt 40pt,clip]{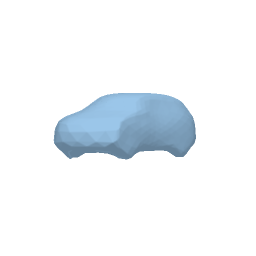} &
        \includegraphics[width=0.154\linewidth,trim=10pt 60pt 10pt 40pt,clip]{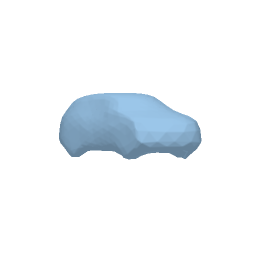} &
        \includegraphics[width=0.154\linewidth,trim=10pt 60pt 10pt 40pt,clip]{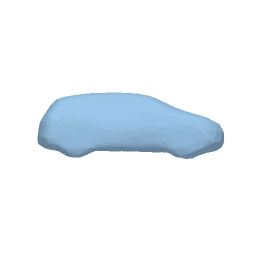} &
        \includegraphics[width=0.154\linewidth,trim=10pt 60pt 10pt 40pt,clip]{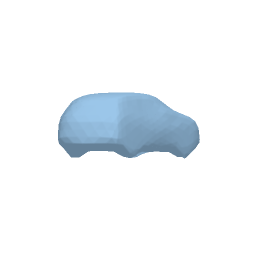} &
        \includegraphics[width=0.154\linewidth,trim=10pt 60pt 10pt 40pt,clip]{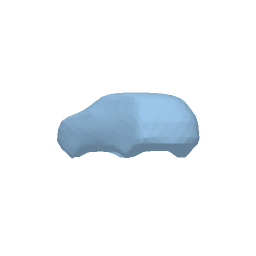}
    \end{tabular}
    \vspace{-5pt}
    \caption{Some of the meanshapes learned during training on Pascal3D+. First group: aeroplane class ($8$ meanshapes); second group: car class ($10$ meanshapes); third group: aeroplane and car classes ($2$ meanshapes).}
    \label{fig:meanshapes_aerocar_maskrcnn}
    \vspace{-1em}
\end{figure}

\subsubsec{Pascal3D+}
We show the results of our method compared to the state of the art on the Pascal3D+ dataset in Table~\ref{tab:pascal_single_col}, using the 3D IoU metric as proposed by Tulsiani \etal \cite{tulsiani2017multi}. 
We present two different versions of our method.
Firstly, we employ the same approach used by competitors: train a different model for each class of Pascal3D+ (experiments marked as ``independent training'').
In this case, we set the number of meanshapes equal to the number of subclasses of Pascal3D+, \ie $N = 8$ for the aeroplane class, $N = 10$ for the car class.
As reported in the second-to-last row of Table \ref{tab:pascal_single_col}, our method 
can leverage the use of multiple meanshapes and the dynamic subdivision
obtaining state-of-the-art results on this dataset.
In addition, we jointly train our method on both the aeroplane and the car classes using $2$ meanshapes, and letting the network distinguish between the two classes.
Even in this more complex scenario, we obtain comparable or state-of-the-art scores on both classes (see last row of Table \ref{tab:pascal_single_col}).
The learned meanshapes for these three experiments, \ie training on aeroplanes, on cars, and on aeroplanes and cars jointly, are shown in Figure~\ref{fig:meanshapes_aerocar_maskrcnn}.
We observe that the set of meanshapes on the single classes contains both recognizable and less explainable shapes (Figure~\ref{fig:meanshapes_aerocar_maskrcnn}, top and middle): we refer the reader to the supplementary material for an analysis of the impact of the learned shapes on the weighted meanshape.
On the other hand, the two meanshapes learned in an unsupervised manner using images of aeroplanes and cars correspond to these two classes (Figure~\ref{fig:meanshapes_aerocar_maskrcnn}, bottom). We show qualitative results of the joint setting on aeroplanes and cars in Figure~\ref{fig:qualitative_results} (second block).

\begin{table}[t]
    \begin{center}
    \renewcommand{\arraystretch}{1.1}
    \resizebox{1\linewidth}{!}{
    \begin{tabular}{l|cc|ccc}
        \multirow{2}{*}{\textbf{Approach}} & \multicolumn{2}{c|}{\textbf{Mask IoU} $\uparrow$} & \multicolumn{3}{c}{\textbf{Texture metrics}} \\
        & \textbf{Pred cam} & \textbf{GT cam} & \textbf{SSIM} $\uparrow$ &  \textbf{L1} $\downarrow$ & \textbf{FID} $\downarrow$ \\
        \midrule
        CMR~\cite{kanazawa2018learning} & $\mathbf{0.706}$ & $0.734$ & $\mathbf{0.718}$ & $\mathbf{0.063}$ & $290.32$ \\
        DIB-R~\cite{chen2019learning} & - & $\mathbf{0.757}$ & - & - & - \\
        U-CMR~\cite{goel2020shape} & $0.637$ & - & $0.689$ & $0.077$ & $\mathbf{190.35}$ \\
        \textbf{Ours} ($1$ meanshape) & $0.658$ & $0.721$ & $0.717$ & $0.064$ & $227.24$ \\
        \midrule
        \textbf{Ours} ($14$ meanshapes) & $0.642$ & $0.723$ & $0.715$ & $0.065$ & $231.95$ \\
        \bottomrule
    \end{tabular}
    }
    \end{center}
    \vspace{-5pt}
    \caption{Mask IoU and texture metrics on CUB dataset~\cite{WahCUB_200_2011}.
    Our method is trained using $1$ or $14$ meanshapes.
    }
    \label{tab:cub_single_col}
    \vspace{-0.5em}
\end{table}

\begin{figure}[t]
    \centering
    \setlength{\tabcolsep}{1.5pt}
    \renewcommand{\arraystretch}{1}
    \begin{tabular}{cccccc}
        & \footnotesize{$+60\degree$} & \footnotesize{$+120\degree$} & \footnotesize{$+180\degree$} & \footnotesize{$+240\degree$} & \footnotesize{$+300\degree$} \\
        \includegraphics[width=0.154\linewidth,trim=30pt 80pt 30pt 40pt,clip]{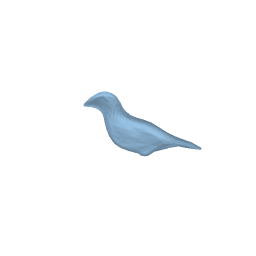} &
        \includegraphics[width=0.154\linewidth,trim=30pt 80pt 30pt 40pt,clip]{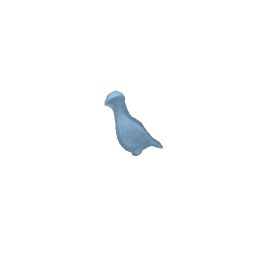} &
        \includegraphics[width=0.154\linewidth,trim=30pt 80pt 30pt 40pt,clip]{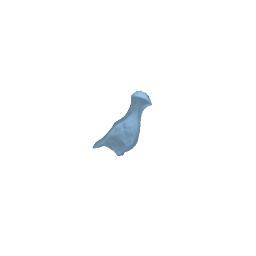} &
        \includegraphics[width=0.154\linewidth,trim=30pt 80pt 30pt 40pt,clip]{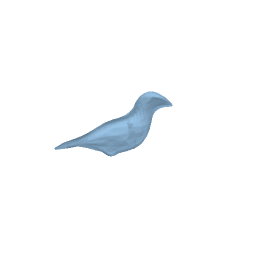} &
        \includegraphics[width=0.154\linewidth,trim=30pt 80pt 30pt 40pt,clip]{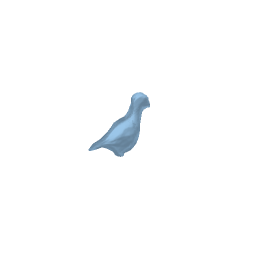} &
        \includegraphics[width=0.154\linewidth,trim=30pt 80pt 30pt 40pt,clip]{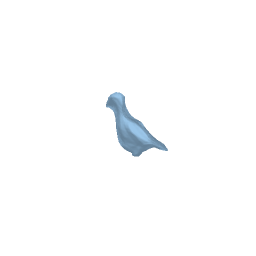} \\
        \includegraphics[width=0.154\linewidth,trim=30pt 60pt 30pt 40pt,clip]{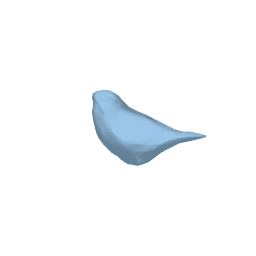} &
        \includegraphics[width=0.154\linewidth,trim=30pt 60pt 30pt 40pt,clip]{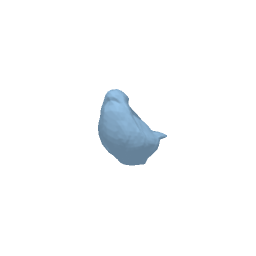} &
        \includegraphics[width=0.154\linewidth,trim=30pt 60pt 30pt 40pt,clip]{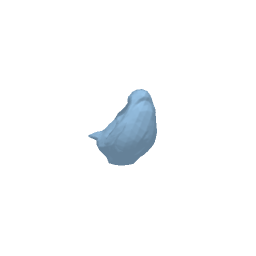} &
        \includegraphics[width=0.154\linewidth,trim=30pt 60pt 30pt 40pt,clip]{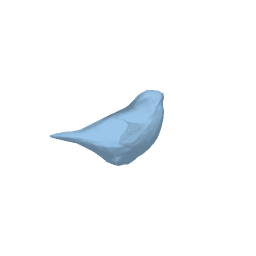} &
        \includegraphics[width=0.154\linewidth,trim=30pt 60pt 30pt 40pt,clip]{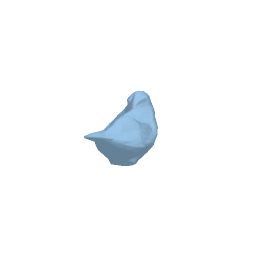} &
        \includegraphics[width=0.154\linewidth,trim=30pt 60pt 30pt 40pt,clip]{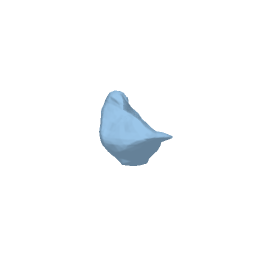} \\
        \includegraphics[width=0.154\linewidth,trim=30pt 60pt 30pt 40pt,clip]{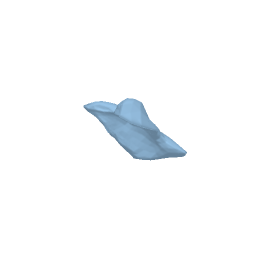} &
        \includegraphics[width=0.154\linewidth,trim=30pt 60pt 30pt 40pt,clip]{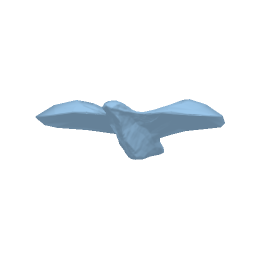} &
        \includegraphics[width=0.154\linewidth,trim=30pt 60pt 30pt 40pt,clip]{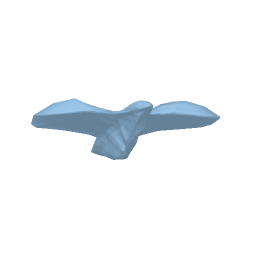} &
        \includegraphics[width=0.154\linewidth,trim=30pt 60pt 30pt 40pt,clip]{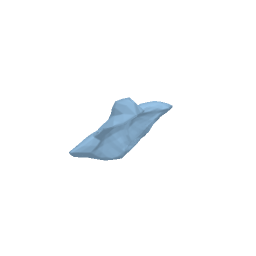} &
        \includegraphics[width=0.154\linewidth,trim=30pt 60pt 30pt 40pt,clip]{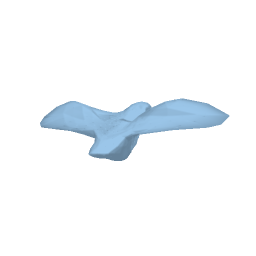} &
        \includegraphics[width=0.154\linewidth,trim=30pt 60pt 30pt 40pt,clip]{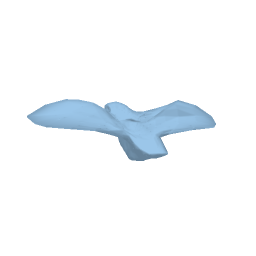} \\
        \includegraphics[width=0.154\linewidth,trim=30pt 60pt 30pt 40pt,clip]{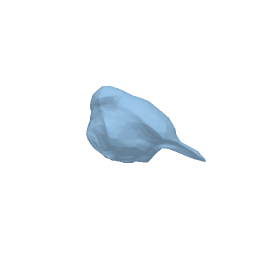} &
        \includegraphics[width=0.154\linewidth,trim=30pt 60pt 30pt 40pt,clip]{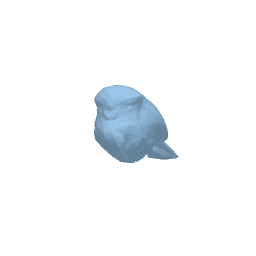} &
        \includegraphics[width=0.154\linewidth,trim=30pt 60pt 30pt 40pt,clip]{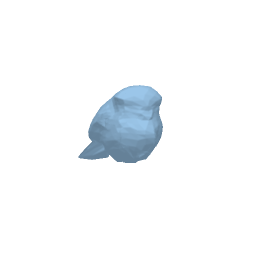} &
        \includegraphics[width=0.154\linewidth,trim=30pt 60pt 30pt 40pt,clip]{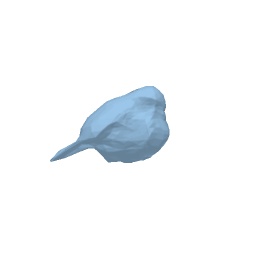} &
        \includegraphics[width=0.154\linewidth,trim=30pt 60pt 30pt 40pt,clip]{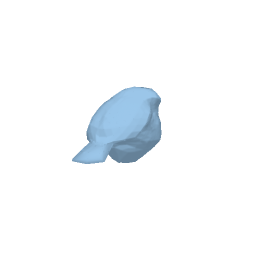} &
        \includegraphics[width=0.154\linewidth,trim=30pt 60pt 30pt 40pt,clip]{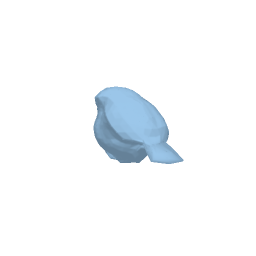} \\
        \includegraphics[width=0.154\linewidth,trim=30pt 60pt 30pt 40pt,clip]{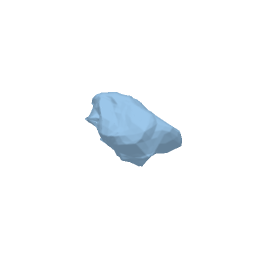} &
        \includegraphics[width=0.154\linewidth,trim=30pt 60pt 30pt 40pt,clip]{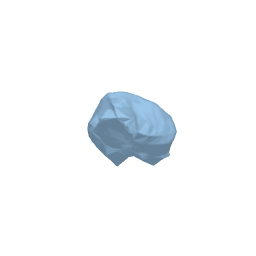} &
        \includegraphics[width=0.154\linewidth,trim=30pt 60pt 30pt 40pt,clip]{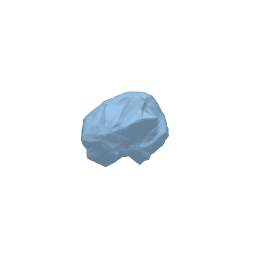} &
        \includegraphics[width=0.154\linewidth,trim=30pt 60pt 30pt 40pt,clip]{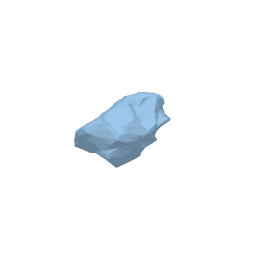} &
        \includegraphics[width=0.154\linewidth,trim=30pt 60pt 30pt 40pt,clip]{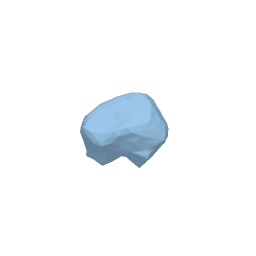} &
        \includegraphics[width=0.154\linewidth,trim=30pt 60pt 30pt 40pt,clip]{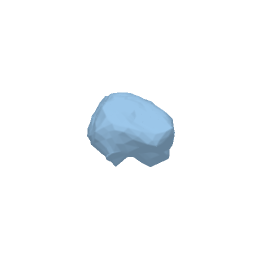} \\
        \includegraphics[width=0.154\linewidth,trim=30pt 60pt 30pt 40pt,clip]{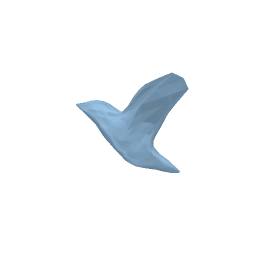} &
        \includegraphics[width=0.154\linewidth,trim=30pt 60pt 30pt 40pt,clip]{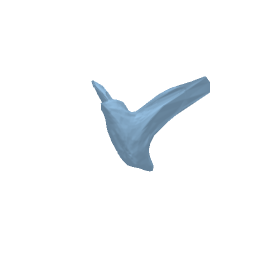} &
        \includegraphics[width=0.154\linewidth,trim=30pt 60pt 30pt 40pt,clip]{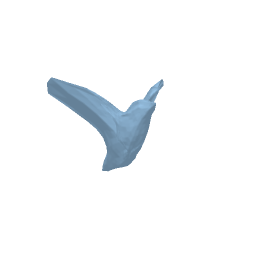} &
        \includegraphics[width=0.154\linewidth,trim=30pt 60pt 30pt 40pt,clip]{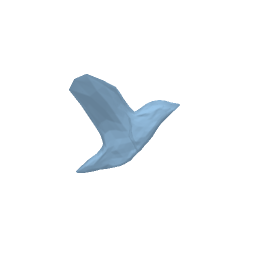} &
        \includegraphics[width=0.154\linewidth,trim=30pt 60pt 30pt 40pt,clip]{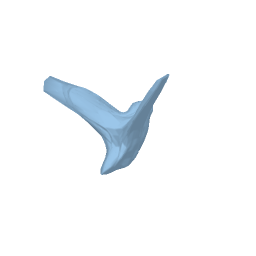} &
        \includegraphics[width=0.154\linewidth,trim=30pt 60pt 30pt 40pt,clip]{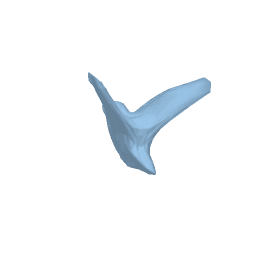}
    \end{tabular}
    \vspace{-5pt}
    \caption{Some of the meanshapes learned during training on the CUB dataset using our method initialized with $14$ spherical meanshapes.}
    \label{fig:meanshapes_cub}
    \vspace{-1em}
\end{figure}

\begin{table*}[t]
    \begin{center}
    \renewcommand{\arraystretch}{1.1}
    \resizebox{0.8\linewidth}{!}{
    \begin{tabular}{c|c|c|cc|ccc}
        \multirow{2}{*}{\textbf{Training classes}} & \textbf{Number of} & \multirow{2}{*}{\textbf{3D IoU} $\uparrow$} & \multicolumn{2}{c|}{\textbf{Mask IoU} $\uparrow$} & \multicolumn{3}{c}{\textbf{Texture metrics}} \\
        & \textbf{meanshapes} & & \textbf{Pred cam} & \textbf{GT cam} & \textbf{SSIM} $\uparrow$ &  \textbf{L1} $\downarrow$ & \textbf{FID} $\downarrow$ \\
        \midrule
        aeroplane, car & 1 & $0.532$ & $0.592$ & $0.689$ & $0.736$ & $0.066$ & $365.01$ \\
        aeroplane, car & 2 & $\mathbf{0.552}$ & $\mathbf{0.671}$ & $\mathbf{0.702}$ & $\mathbf{0.737}$ & $\mathbf{0.062}$ & $\mathbf{344.80}$ \\
        \midrule
        bicycle, bus, car, motorbike & 1 & $0.517$ & $0.665$ & $0.751$ & $0.601$ & $0.100$ & $390.41$ \\
        bicycle, bus, car, motorbike & 4 & $\mathbf{0.543}$ & $\mathbf{0.711}$ & $\mathbf{0.759}$ & $\mathbf{0.607}$ & $\mathbf{0.094}$ & $\mathbf{380.15}$ \\
        \midrule
        $12$ Pascal3D+ classes & 1 & $0.409$ & $0.602$ & $0.670$ & $0.660$ & $0.088$ & $357.51$ \\
        $12$ Pascal3D+ classes & 12 & $\mathbf{0.425}$ & $\mathbf{0.620}$ & $\mathbf{0.685}$ & $\mathbf{0.665}$ & $\mathbf{0.086}$ & $\mathbf{345.90}$ \\
        \bottomrule
    \end{tabular}
    }
    \end{center}
    \vspace{-5pt}
    \caption{Ablation study comparing the usage of several meanshapes (our proposal) against a single meanshape (as a baseline) on Pascal3D+ dataset~\cite{xiang2014beyond} using segmentation masks obtained with PointRend~\cite{kirillov2020pointrend}.}
    \label{tab:pascal_meanshapes}
    \vspace{-0.5em}
\end{table*}

\subsubsec{CUB}
We also evaluate our method on the CUB dataset.
Results in terms of foreground mask IoU and texture metrics (SSIM~\cite{wang2004image}, L1, and FID~\cite{heusel2017gans,lucic2017gans}) are reported in Table~\ref{tab:cub_single_col}.
Differently from the previous case, the CUB dataset does not have a clear subdivision in classes 
and literature approaches have only tested on the whole dataset.
Thus, we test our method in two different settings.
On the one hand, we evaluate the use of a single meanshape (as done by competitors).
On the other hand, we test our method initializing $N$ deformable meanshapes, as done in previous experiments.
We empirically set $N=14$, which is equal to the number of different values of the annotated categorical attribute ``has\_shape'',
and refer the reader to the supplementary material for an analysis of using different numbers of meanshapes on the CUB dataset.
As shown, even if this dataset does contain objects of the same class ``bird'', our method obtains comparable results with respect to literature approaches, on both shape and texture metrics.
Even if the experiment with multiple shapes does not seem to increase the overall scores, it produces a set of insightful meanshapes learned in an unsupervised manner, as shown in Figure~\ref{fig:meanshapes_cub}.
Qualitative results are reported in Figure~\ref{fig:qualitative_results} (first block) and in the supplementary material.

\begin{figure}[t]
    \centering
    \setlength{\tabcolsep}{1.5pt}
    \renewcommand{\arraystretch}{1}
    \begin{tabular}{cccccc}
        & \footnotesize{$+60\degree$} & \footnotesize{$+120\degree$} & \footnotesize{$+180\degree$} & \footnotesize{$+240\degree$} & \footnotesize{$+300\degree$} \\
        \includegraphics[width=0.154\linewidth,trim=10pt 40pt 10pt 40pt,clip]{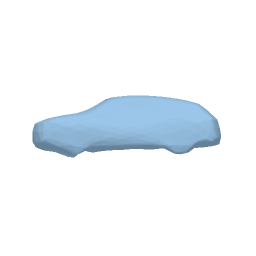} &
        \includegraphics[width=0.154\linewidth,trim=10pt 40pt 10pt 40pt,clip]{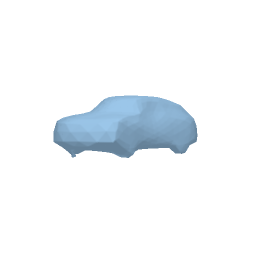} &
        \includegraphics[width=0.154\linewidth,trim=10pt 40pt 10pt 40pt,clip]{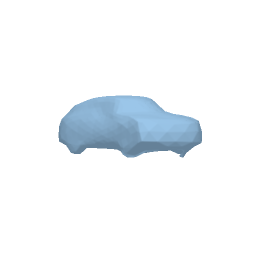} &
        \includegraphics[width=0.154\linewidth,trim=10pt 40pt 10pt 40pt,clip]{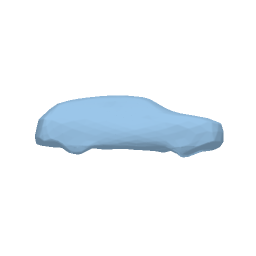} &
        \includegraphics[width=0.154\linewidth,trim=10pt 40pt 10pt 40pt,clip]{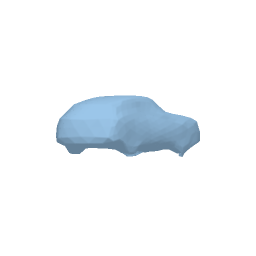} &
        \includegraphics[width=0.154\linewidth,trim=10pt 40pt 10pt 40pt,clip]{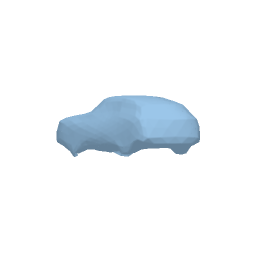} \\
        \includegraphics[width=0.154\linewidth,trim=10pt 40pt 10pt 40pt,clip]{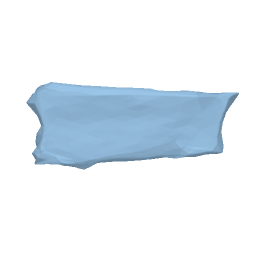} &
        \includegraphics[width=0.154\linewidth,trim=10pt 40pt 10pt 40pt,clip]{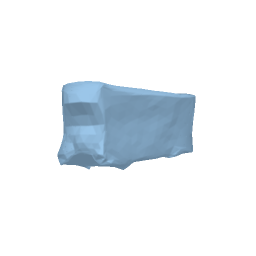} &
        \includegraphics[width=0.154\linewidth,trim=10pt 40pt 10pt 40pt,clip]{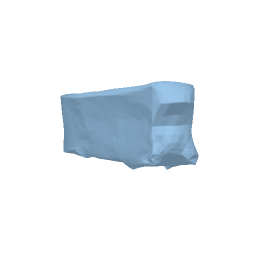} &
        \includegraphics[width=0.154\linewidth,trim=10pt 40pt 10pt 40pt,clip]{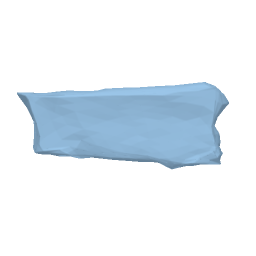} &
        \includegraphics[width=0.154\linewidth,trim=10pt 40pt 10pt 40pt,clip]{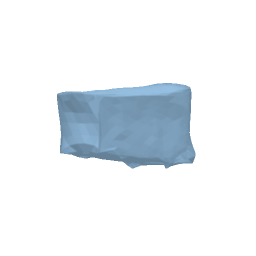} &
        \includegraphics[width=0.154\linewidth,trim=10pt 40pt 10pt 40pt,clip]{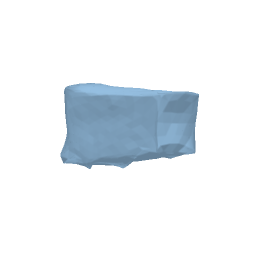} \\
        \includegraphics[width=0.154\linewidth,trim=10pt 40pt 10pt 40pt,clip]{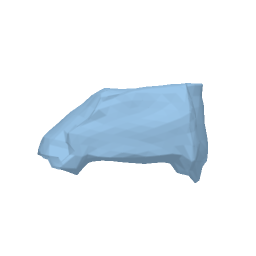} &
        \includegraphics[width=0.154\linewidth,trim=10pt 40pt 10pt 40pt,clip]{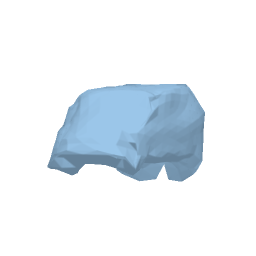} &
        \includegraphics[width=0.154\linewidth,trim=10pt 40pt 10pt 40pt,clip]{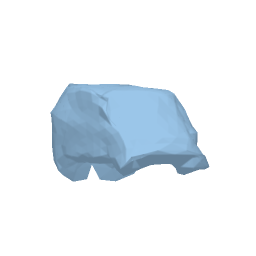} &
        \includegraphics[width=0.154\linewidth,trim=10pt 40pt 10pt 40pt,clip]{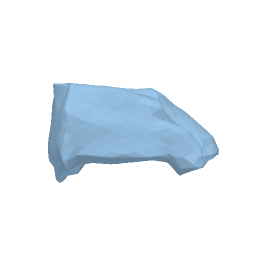} &
        \includegraphics[width=0.154\linewidth,trim=10pt 40pt 10pt 40pt,clip]{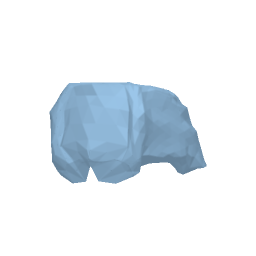} &
        \includegraphics[width=0.154\linewidth,trim=10pt 40pt 10pt 40pt,clip]{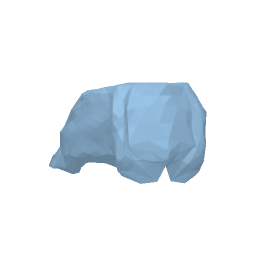} \\
        \includegraphics[width=0.154\linewidth,trim=10pt 40pt 10pt 40pt,clip]{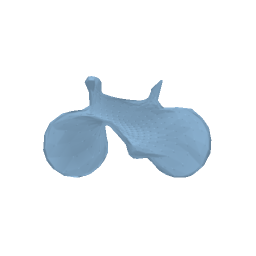} &
        \includegraphics[width=0.154\linewidth,trim=10pt 40pt 10pt 40pt,clip]{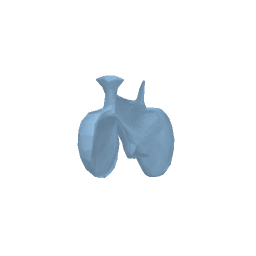} &
        \includegraphics[width=0.154\linewidth,trim=10pt 40pt 10pt 40pt,clip]{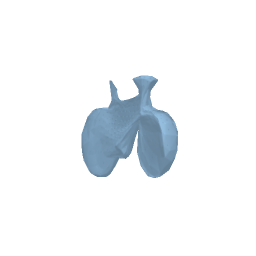} &
        \includegraphics[width=0.154\linewidth,trim=10pt 40pt 10pt 40pt,clip]{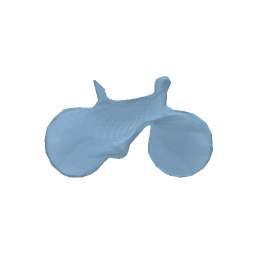} &
        \includegraphics[width=0.154\linewidth,trim=10pt 40pt 10pt 40pt,clip]{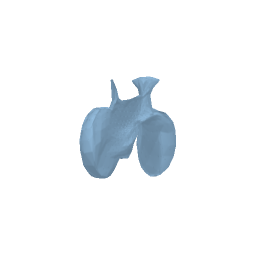} &
        \includegraphics[width=0.154\linewidth,trim=10pt 40pt 10pt 40pt,clip]{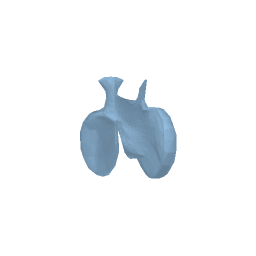}
    \end{tabular}
    \vspace{-5pt}
    \caption{Meanshapes learned during training on the classes bicycle, bus, car, motorbike of the Pascal3D+ dataset~\cite{xiang2014beyond}.}
    \label{fig:meanshapes_4macroclasses}
    \vspace{-1em}
\end{figure}

\subsection{Ablation study}
In this section, we investigate the impact of using one or multiple meanshapes. In addition, we evaluate the influence of the dynamic subdivision approach compared to the static one.
In these experiments, we use the Pascal3D+ dataset and extract precise foreground masks with PointRend~\cite{kirillov2020pointrend}.
Additional ablation studies and qualitative results are available in the supplementary material.

\subsubsec{Unsupervised shape selection}
As our first analysis, we evaluate the impact of the proposed unsupervised shape selection, which enables the training with multiple meanshapes and classes.
We test three different training settings using the following object categories: (i) \textit{aeroplane}, \textit{car}, (ii) \textit{bicycle}, \textit{bus}, \textit{car}, \textit{motorbike}, (iii) all the $12$ Pascal3D+ classes. Each setting has been tested using both a single meanshape or a set of $N$ meanshapes, in order to verify the contribution of the usage of multiple learnable shapes and their unsupervised selection.
The obtained results are reported in Table~\ref{tab:pascal_meanshapes} in terms of 3D IoU, foreground mask IoU and texture metrics. Our approach with multiple meanshapes provide the best results in all the experimental settings.
Furthermore, the meanshapes learned with the four-category setting are depicted in Figure~\ref{fig:meanshapes_4macroclasses}.
Even if the meanshapes do not exactly correspond to the four classes (\eg, the motorbike is missing), the meanshapes are meaningful and represent different object categories.
Qualitative results are shown in Figure~\ref{fig:qualitative_results}. 
In the supplementary material, we further evaluate the average usage of each learned meanshape throughout the test set and the classification accuracy of the unsupervised shape selection module when used as a category classifier.

\begin{table}[t]
    \begin{center}
    \renewcommand{\arraystretch}{1.1}
    \resizebox{1\linewidth}{!}{
    \begin{tabular}{c|cc|ccc}
        \textbf{Subdivision} & \multicolumn{2}{c|}{\textbf{Mask IoU} $\uparrow$} & \multicolumn{3}{c}{\textbf{Texture metrics}} \\
        \textbf{level} & \textbf{Pred cam} & \textbf{GT cam} & \textbf{SSIM} $\uparrow$ &  \textbf{L1} $\downarrow$ & \textbf{FID} $\downarrow$ \\
        \midrule
        3 & $0.701$ & $\mathbf{0.759}$ & $0.600$ & $0.096$ & $395.96$ \\
        4 & $0.685$ & $0.756$ & $0.593$ & $0.101$ & $385.68$ \\
        3 $\rightarrow$ 4 & $\mathbf{0.711}$ & $\mathbf{0.759}$ & $\mathbf{0.607}$ & $\mathbf{0.094}$ & $\mathbf{380.15}$ \\
        \bottomrule
    \end{tabular}
    }
    \end{center}
    \vspace{-5pt}
    \caption{Ablation study comparing different subdivision levels on Pascal3D+ dataset~\cite{xiang2014beyond}. Model trained on 4 classes (bicycle, bus, car, motorbike) using 4 meanshapes.}
    \label{tab:pascal_subdivision_single_col}
    \vspace{-1em}
\end{table}

\begin{figure*}
    \centering
    \setlength{\tabcolsep}{1.6pt}
    \renewcommand{\arraystretch}{0.6}
    \begin{tabular}{c cccccc cc cccccc}
        & \small{Input} & \small{Weighted} & \small{Predicted} & \multicolumn{3}{c}{\small{Predicted shape}} & & & \small{Input} & \small{Weighted} & \small{Predicted} & \multicolumn{3}{c}{\small{Predicted shape}} \\
        & \small{{image}}\vspace{2pt} & \small{{meanshape}} & \small{{shape}} & \multicolumn{3}{c}{\small{{with texture}}} & & & \small{{image}} & \small{{meanshape}} & \small{{shape}} & \multicolumn{3}{c}{\small{{with texture}}} \\
        & \small{${I}$}\vspace{3pt} & \small{${M}$} & \small{${\hat{M}}$} & \multicolumn{3}{c}{${\hat{M}} + {\hat{I}_\text{tex}}$} & & & \small{${I}$} & \small{${M}$} & \small{${\hat{M}}$} & \multicolumn{3}{c}{${\hat{M}} + {\hat{I}_\text{tex}}$} \\
        &
        \includegraphics[width=0.06\linewidth]{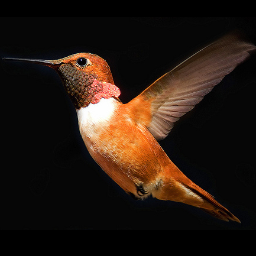} &
        \includegraphics[width=0.075\linewidth,trim=64px 0 64px 0,clip]{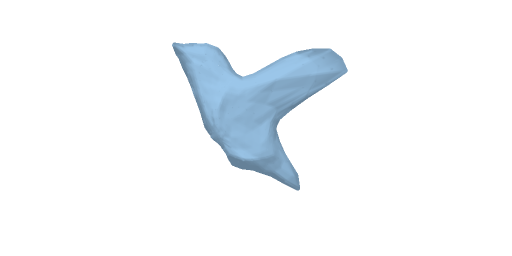} &
        \includegraphics[width=0.075\linewidth,trim=64px 0 64px 0,clip]{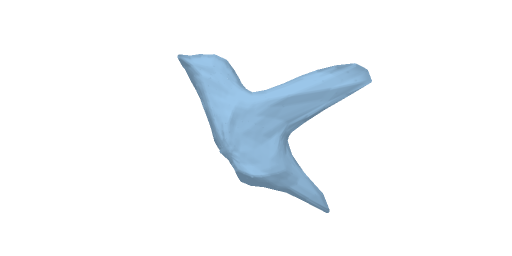} &
        \includegraphics[width=0.075\linewidth,trim=64px 0 64px 0,clip]{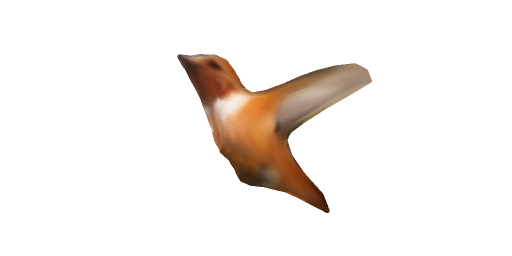} &
        \includegraphics[width=0.075\linewidth,trim=64px 0 64px 0,clip]{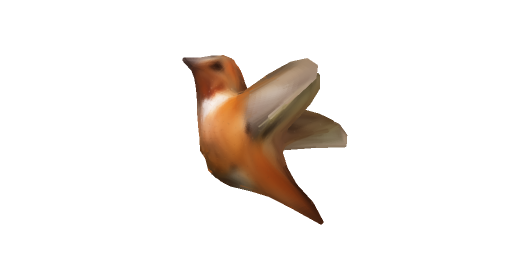} &
        \includegraphics[width=0.075\linewidth,trim=64px 0 64px 0,clip]{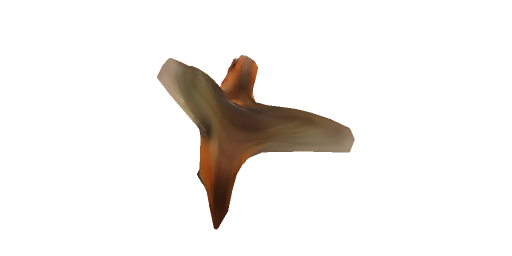} & &
        &
        \includegraphics[width=0.06\linewidth]{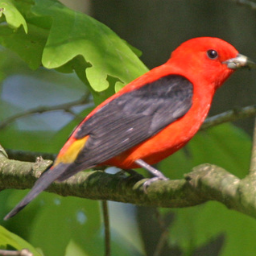} &
        \includegraphics[width=0.075\linewidth,trim=64px 0 64px 0,clip]{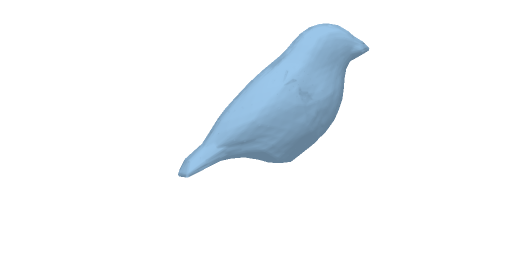} &
        \includegraphics[width=0.075\linewidth,trim=64px 0 64px 0,clip]{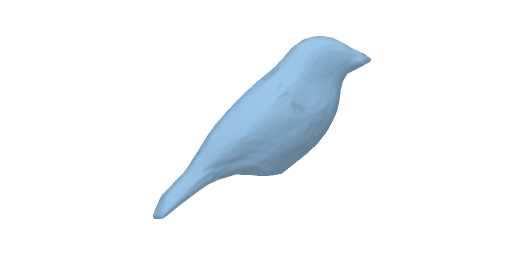} &
        \includegraphics[width=0.075\linewidth,trim=64px 0 64px 0,clip]{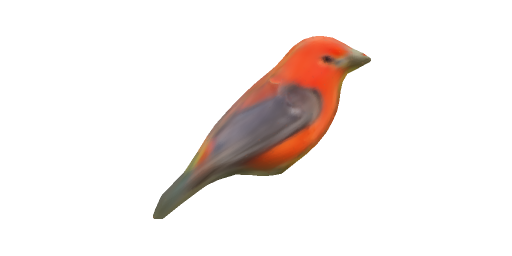} &
        \includegraphics[width=0.075\linewidth,trim=64px 0 64px 0,clip]{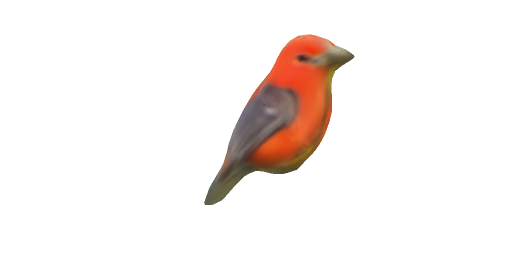} &
        \includegraphics[width=0.075\linewidth,trim=64px 0 64px 0,clip]{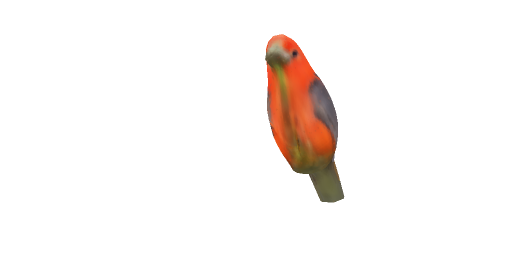} \\
        &
        \includegraphics[width=0.06\linewidth]{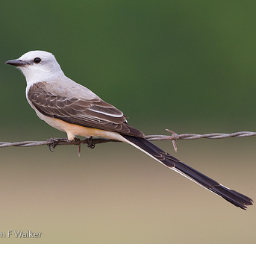} &
        \includegraphics[width=0.075\linewidth,trim=32px 0 96px 0,clip]{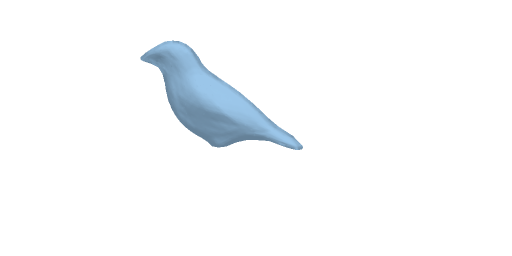} &
        \includegraphics[width=0.075\linewidth,trim=48px 0 80px 0,clip]{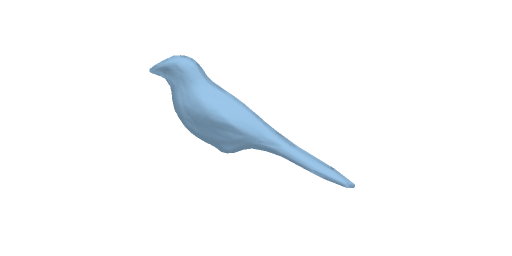} &
        \includegraphics[width=0.075\linewidth,trim=32px 0 96px 0,clip]{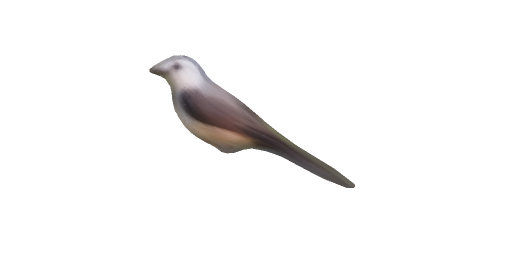} &
        \includegraphics[width=0.075\linewidth,trim=32px 0 96px 0,clip]{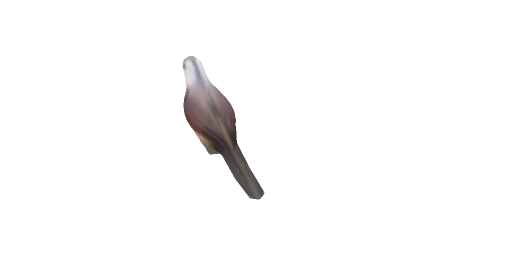} &
        \includegraphics[width=0.075\linewidth,trim=0 0 128px 0,clip]{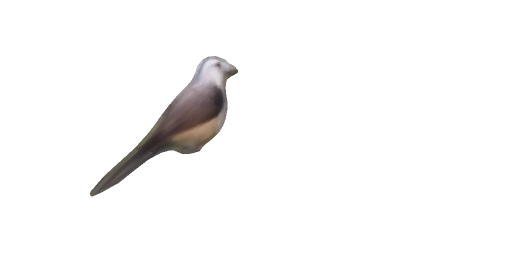} & &
        &
        \includegraphics[width=0.06\linewidth]{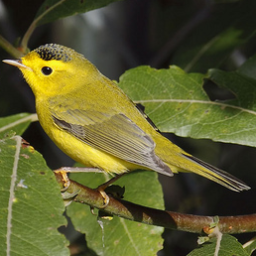} &
        \includegraphics[width=0.075\linewidth,trim=32px 0 96px 0,clip]{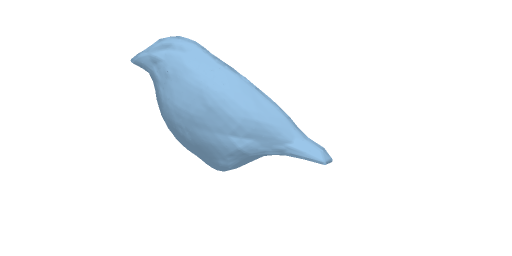} &
        \includegraphics[width=0.075\linewidth,trim=64px 0 64px 0,clip]{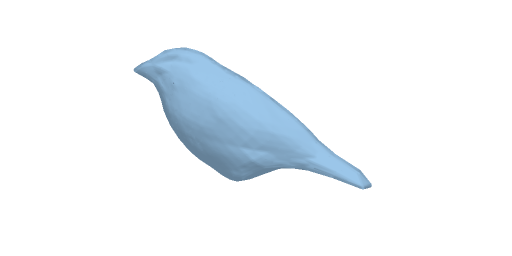} &
        \includegraphics[width=0.075\linewidth,trim=64px 0 64px 0,clip]{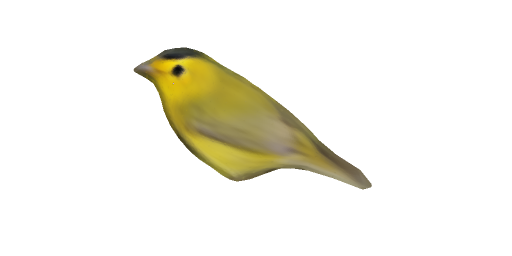} &
        \includegraphics[width=0.075\linewidth,trim=32px 0 96px 0,clip]{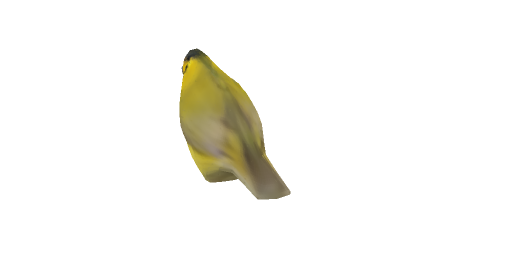} &
        \includegraphics[width=0.075\linewidth,trim=-32px 0 160px 0,clip]{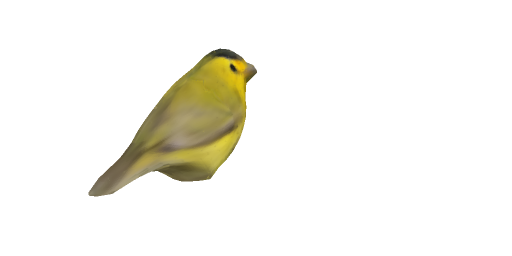} \\
        \midrule
        %
        %
        \rotatebox{90}{\parbox[t]{0.06\linewidth}{\hspace*{\fill}\textbf{\tiny{aeroplane}}\hspace*{\fill}}} &
        \includegraphics[width=0.06\linewidth]{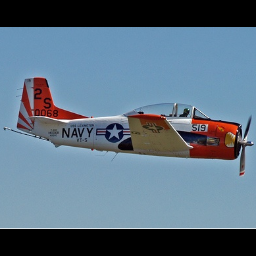} &
        \includegraphics[width=0.07\linewidth,trim=64px 0 64px 0,clip]{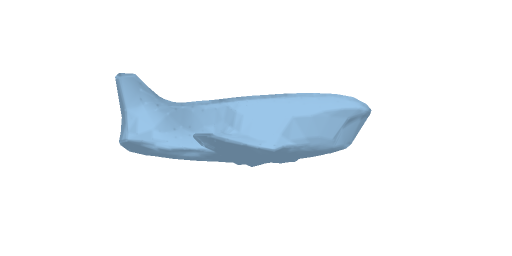} &
        \includegraphics[width=0.07\linewidth,trim=64px 0 64px 0,clip]{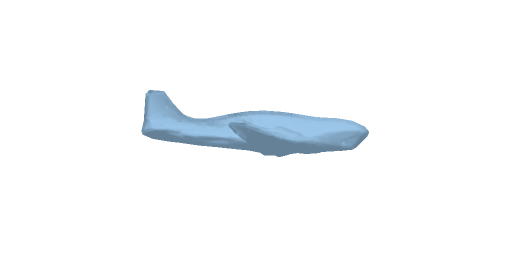} &
        \includegraphics[width=0.07\linewidth,trim=64px 0 64px 0,clip]{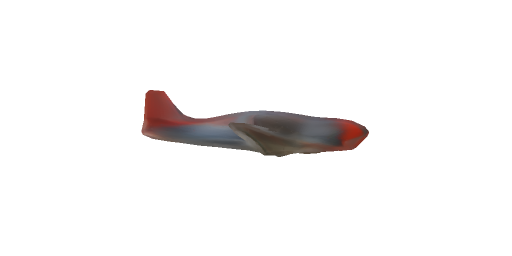} &
        \includegraphics[width=0.07\linewidth,trim=64px 0 64px 0,clip]{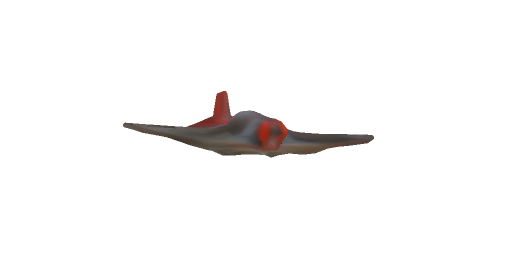} &
        \includegraphics[width=0.07\linewidth,trim=64px 0 64px 0,clip]{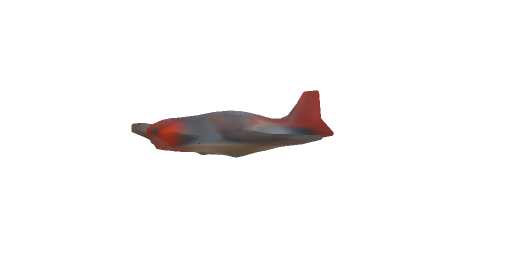} & &
        \rotatebox{90}{\parbox[t]{0.06\linewidth}{\hspace*{\fill}\textbf{\tiny{car}}\hspace*{\fill}}} &
        \includegraphics[width=0.06\linewidth]{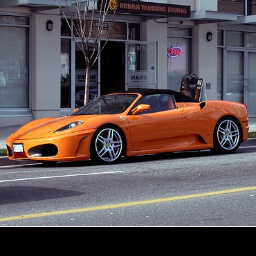} &
        \includegraphics[width=0.07\linewidth,trim=64px 0 64px 0,clip]{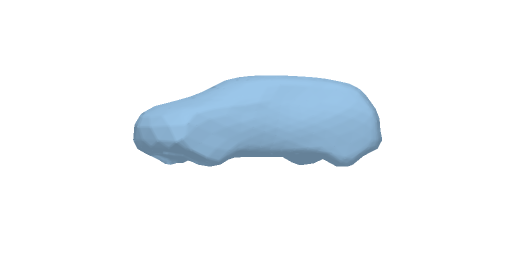} &
        \includegraphics[width=0.07\linewidth,trim=64px 0 64px 0,clip]{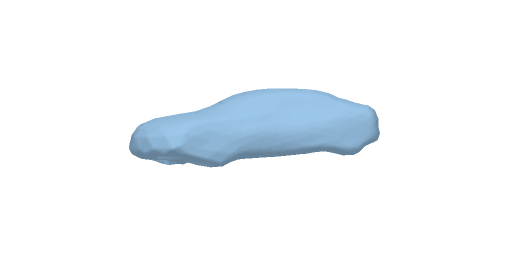} &
        \includegraphics[width=0.07\linewidth,trim=64px 0 64px 0,clip]{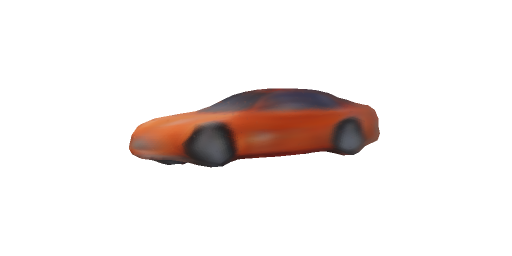} &
        \includegraphics[width=0.07\linewidth,trim=64px 0 64px 0,clip]{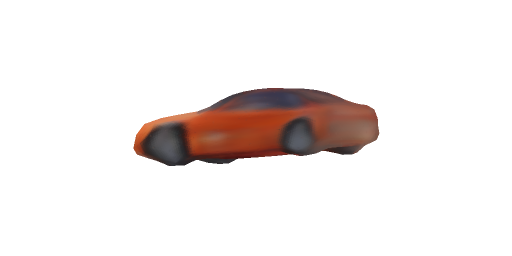} &
        \includegraphics[width=0.07\linewidth,trim=64px 0 64px 0,clip]{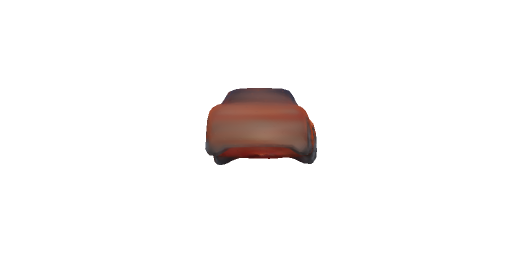} \\
        \midrule
        %
        %
        \rotatebox{90}{\parbox[t]{0.06\linewidth}{\hspace*{\fill}\textbf{\tiny{bicycle}}\hspace*{\fill}}} &
        \includegraphics[width=0.06\linewidth]{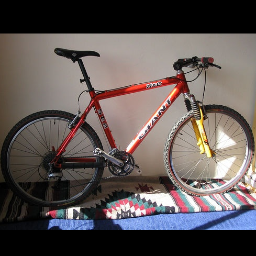} &
        \includegraphics[width=0.075\linewidth,trim=64px 0 64px 0,clip]{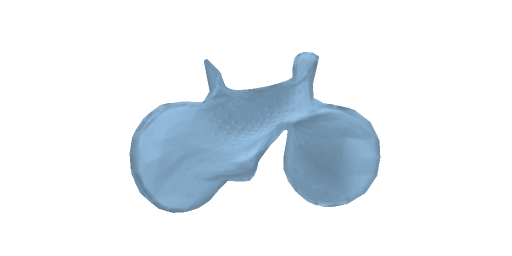} &
        \includegraphics[width=0.075\linewidth,trim=64px 0 64px 0,clip]{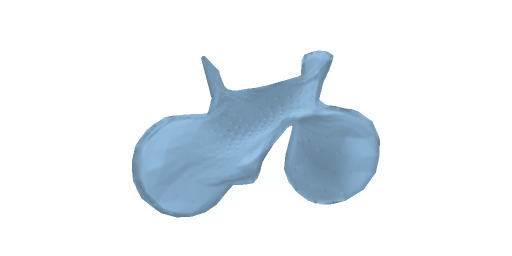} &
        \includegraphics[width=0.075\linewidth,trim=64px 0 64px 0,clip]{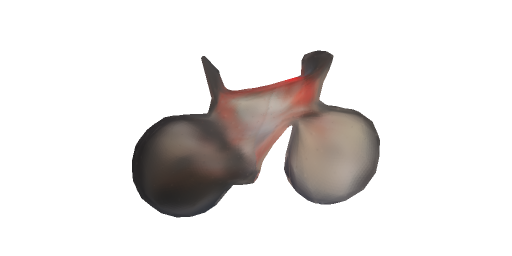} &
        \includegraphics[width=0.075\linewidth,trim=64px 0 64px 0,clip]{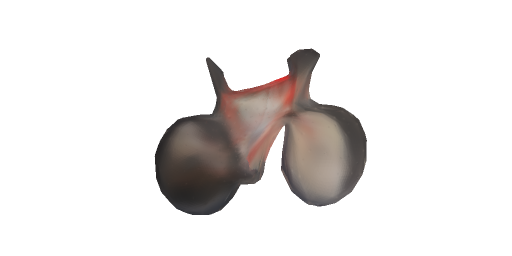} &
        \includegraphics[width=0.075\linewidth,trim=64px 0 64px 0,clip]{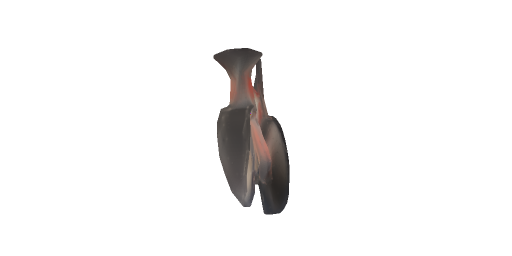} & &
        \rotatebox{90}{\parbox[t]{0.06\linewidth}{\hspace*{\fill}\textbf{\tiny{bus}}\hspace*{\fill}}} &
        \includegraphics[width=0.06\linewidth]{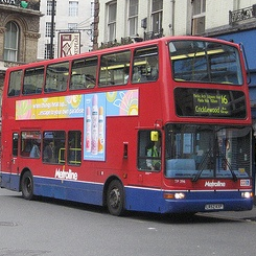} &
        \includegraphics[width=0.075\linewidth,trim=64px 0 64px 0,clip]{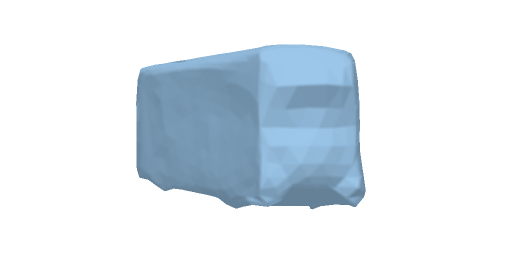} &
        \includegraphics[width=0.075\linewidth,trim=64px 0 64px 0,clip]{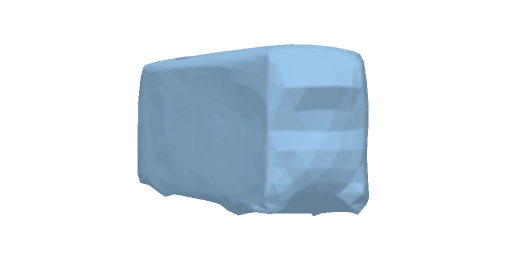} &
        \includegraphics[width=0.075\linewidth,trim=64px 0 64px 0,clip]{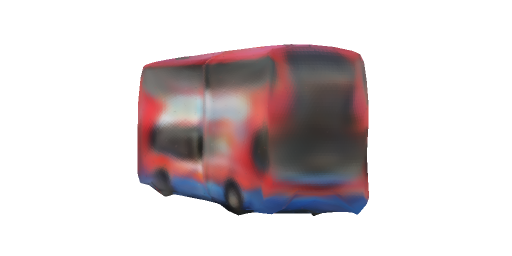} &
        \includegraphics[width=0.075\linewidth,trim=64px 0 64px 0,clip]{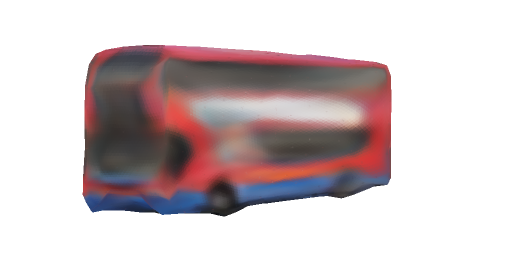} &
        \includegraphics[width=0.075\linewidth,trim=46px 0 82px 0,clip]{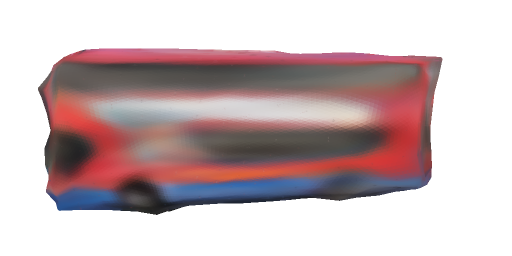} \\
        \rotatebox{90}{\parbox[t]{0.06\linewidth}{\hspace*{\fill}\textbf{\tiny{car}}\hspace*{\fill}}} &
        \includegraphics[width=0.06\linewidth]{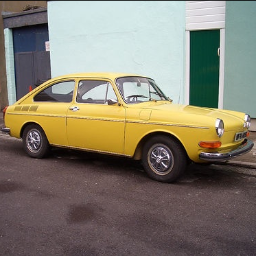} &
        \includegraphics[width=0.075\linewidth,trim=64px 0 64px 0,clip]{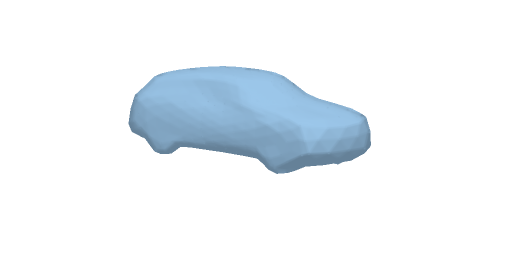} &
        \includegraphics[width=0.075\linewidth,trim=64px 0 64px 0,clip]{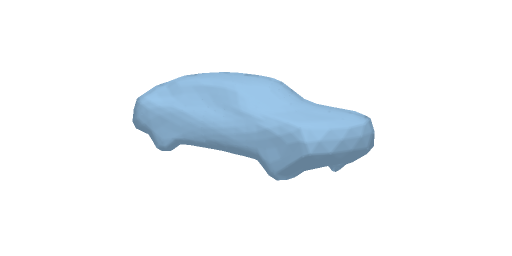} &
        \includegraphics[width=0.075\linewidth,trim=64px 0 64px 0,clip]{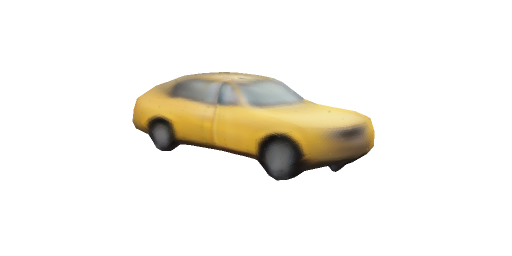} &
        \includegraphics[width=0.075\linewidth,trim=64px 0 64px 0,clip]{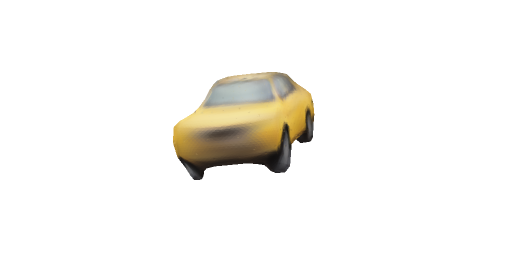} &
        \includegraphics[width=0.075\linewidth,trim=64px 0 64px 0,clip]{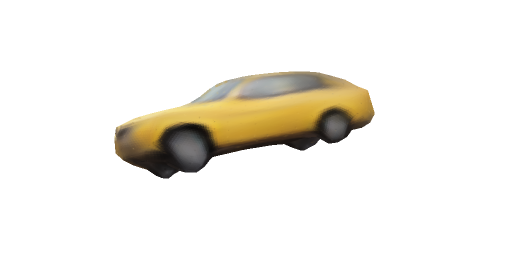} & &
        \rotatebox{90}{\parbox[t]{0.06\linewidth}{\hspace*{\fill}\textbf{\tiny{motorbike}}\hspace*{\fill}}} &
        \includegraphics[width=0.06\linewidth]{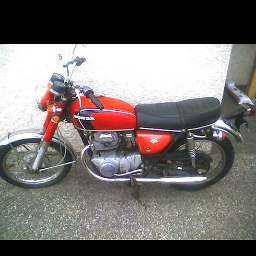} &
        \includegraphics[width=0.075\linewidth,trim=64px 0 64px 0,clip]{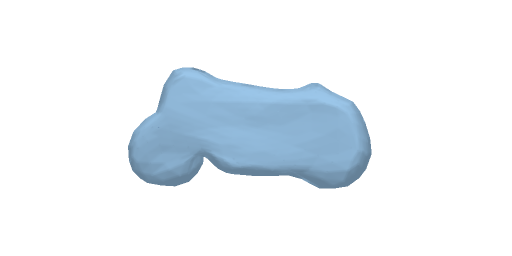} &
        \includegraphics[width=0.075\linewidth,trim=64px 0 64px 0,clip]{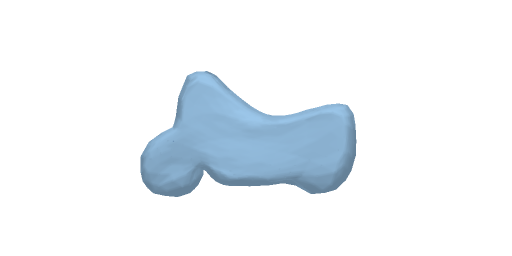} &
        \includegraphics[width=0.075\linewidth,trim=64px 0 64px 0,clip]{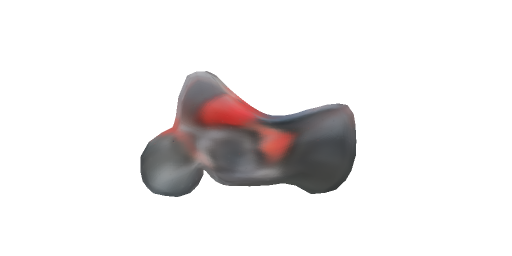} &
        \includegraphics[width=0.075\linewidth,trim=64px 0 64px 0,clip]{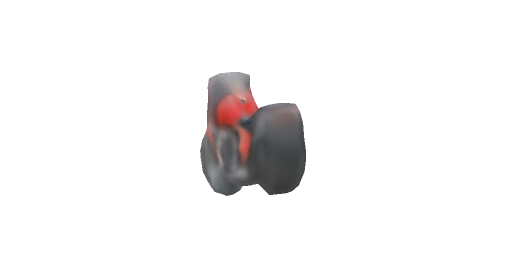} &
        \includegraphics[width=0.075\linewidth,trim=64px 0 64px 0,clip]{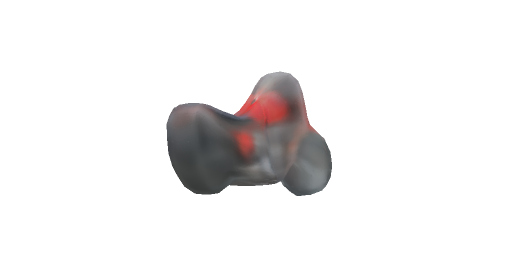} \\
        \midrule
        %
        %
        \rotatebox{90}{\parbox[t]{0.06\linewidth}{\hspace*{\fill}\textbf{\tiny{aeroplane}}\hspace*{\fill}}} &
        \includegraphics[width=0.06\linewidth]{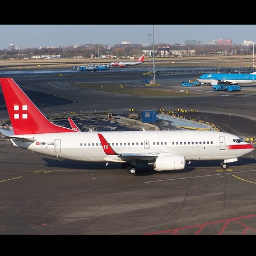} &
        \includegraphics[width=0.075\linewidth,trim=64px 0 64px 0,clip]{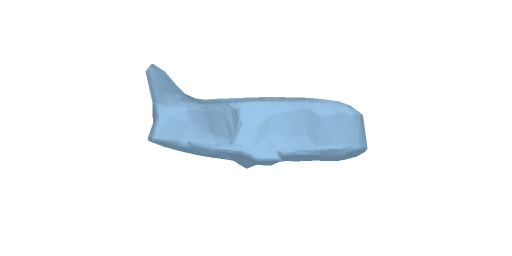} &
        \includegraphics[width=0.075\linewidth,trim=64px 0 64px 0,clip]{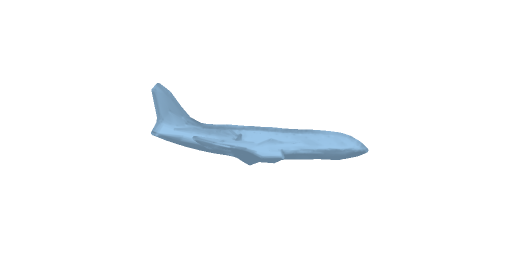} &
        \includegraphics[width=0.075\linewidth,trim=64px 0 64px 0,clip]{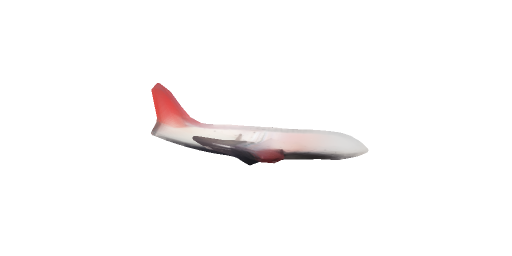} &
        \includegraphics[width=0.075\linewidth,trim=64px 0 64px 0,clip]{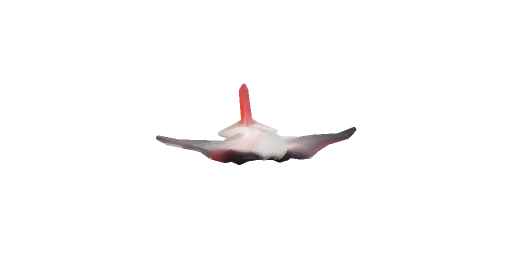} &
        \includegraphics[width=0.075\linewidth,trim=64px 0 64px 0,clip]{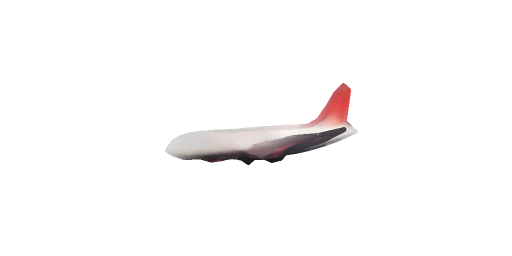} & &
        \rotatebox{90}{\parbox[t]{0.06\linewidth}{\hspace*{\fill}\textbf{\tiny{bicycle}}\hspace*{\fill}}} &
        \includegraphics[width=0.06\linewidth]{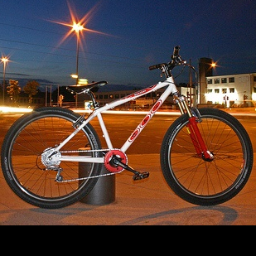} &
        \includegraphics[width=0.075\linewidth,trim=64px 0 64px 0,clip]{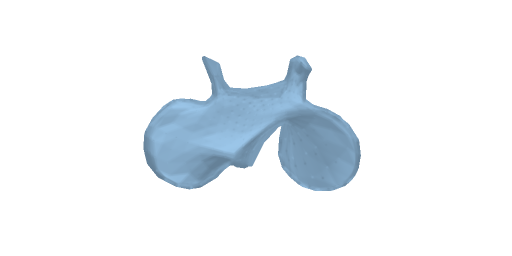} &
        \includegraphics[width=0.075\linewidth,trim=64px 0 64px 0,clip]{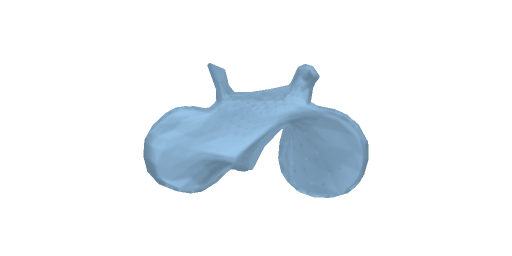} &
        \includegraphics[width=0.075\linewidth,trim=64px 0 64px 0,clip]{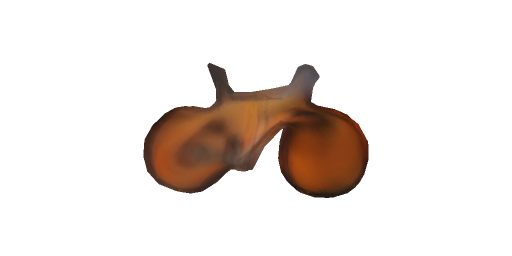} &
        \includegraphics[width=0.075\linewidth,trim=64px 0 64px 0,clip]{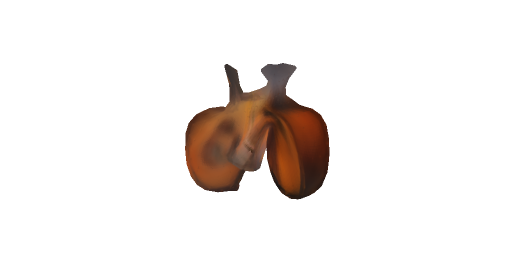} &
        \includegraphics[width=0.075\linewidth,trim=64px 0 64px 0,clip]{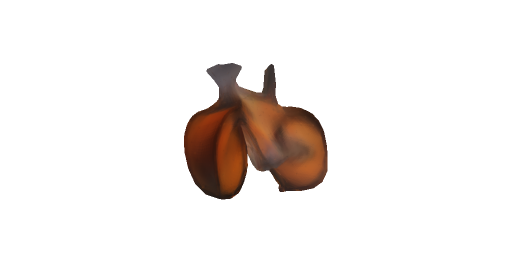} \\
        \rotatebox{90}{\parbox[t]{0.06\linewidth}{\hspace*{\fill}\textbf{\tiny{boat}}\hspace*{\fill}}} &
        \includegraphics[width=0.06\linewidth]{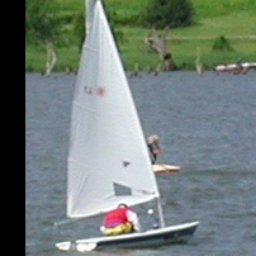} &
        \includegraphics[width=0.075\linewidth,trim=64px 0 64px 0,clip]{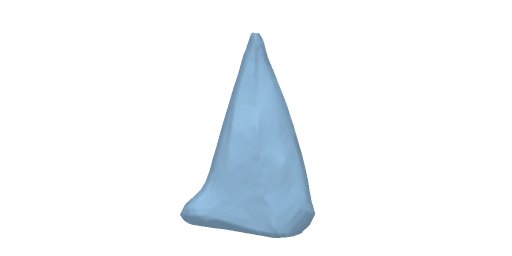} &
        \includegraphics[width=0.075\linewidth,trim=64px 0 64px 0,clip]{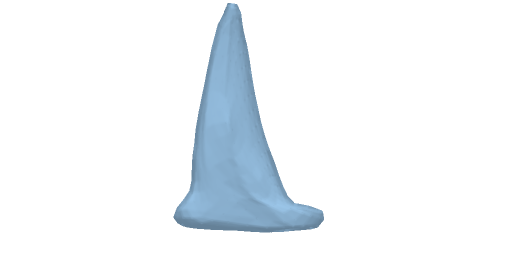} &
        \includegraphics[width=0.075\linewidth,trim=64px 0 64px 0,clip]{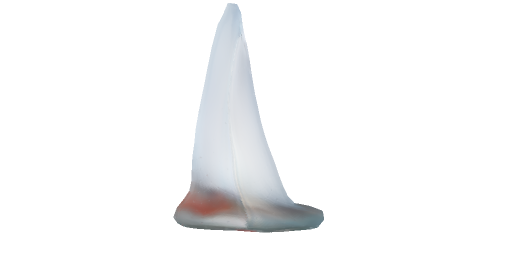} &
        \includegraphics[width=0.075\linewidth,trim=64px 0 64px 0,clip]{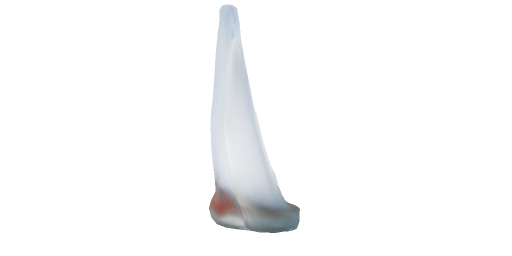} &
        \includegraphics[width=0.075\linewidth,trim=64px 0 64px 0,clip]{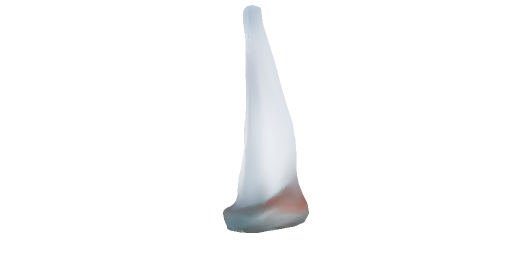} & &
        \rotatebox{90}{\parbox[t]{0.06\linewidth}{\hspace*{\fill}\textbf{\tiny{bottle}}\hspace*{\fill}}} &
        \includegraphics[width=0.06\linewidth]{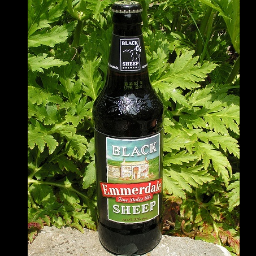} &
        \includegraphics[width=0.075\linewidth,trim=64px 0 64px 0,clip]{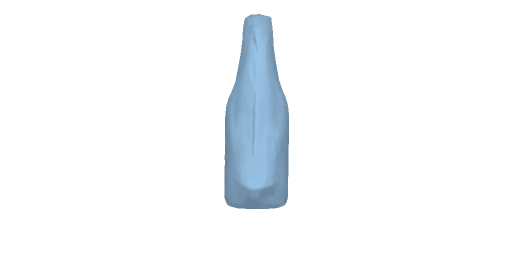} &
        \includegraphics[width=0.075\linewidth,trim=64px 0 64px 0,clip]{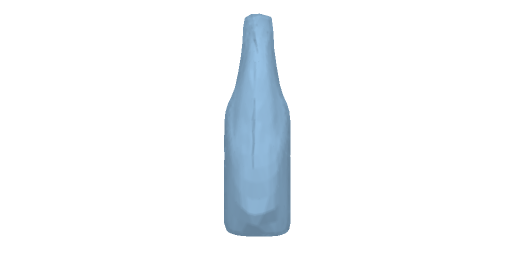} &
        \includegraphics[width=0.075\linewidth,trim=64px 0 64px 0,clip]{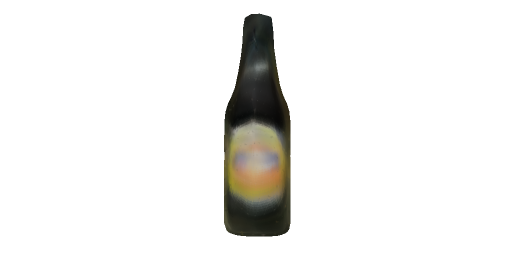} &
        \includegraphics[width=0.075\linewidth,trim=64px 0 64px 0,clip]{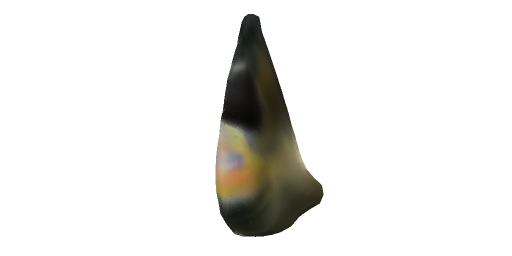} &
        \includegraphics[width=0.075\linewidth,trim=64px 0 64px 0,clip]{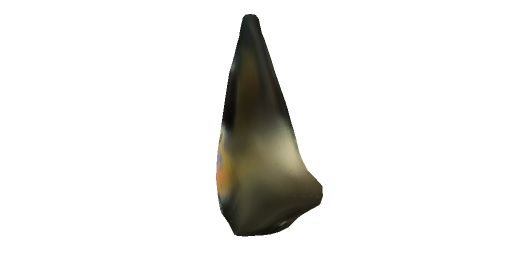} \\
        \rotatebox{90}{\parbox[t]{0.06\linewidth}{\hspace*{\fill}\textbf{\tiny{bus}}\hspace*{\fill}}} &
        \includegraphics[width=0.06\linewidth]{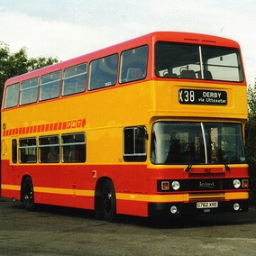} &
        \includegraphics[width=0.075\linewidth,trim=64px 0 64px 0,clip]{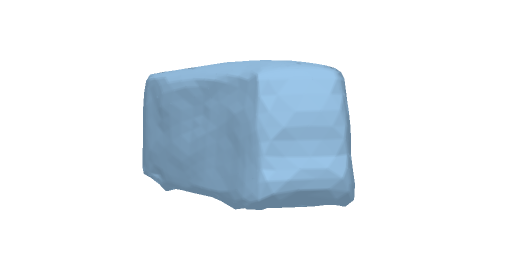} &
        \includegraphics[width=0.075\linewidth,trim=64px 0 64px 0,clip]{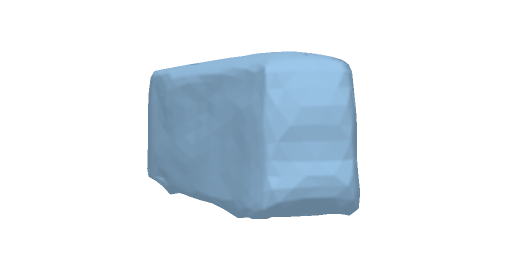} &
        \includegraphics[width=0.075\linewidth,trim=64px 0 64px 0,clip]{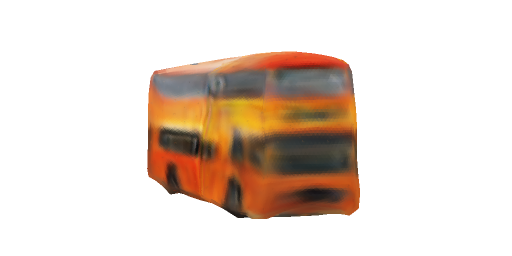} &
        \includegraphics[width=0.075\linewidth,trim=64px 0 64px 0,clip]{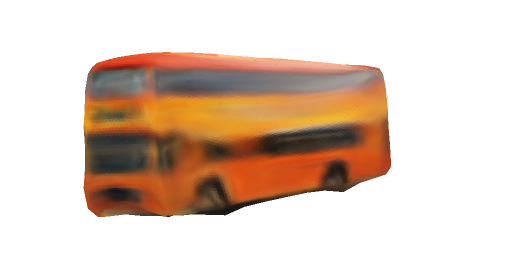} &
        \includegraphics[width=0.075\linewidth,trim=48px -12px 48px -12px,clip]{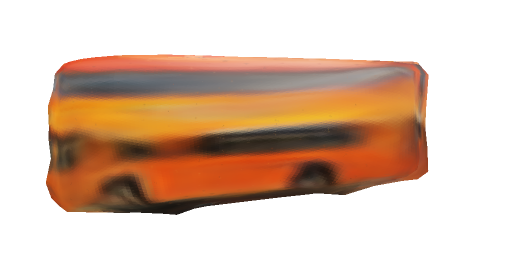} & &
        \rotatebox{90}{\parbox[t]{0.06\linewidth}{\hspace*{\fill}\textbf{\tiny{car}}\hspace*{\fill}}} &
        \includegraphics[width=0.06\linewidth]{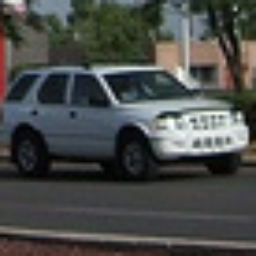} &
        \includegraphics[width=0.075\linewidth,trim=64px 0 64px 0,clip]{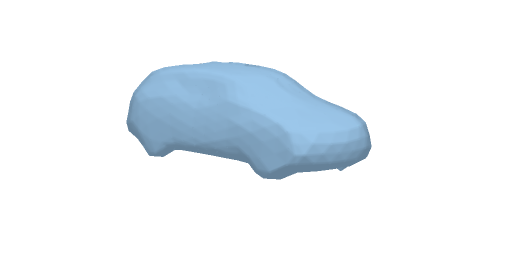} &
        \includegraphics[width=0.075\linewidth,trim=64px 0 64px 0,clip]{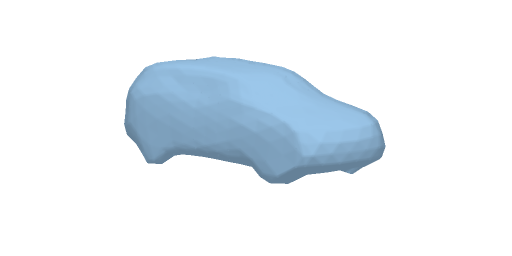} &
        \includegraphics[width=0.075\linewidth,trim=64px 0 64px 0,clip]{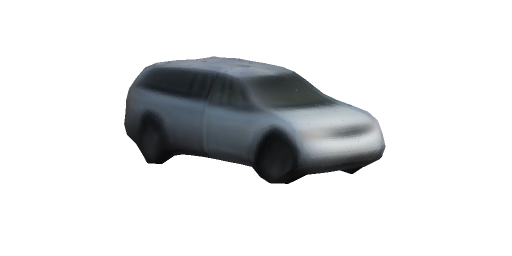} &
        \includegraphics[width=0.075\linewidth,trim=64px 0 64px 0,clip]{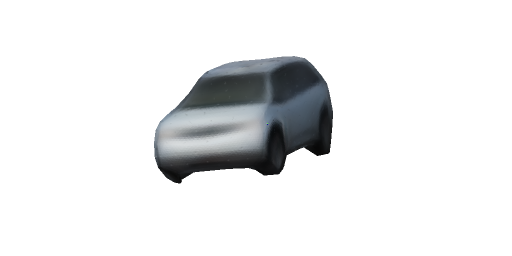} &
        \includegraphics[width=0.075\linewidth,trim=64px 0 64px 0,clip]{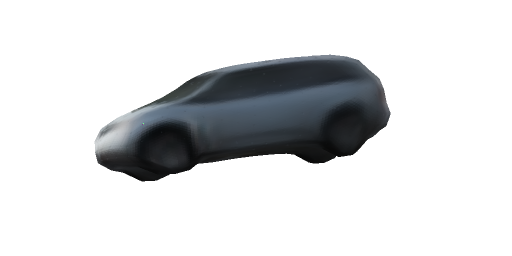} \\
        \rotatebox{90}{\parbox[t]{0.06\linewidth}{\hspace*{\fill}\textbf{\tiny{chair}}\hspace*{\fill}}} &
        \includegraphics[width=0.06\linewidth]{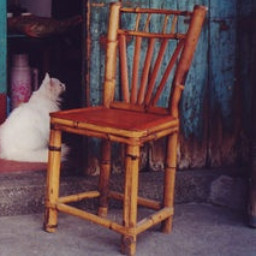} &
        \includegraphics[width=0.075\linewidth,trim=64px 0 64px 0,clip]{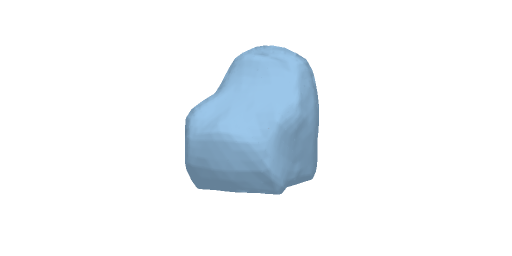} &
        \includegraphics[width=0.075\linewidth,trim=64px 0 64px 0,clip]{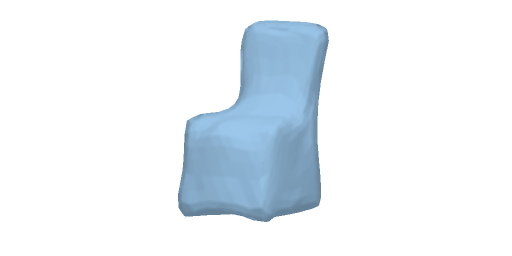} &
        \includegraphics[width=0.075\linewidth,trim=64px 0 64px 0,clip]{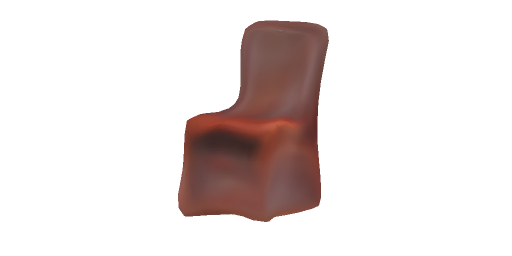} &
        \includegraphics[width=0.075\linewidth,trim=64px 0 64px 0,clip]{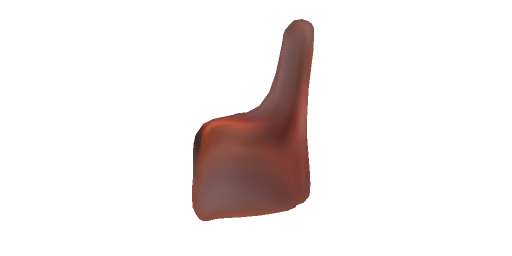} &
        \includegraphics[width=0.075\linewidth,trim=64px 0 64px 0,clip]{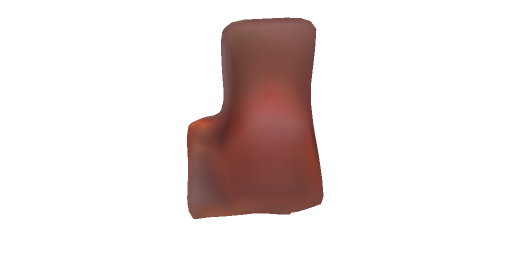} & &
        \rotatebox{90}{\parbox[t]{0.06\linewidth}{\hspace*{\fill}\textbf{\tiny{table}}\hspace*{\fill}}} &
        \includegraphics[width=0.06\linewidth]{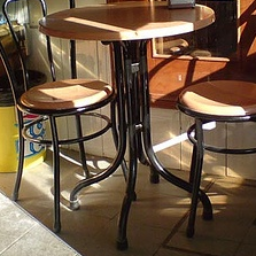} &
        \includegraphics[width=0.075\linewidth,trim=64px 0 64px 0,clip]{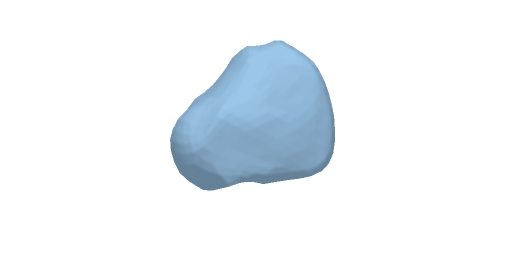} &
        \includegraphics[width=0.075\linewidth,trim=64px 0 64px 0,clip]{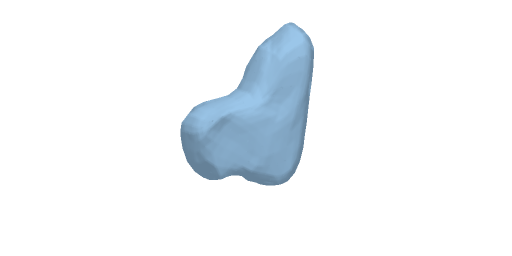} &
        \includegraphics[width=0.075\linewidth,trim=64px 0 64px 0,clip]{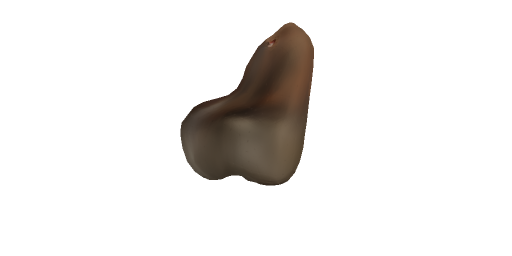} &
        \includegraphics[width=0.075\linewidth,trim=64px 0 64px 0,clip]{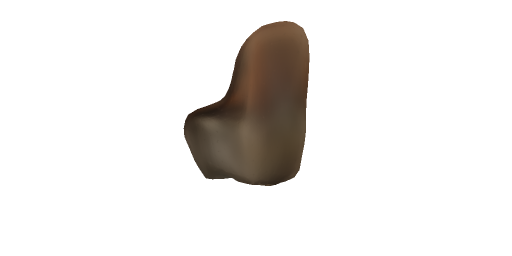} &
        \includegraphics[width=0.075\linewidth,trim=64px 0 64px 0,clip]{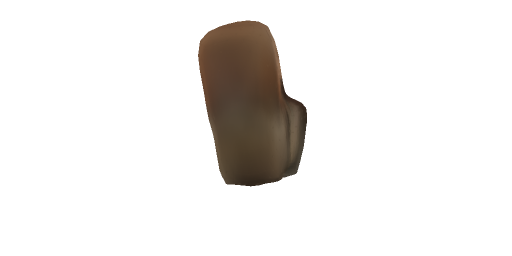} \\
        \rotatebox{90}{\parbox[t]{0.06\linewidth}{\hspace*{\fill}\textbf{\tiny{motorbike}}\hspace*{\fill}}} &
        \includegraphics[width=0.06\linewidth]{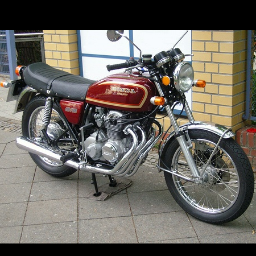} &
        \includegraphics[width=0.075\linewidth,trim=64px 0 64px 0,clip]{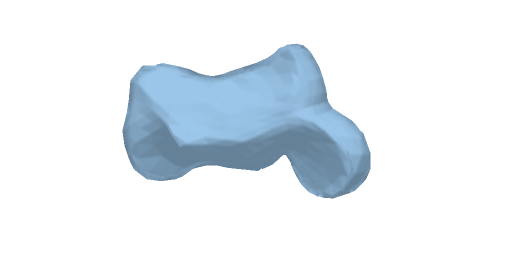} &
        \includegraphics[width=0.075\linewidth,trim=64px 0 64px 0,clip]{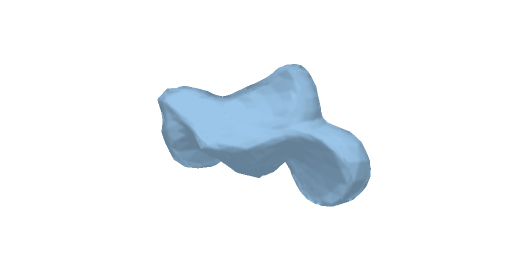} &
        \includegraphics[width=0.075\linewidth,trim=64px 0 64px 0,clip]{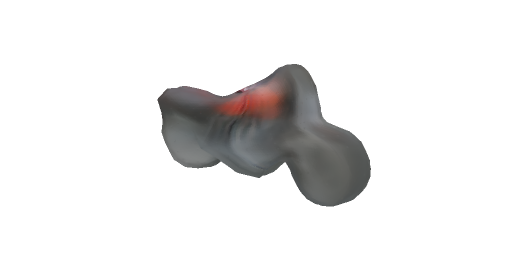} &
        \includegraphics[width=0.075\linewidth,trim=64px 0 64px 0,clip]{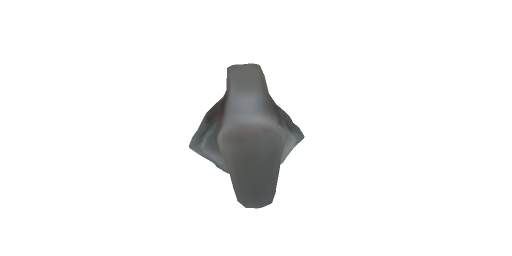} &
        \includegraphics[width=0.075\linewidth,trim=64px 0 64px 0,clip]{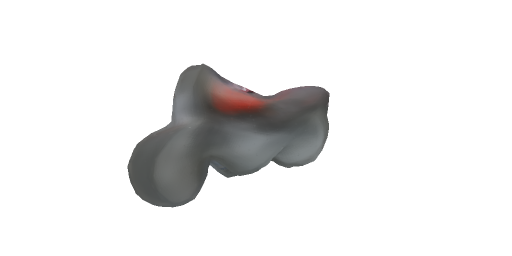} & &
        \rotatebox{90}{\parbox[t]{0.06\linewidth}{\hspace*{\fill}\textbf{\tiny{sofa}}\hspace*{\fill}}} &
        \includegraphics[width=0.06\linewidth]{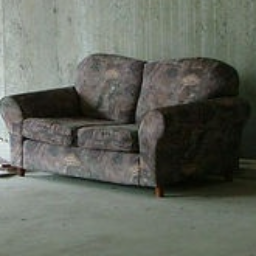} &
        \includegraphics[width=0.075\linewidth,trim=64px 0 64px 0,clip]{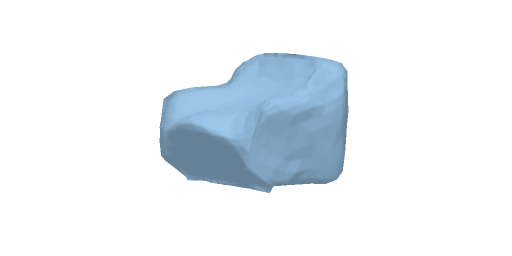} &
        \includegraphics[width=0.075\linewidth,trim=64px 0 64px 0,clip]{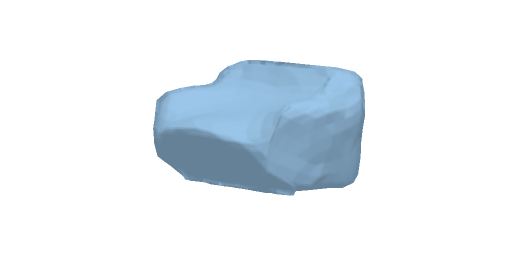} &
        \includegraphics[width=0.075\linewidth,trim=64px 0 64px 0,clip]{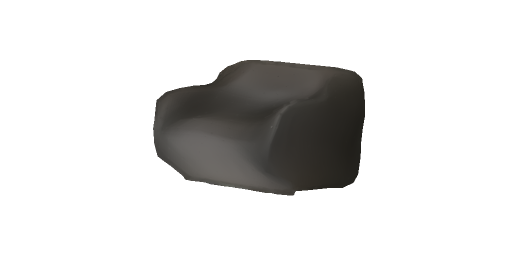} &
        \includegraphics[width=0.075\linewidth,trim=64px 0 64px 0,clip]{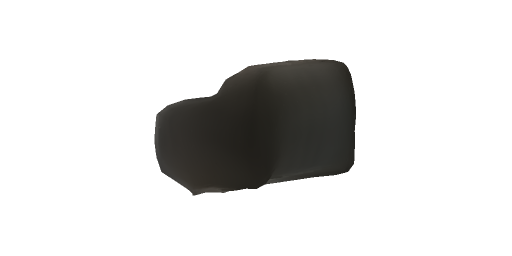} &
        \includegraphics[width=0.075\linewidth,trim=64px 0 64px 0,clip]{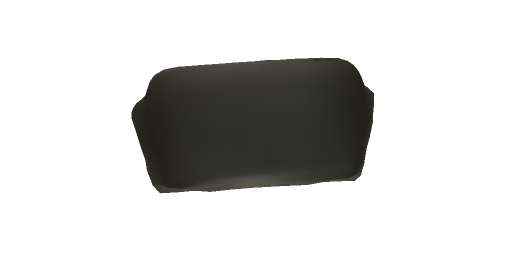} \\
        \rotatebox{90}{\parbox[t]{0.06\linewidth}{\hspace*{\fill}\textbf{\tiny{train}}\hspace*{\fill}}} &
        \includegraphics[width=0.06\linewidth]{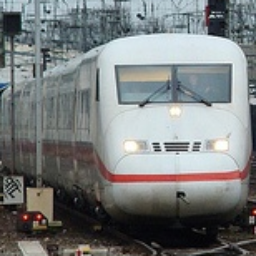} &
        \includegraphics[width=0.075\linewidth,trim=220px 0 220px 0,clip]{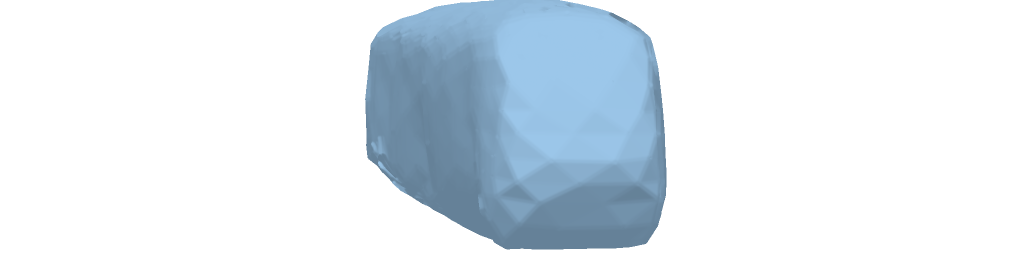} &
        \includegraphics[width=0.075\linewidth,trim=256px 0 256px 0,clip]{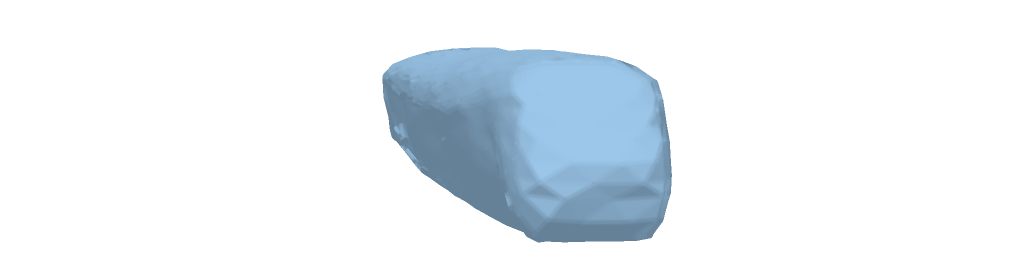} &
        \includegraphics[width=0.075\linewidth,trim=256px 0px 256px 0px,clip]{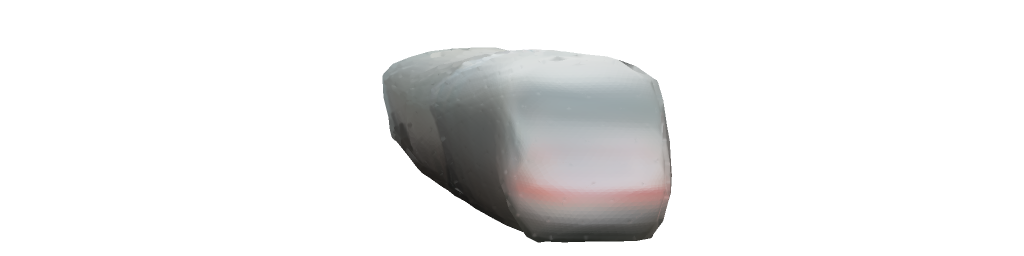} &
        \includegraphics[width=0.075\linewidth,trim=256px 0px 256px 0px,clip]{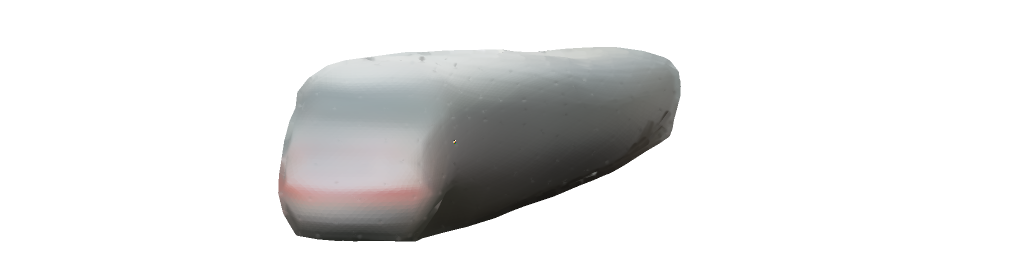} &
        \includegraphics[width=0.075\linewidth,trim=200px 0px 200px 0px,clip]{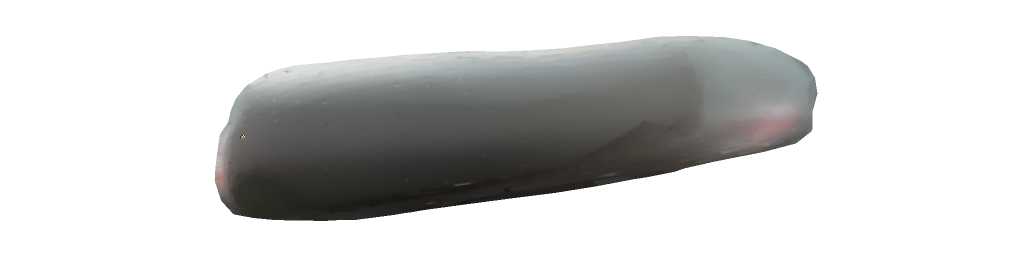} & &
        \rotatebox{90}{\parbox[t]{0.06\linewidth}{\hspace*{\fill}\textbf{\tiny{monitor}}\hspace*{\fill}}} &
        \includegraphics[width=0.06\linewidth]{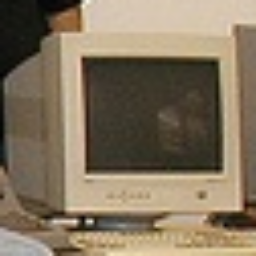} &
        \includegraphics[width=0.075\linewidth,trim=64px 0 64px 0,clip]{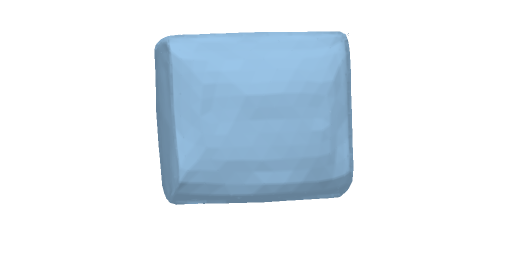} &
        \includegraphics[width=0.075\linewidth,trim=64px 0 64px 0,clip]{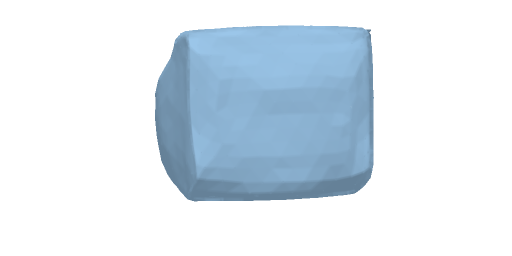} &
        \includegraphics[width=0.075\linewidth,trim=64px 0 64px 0,clip]{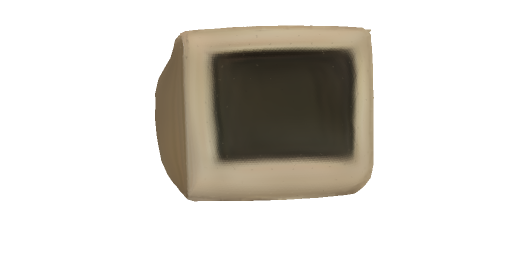} &
        \includegraphics[width=0.075\linewidth,trim=64px 0 64px 0,clip]{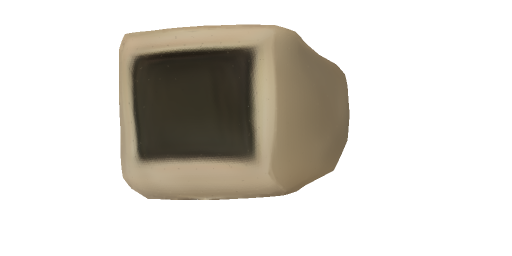} &
        \includegraphics[width=0.075\linewidth,trim=64px 0 64px 0,clip]{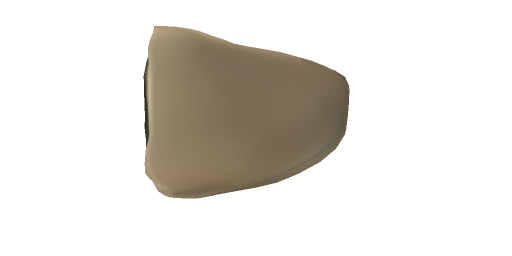} \\
    \end{tabular}
    \caption{Qualitative results on different settings:
    CUB~\cite{WahCUB_200_2011} (birds) and Pascal3D+~\cite{xiang2014beyond} (aeroplane and car, $4$ automotive classes, all $12$ classes).
    We show the input image $I$, the output $M$ of the unsupervised shape selection module, the predicted shape $\hat{M}$ and the predicted textured shape ${\hat{M}} + {\hat{I}_\text{tex}}$ under several 3D rotations over the vertical axis of the predicted pose $\hat{\pi}$.}
    \vspace{-1em}
    \label{fig:qualitative_results}
\end{figure*}

\subsubsec{Dynamic mesh subdivision}
We evaluate the contribution of the dynamic mesh subdivision during the training process using the four automotive classes.
We compare three different settings of the 3D mesh connectivity, in terms of icosphere subdivision level:
(i) level set to 3, (ii) level set to 4, and (iii) dynamic subdivision starting from level 3 and going up to level 4.
Results are reported in Table~\ref{tab:pascal_subdivision_single_col}.
As shown,  the method can converge to good results even using a fixed subdivision level. However, a higher level does not always lead to better scores, as in the case of fixed subdivision level 4.
On the contrary, increasing the subdivision level during training leads to higher results in terms of both mask IoU and texture metrics. 
Indeed, dynamic subdivision allows to 
take advantage of low subdivision levels during the initial training phase -- optimizing the shape smoothness in a faster and easier way -- 
and at the same time leveraging the higher number of faces of high subdivision levels in the second part of the training -- improving the finer details and the quality of the texture.

\section{Conclusion}
In this paper, we show how the 3D mesh reconstruction of objects can be learned jointly on multiple classes using only foreground masks and coarse camera poses as supervision.
The proposed approach discerns between different object categories and learns meaningful category-level meanshapes, which were initialized as spheres, in an unsupervised manner.
In addition, a novel approach to predict the instance-specific deformation at vertex level is presented.
The network produces smooth deformations and is independent of the number of the mesh vertices, allowing the dynamic subdivision of the mesh during training.
Quantitative and qualitative results on two public datasets show the effectiveness of the proposed method.

\subsubsec{Acknowledgements}
This research was supported by MIUR PRIN project ``PREVUE: PRediction of activities and Events by Vision in an Urban Environment", grant ID E94I19000650001.

{\small
\bibliographystyle{ieee_fullname}
\bibliography{egbib}
}

\section*{Supplementary Material}

\appendix

In this supplementary material, 
we report architectural details of the proposed method in Section~\ref{sec:arch}, followed by an analysis of the computational performance in Section~\ref{sec:perf}.
In section~\ref{sec:ablation}, we present additional ablation studies, including the impact of pre-training and foreground mask quality; an analysis of the meanshape learning process; the use of a different number of meanshapes on CUB; a study on the unsupervised shape selection module, in terms of classification accuracy and average meanshape weight.
Finally, additional qualitative results and failure cases are reported in Section~\ref{sec:qualitative}.

\section{Architectural details}\label{sec:arch}
In this section, we firstly describe the architectural details of our method.
Then, we report the weights used to balance the losses
during the training process.

\subsection{Network}
Here, we report the implementation details of each module of the proposed framework.

\subsubsec{Feature extraction}
We use ResNet-18~\cite{he2016deep} as visual encoder, 
replacing the classification layer with an additional convolutional layer with kernel size $k=4$, stride $s=2$, and $256$ filters.
Taking as input an RGB image $I \in \mathbb{R}^{3 \times 256 \times 256}$, the encoder outputs a feature map $f_{\text{tex}} \in \mathbb{R}^{256 \times 4 \times 4}$.
These features are then flattened and given as input to a 256-d fully connected layer with batch normalization and a leaky ReLU activation function, obtaining a 256-d feature vector $f_{\text{shape}}$.
The visual encoder is pre-trained on ImageNet~\cite{deng2009imagenet}. We investigate the impact of pre-training on the unsupervised shape selection in Section~\ref{subsec:pre-training}.

\subsubsec{Unsupervised shape selection} 
The unsupervised shape selection module is a network that smoothly approximates the argmax function over the $N$ meanshapes.
It is composed of two fully-connected layers: (i) a $64$-d layer with batch normalization and leaky ReLU, (ii) a $N$-d layer followed by a softmax activation function that outputs the $N$ weighting scores.
The input of the module are the 
features $f_{\text{shape}}$.

\subsubsec{Vertex deformation} 
Inspired by the work of Park \etal~\cite{park2019deepsdf}, the vertex deformation network is composed of four 512-d fully connected layers with weight normalization, random dropout of $0.2$, and the ReLU activation function.
An additional $3$-d fully connected layer with a tanh activation function outputs the displacement 
$\Delta v_j = (\Delta x, \Delta y, \Delta z)$ 
of the vertex $v_j$, which is given as input along with the features $f_{\text{shape}}$ and the weighting scores of the previous module.
The input features (\ie vertex location, $f_{\text{shape}}$, and weighting scores) are also concatenated to the output of the second layer, before applying the third one.

\begin{table}[t]
    \begin{center}
    \renewcommand{\arraystretch}{1.1}
    \resizebox{1\linewidth}{!}{
    \begin{tabular}{l|cccc|cc|ccc}
        \textbf{Dataset} & $\lambda_1$ & $\lambda_2$ & $\lambda_3$ & $\lambda_4$ & $\lambda_5$ & $\lambda_6$ & $\lambda_7$ & $\lambda_8$ & $\lambda_9$ \\
        \midrule
        Pascal3D+ & $100.0$ & $6.0$ & $1.8$ & $0.05$ & $20.0$ & $2.0$ & $0.03$ & $0.05$ & $0.8$ \\
        CUB & $20.0$ & $1.2$ & $0.18$ & $0.005$ & $2.0$ & $0.1$ & $0.12$ & $0.02$ & $3.2$ \\
        \bottomrule
    \end{tabular}
    }
    \end{center}
    \vspace{-5pt}
    \caption{Loss weights on Pascal3D+~\cite{xiang2014beyond} and CUB~\cite{WahCUB_200_2011}.}
    \label{tab:loss_weights}
\end{table}

\begin{table}[t]
    \begin{center}
    \renewcommand{\arraystretch}{1.1}
    \resizebox{1\linewidth}{!}{
    \begin{tabular}{l|ccc}
        \textbf{Method} & \textbf{Params} (M) & \textbf{Memory} (GB) & \textbf{Inference} (ms) \\
        \midrule
        CMR~\cite{kanazawa2018learning} & $84.25$ & $2.28$ & $3.56 \pm 0.14$ \\
        U-CMR~\cite{goel2020shape} & $19.89$ & $3.38$ & $5.10 \pm 2.91$ \\
        \textbf{Ours} & $20.10$ & $3.74$ & $4.43\pm0.19$ \\
        \bottomrule
    \end{tabular}
    }
    \end{center}
    \vspace{-5pt}
    \caption{Performance analysis of our multi-category approach against open-sourced single-category competitors.}
    \label{tab:performance}
\end{table}

\subsubsec{3D pose regression}
The prediction of the object viewpoint is tackled as a regression problem using two fully connected layers: 
(i) a $64$-d layer with batch normalization, random dropout of $0.5$, and leaky ReLU, (ii) a $7$-d layer that outputs the object pose $\hat{\pi}=(\hat{s},\hat{t},\hat{q}) \in (\mathbb{R}^{1}, \mathbb{R}^{2}, \mathbb{R}^{4})$.
The input of the module are the features $f_{\text{shape}}$.

\begin{table*}[t]
    \begin{center}
    \renewcommand{\arraystretch}{1.1}
    \resizebox{0.9\linewidth}{!}{
    \begin{tabular}{c|c|c|c|cc|ccc}
        \multirow{2}{*}{\textbf{Training classes}} & \textbf{Segmentation} & \textbf{Number of} & \multirow{2}{*}{\textbf{3D IoU} $\uparrow$} & \multicolumn{2}{c|}{\textbf{Mask IoU} $\uparrow$} & \multicolumn{3}{c}{\textbf{Texture metrics}} \\
        & \textbf{Method} & \textbf{meanshapes} & & \textbf{Pred cam} & \textbf{GT cam} & \textbf{SSIM} $\uparrow$ &  \textbf{L1} $\downarrow$ & \textbf{FID} $\downarrow$ \\
        \midrule
        aeroplane, car & Mask R-CNN & 2 & $\mathbf{0.556}$ & $0.648$ & $0.699$ & $\mathbf{0.739}$ & $0.064$ & $350.12$ \\
        aeroplane, car & PointRend & 2 & $0.552$ & $\mathbf{0.671}$ & $\mathbf{0.702}$ & $0.737$ & $\mathbf{0.062}$ & $\mathbf{344.80}$ \\
        \midrule
        bicycle, bus, car, motorbike & Mask R-CNN & 4 & $0.530$ & $0.677$ & $0.756$ & $0.605$ & $0.098$ & $390.55$ \\
        bicycle, bus, car, motorbike & PointRend & 4 & $\mathbf{0.543}$ & $\mathbf{0.711}$ & $\mathbf{0.759}$ & $\mathbf{0.607}$ & $\mathbf{0.094}$ & $\mathbf{380.15}$ \\
        \hline
    \end{tabular}
    }
    \end{center}
    \vspace{-5pt}
    \caption{Evaluation on Pascal3D+~\cite{xiang2014beyond} using segmentation masks obtained with Mask R-CNN~\cite{he2017mask} or PointRend~\cite{kirillov2020pointrend}.}
    \label{tab:pascal_segmentation_masks}
    \vspace{-0.5em}
\end{table*}

\subsubsec{Texture prediction} 
Inspired by the decoder of the SPADE architecture proposed by Park \etal~\cite{park2019semantic}, our texture decoder is composed of 6 upsampling steps with bilinear interpolation, in order to output a texture image $I_{\text{tex}} \in \mathbb{R}^{3 \times 256 \times 256}$. 
Differently from the original implementation, we use only the convolutional layers with skip connections and leaky ReLU activation functions. 
We test different types of normalization (\eg batch, instance), but we obtain the best results without it.
The decoder takes as input the features $f_{\text{tex}}$ and the output is finally passed through a sigmoid activation function in order to obtain valid RGB color values.

\subsection{Loss weights}
In the following, we reintroduce the losses used during training in order to show their weighting parameters, whose values are reported in Table~\ref{tab:loss_weights}.
We select different weights for each dataset,
exploiting their validation set.

For the shape prediction, the loss is defined by:
\begin{equation}
    \mathcal{L}_\text{shape} = \lambda_1\mathcal{L}_\text{mask} + \lambda_2\mathcal{L}_\text{smooth}^{\hat{M}} + \lambda_3\mathcal{L}_\text{smooth}^{\Delta V} + \lambda_4\mathcal{L}_\text{def}
    \label{eq:shape}
\end{equation}
where the smoothness prior is applied to both the vertices of deformed shape $\hat{M}$ and the predicted deformations $\Delta V$.
For the pose regression, the loss is defined as:
\begin{equation}
    \mathcal{L}_\text{cam} = \lambda_5\mathcal{L}_\text{pose} + \lambda_6\mathcal{L}_\text{pose\_reg}
\end{equation}
while the texture prediction loss is represented by:
\begin{equation}
    \mathcal{L}_\text{tex} = \lambda_7\mathcal{L}_\text{color} + \lambda_8\mathcal{L}_\text{style} + \lambda_9\mathcal{L}_\text{percept}
\end{equation}

\section{Computational Performance}\label{sec:perf}
In this section, we assess the computational requirements of our method and some open-sourced competitors. 
Compared to previous category-specific methods, our approach does not require an initial shape classifier and the training on N independent models, thus being faster and requiring less memory during inference.
Indeed, our multi-category method has comparable network size, memory usage, and inference time with respect to the single-category competitors, as reported in Table~\ref{tab:performance}.
Their evaluation is conducted on a workstation with an \textit{Intel Core} i7-7700K and a \textit{Nvidia GeForce} GTX 1080 Ti.

\begin{figure}[t]
    \centering
    \setlength{\tabcolsep}{1.5pt}
    \renewcommand{\arraystretch}{1}
    \begin{tabular}{cccccc}
    & \footnotesize{$+60\degree$} & \footnotesize{$+120\degree$} & \footnotesize{$+180\degree$} & \footnotesize{$+240\degree$} & \footnotesize{$+300\degree$} \\
    \includegraphics[width=0.153\linewidth,trim=10pt 60pt 10pt 40pt,clip]{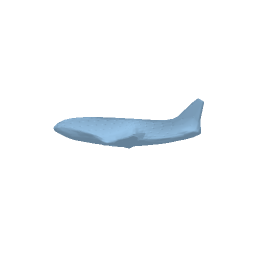} &
    \includegraphics[width=0.153\linewidth,trim=10pt 60pt 10pt 40pt,clip]{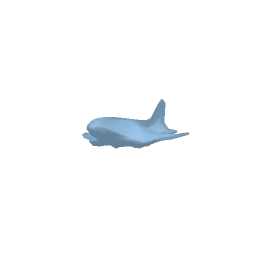} &
    \includegraphics[width=0.153\linewidth,trim=10pt 60pt 10pt 40pt,clip]{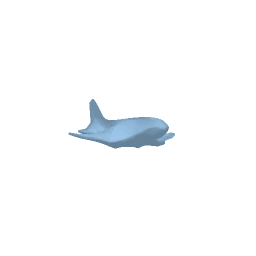} &
    \includegraphics[width=0.153\linewidth,trim=10pt 60pt 10pt 40pt,clip]{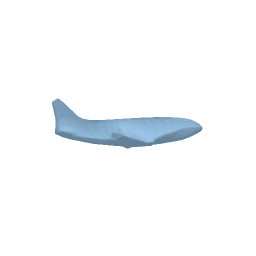} &
    \includegraphics[width=0.153\linewidth,trim=10pt 60pt 10pt 40pt,clip]{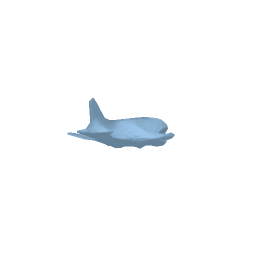} &
    \includegraphics[width=0.153\linewidth,trim=10pt 60pt 10pt 40pt,clip]{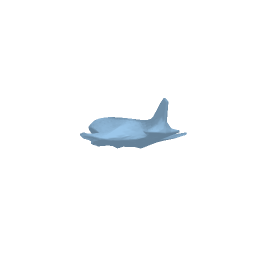} \\
    \includegraphics[width=0.153\linewidth,trim=10pt 60pt 10pt 40pt,clip]{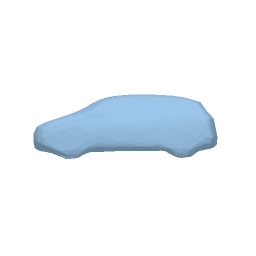} &
    \includegraphics[width=0.153\linewidth,trim=10pt 60pt 10pt 40pt,clip]{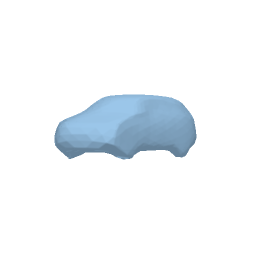} &
    \includegraphics[width=0.153\linewidth,trim=10pt 60pt 10pt 40pt,clip]{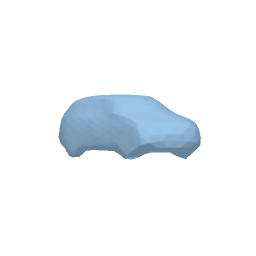} &
    \includegraphics[width=0.153\linewidth,trim=10pt 60pt 10pt 40pt,clip]{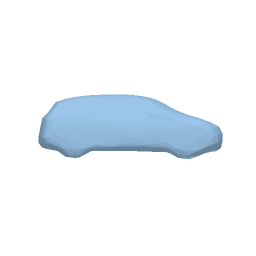} &
    \includegraphics[width=0.153\linewidth,trim=10pt 60pt 10pt 40pt,clip]{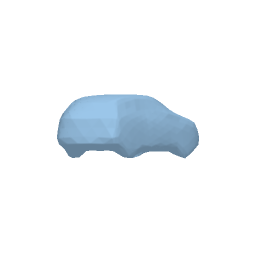} &
    \includegraphics[width=0.153\linewidth,trim=10pt 60pt 10pt 40pt,clip]{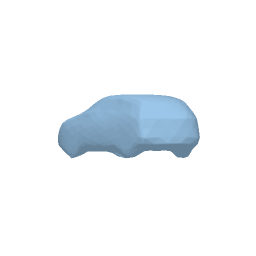} \\
    \midrule
    \includegraphics[width=0.153\linewidth,trim=10pt 60pt 10pt 40pt,clip]{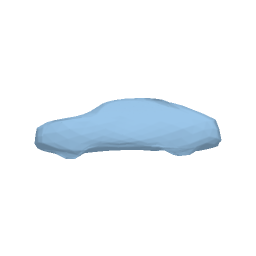} &
    \includegraphics[width=0.153\linewidth,trim=10pt 60pt 10pt 40pt,clip]{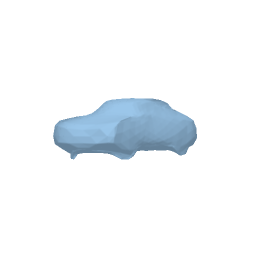} &
    \includegraphics[width=0.153\linewidth,trim=10pt 60pt 10pt 40pt,clip]{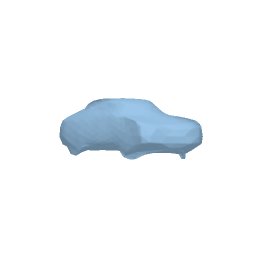} &
    \includegraphics[width=0.153\linewidth,trim=10pt 60pt 10pt 40pt,clip]{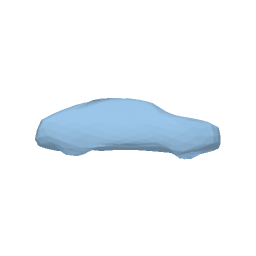} &
    \includegraphics[width=0.153\linewidth,trim=10pt 60pt 10pt 40pt,clip]{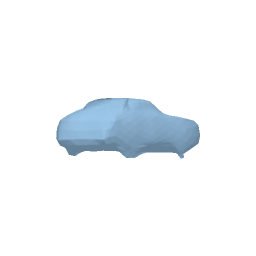} &
    \includegraphics[width=0.153\linewidth,trim=10pt 60pt 10pt 40pt,clip]{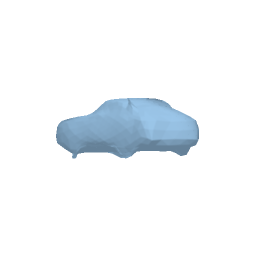} \\
    \includegraphics[width=0.153\linewidth,trim=10pt 60pt 10pt 40pt,clip]{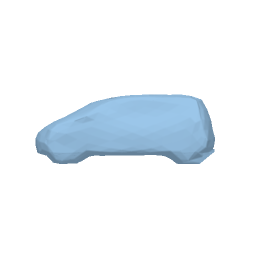} &
    \includegraphics[width=0.153\linewidth,trim=10pt 60pt 10pt 40pt,clip]{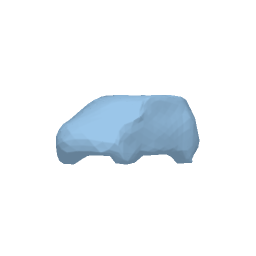} &
    \includegraphics[width=0.153\linewidth,trim=10pt 60pt 10pt 40pt,clip]{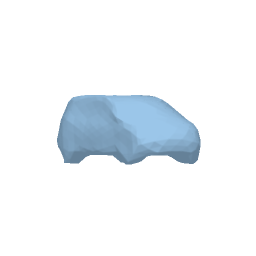} &
    \includegraphics[width=0.153\linewidth,trim=10pt 60pt 10pt 40pt,clip]{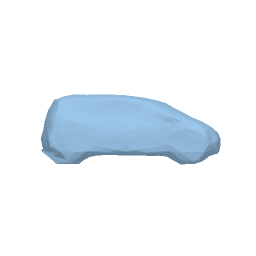} &
    \includegraphics[width=0.153\linewidth,trim=10pt 60pt 10pt 40pt,clip]{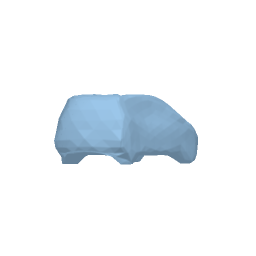} &
    \includegraphics[width=0.153\linewidth,trim=10pt 60pt 10pt 40pt,clip]{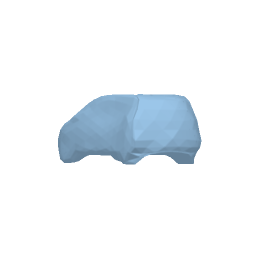} \\
    \includegraphics[width=0.153\linewidth,trim=10pt 60pt 10pt 40pt,clip]{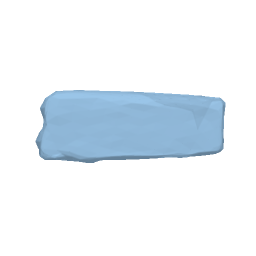} &
    \includegraphics[width=0.153\linewidth,trim=10pt 60pt 10pt 40pt,clip]{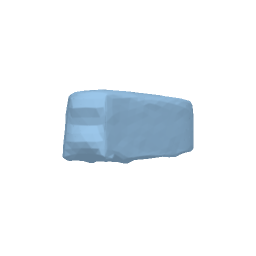} &
    \includegraphics[width=0.153\linewidth,trim=10pt 60pt 10pt 40pt,clip]{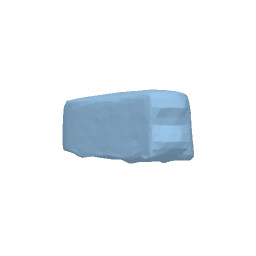} &
    \includegraphics[width=0.153\linewidth,trim=10pt 60pt 10pt 40pt,clip]{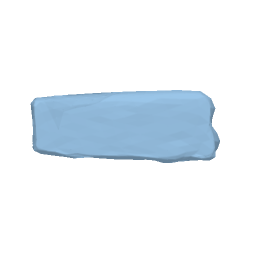} &
    \includegraphics[width=0.153\linewidth,trim=10pt 60pt 10pt 40pt,clip]{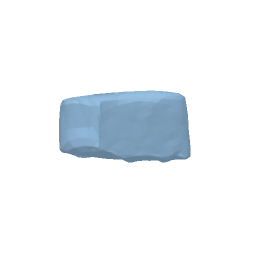} &
    \includegraphics[width=0.153\linewidth,trim=10pt 60pt 10pt 40pt,clip]{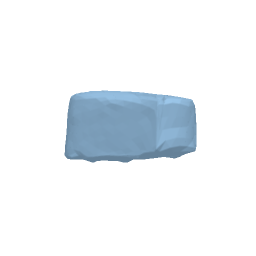} \\
    \includegraphics[width=0.153\linewidth,trim=10pt 40pt 10pt 40pt,clip]{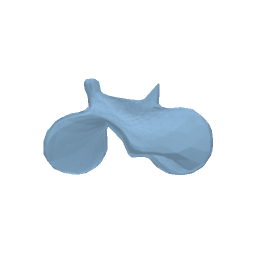} &
    \includegraphics[width=0.153\linewidth,trim=10pt 40pt 10pt 40pt,clip]{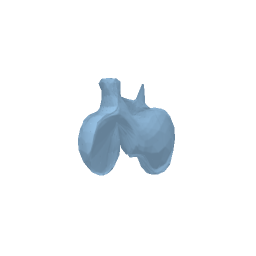} &
    \includegraphics[width=0.153\linewidth,trim=10pt 40pt 10pt 40pt,clip]{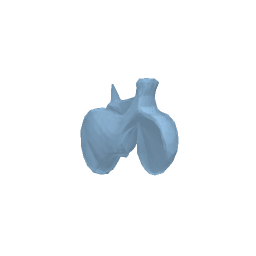} &
    \includegraphics[width=0.153\linewidth,trim=10pt 40pt 10pt 40pt,clip]{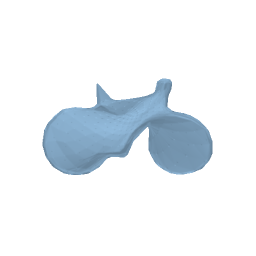} &
    \includegraphics[width=0.153\linewidth,trim=10pt 40pt 10pt 40pt,clip]{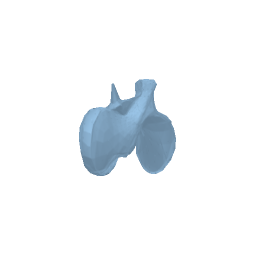} &
    \includegraphics[width=0.153\linewidth,trim=10pt 40pt 10pt 40pt,clip]{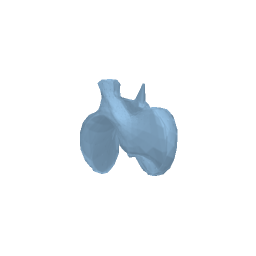}
    \end{tabular}
    \vspace{-5pt}
    \caption{Meanshapes learned by our method trained on aeroplanes and cars, and on 4 automotive classes of Pascal3D+~\cite{xiang2014beyond} without encoder pre-training on ImageNet.}
    \label{fig:meanshapes_4macroclasses_nopretrain}
    \vspace{-1em}
\end{figure}

\begin{figure}[t]
    \centering
    \setlength{\tabcolsep}{1.5pt}
    \renewcommand{\arraystretch}{1}
    \begin{tabular}{cccccc}
    & \footnotesize{$+60\degree$} & \footnotesize{$+120\degree$} & \footnotesize{$+180\degree$} & \footnotesize{$+240\degree$} & \footnotesize{$+300\degree$} \\
    \includegraphics[width=0.153\linewidth,trim=10pt 70pt 10pt 40pt,clip]{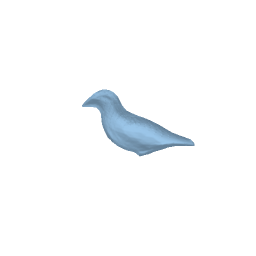} &
    \includegraphics[width=0.153\linewidth,trim=10pt 70pt 10pt 40pt,clip]{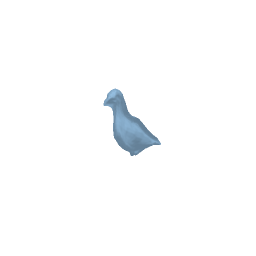} &
    \includegraphics[width=0.153\linewidth,trim=10pt 70pt 10pt 40pt,clip]{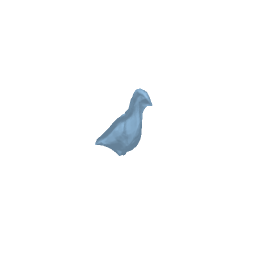} &
    \includegraphics[width=0.153\linewidth,trim=10pt 70pt 10pt 40pt,clip]{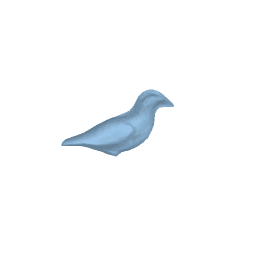} &
    \includegraphics[width=0.153\linewidth,trim=10pt 70pt 10pt 40pt,clip]{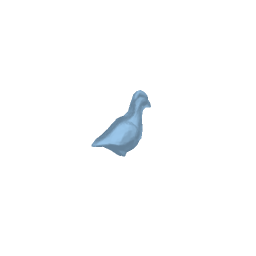} &
    \includegraphics[width=0.153\linewidth,trim=10pt 70pt 10pt 40pt,clip]{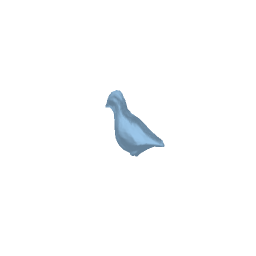} \\
    \includegraphics[width=0.153\linewidth,trim=10pt 40pt 10pt 40pt,clip]{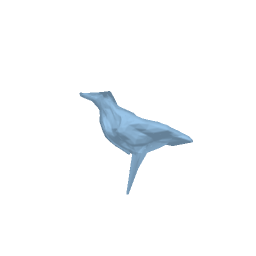} &
    \includegraphics[width=0.153\linewidth,trim=10pt 40pt 10pt 40pt,clip]{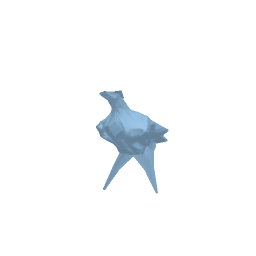} &
    \includegraphics[width=0.153\linewidth,trim=10pt 40pt 10pt 40pt,clip]{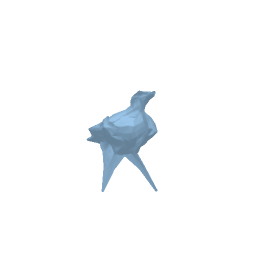} &
    \includegraphics[width=0.153\linewidth,trim=10pt 40pt 10pt 40pt,clip]{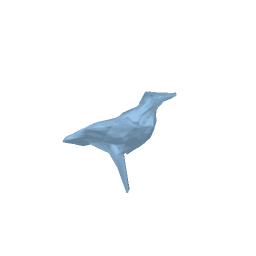} &
    \includegraphics[width=0.153\linewidth,trim=10pt 40pt 10pt 40pt,clip]{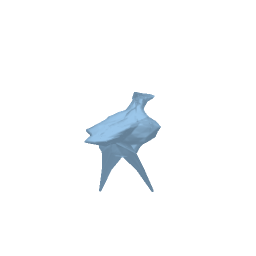} &
    \includegraphics[width=0.153\linewidth,trim=10pt 40pt 10pt 40pt,clip]{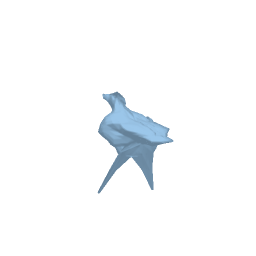} \\
    \includegraphics[width=0.153\linewidth,trim=10pt 60pt 10pt 40pt,clip]{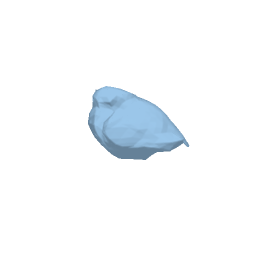} &
    \includegraphics[width=0.153\linewidth,trim=10pt 60pt 10pt 40pt,clip]{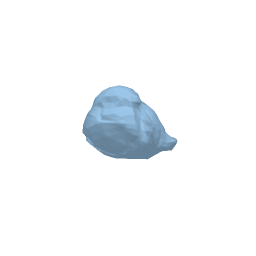} &
    \includegraphics[width=0.153\linewidth,trim=10pt 60pt 10pt 40pt,clip]{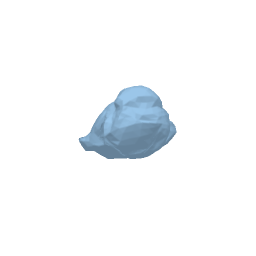} &
    \includegraphics[width=0.153\linewidth,trim=10pt 60pt 10pt 40pt,clip]{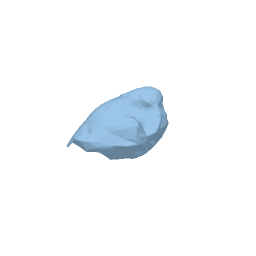} &
    \includegraphics[width=0.153\linewidth,trim=10pt 60pt 10pt 40pt,clip]{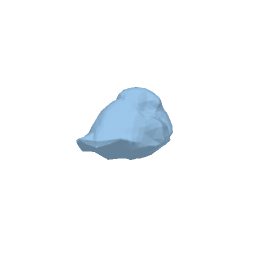} &
    \includegraphics[width=0.153\linewidth,trim=10pt 60pt 10pt 40pt,clip]{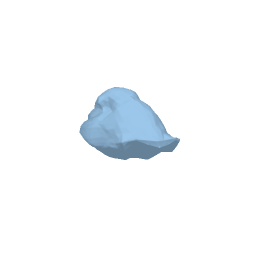} \\
    \includegraphics[width=0.153\linewidth,trim=10pt 60pt 10pt 40pt,clip]{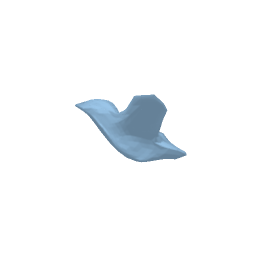} &
    \includegraphics[width=0.153\linewidth,trim=10pt 60pt 10pt 40pt,clip]{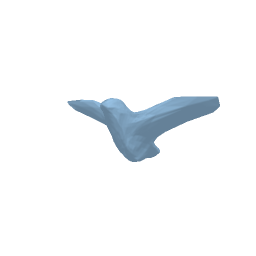} &
    \includegraphics[width=0.153\linewidth,trim=10pt 60pt 10pt 40pt,clip]{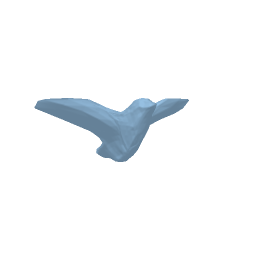} &
    \includegraphics[width=0.153\linewidth,trim=10pt 60pt 10pt 40pt,clip]{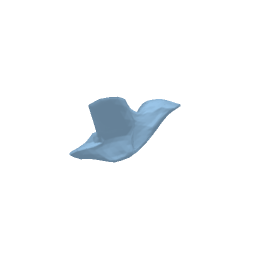} &
    \includegraphics[width=0.153\linewidth,trim=10pt 60pt 10pt 40pt,clip]{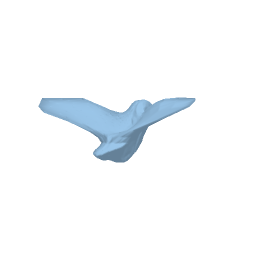} &
    \includegraphics[width=0.153\linewidth,trim=10pt 60pt 10pt 40pt,clip]{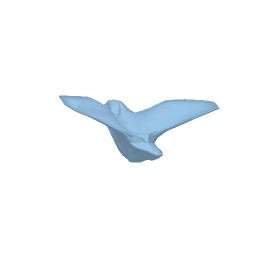} \\
    \includegraphics[width=0.153\linewidth,trim=10pt 60pt 10pt 40pt,clip]{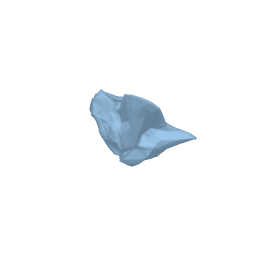} &
    \includegraphics[width=0.153\linewidth,trim=10pt 60pt 10pt 40pt,clip]{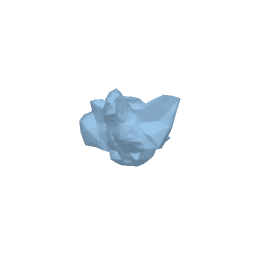} &
    \includegraphics[width=0.153\linewidth,trim=10pt 60pt 10pt 40pt,clip]{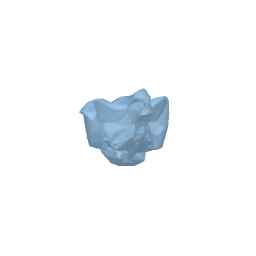} &
    \includegraphics[width=0.153\linewidth,trim=10pt 60pt 10pt 40pt,clip]{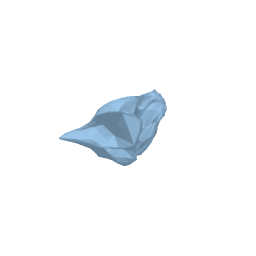} &
    \includegraphics[width=0.153\linewidth,trim=10pt 60pt 10pt 40pt,clip]{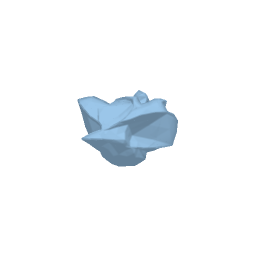} &
    \includegraphics[width=0.153\linewidth,trim=10pt 60pt 10pt 40pt,clip]{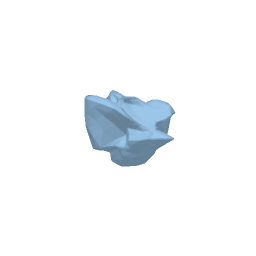} \\
    \includegraphics[width=0.153\linewidth,trim=10pt 40pt 10pt 40pt,clip]{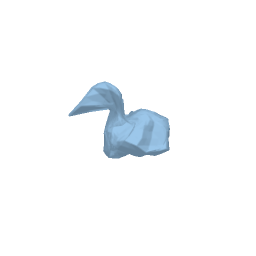} &
    \includegraphics[width=0.153\linewidth,trim=10pt 40pt 10pt 40pt,clip]{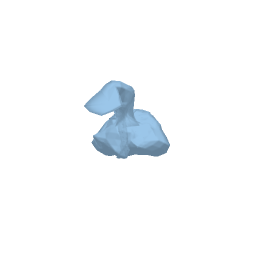} &
    \includegraphics[width=0.153\linewidth,trim=10pt 40pt 10pt 40pt,clip]{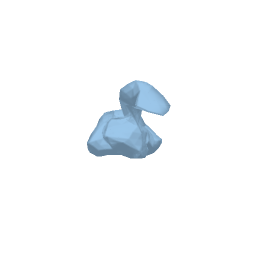} &
    \includegraphics[width=0.153\linewidth,trim=10pt 40pt 10pt 40pt,clip]{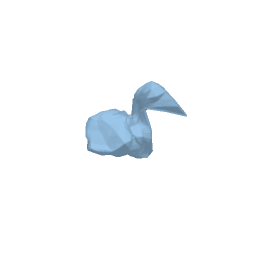} &
    \includegraphics[width=0.153\linewidth,trim=10pt 40pt 10pt 40pt,clip]{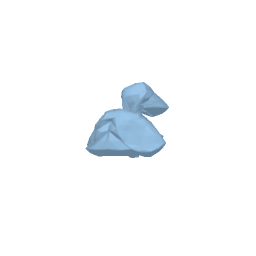} &
    \includegraphics[width=0.153\linewidth,trim=10pt 40pt 10pt 40pt,clip]{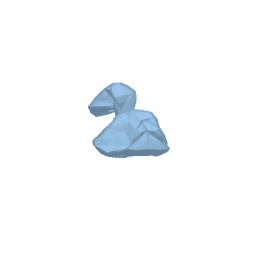}
    \end{tabular}
    \vspace{-5pt}
    \caption{Some of the meanshapes learned by our method trained on CUB~\cite{WahCUB_200_2011}, using $14$ meanshapes, without encoder pre-training on ImageNet.}
    \label{fig:meanshapes_cub_nopretrain}
    \vspace{-1em}
\end{figure}

\section{Additional ablation studies}\label{sec:ablation}
In this section, we present further experiments on the datasets Pascal3D+ and CUB.

\begin{table*}[t]
    \begin{center}
    \renewcommand{\arraystretch}{1.1}
    \resizebox{0.9\linewidth}{!}{
    \begin{tabular}{c|c|c|c|cc|ccc}
        \multirow{2}{*}{\textbf{Training classes}} & \textbf{ImageNet} & \textbf{Number of} & \multirow{2}{*}{\textbf{3D IoU} $\uparrow$} & \multicolumn{2}{c|}{\textbf{Mask IoU} $\uparrow$} & \multicolumn{3}{c}{\textbf{Texture metrics}} \\
        & \textbf{pre-train} & \textbf{meanshapes} & & \textbf{Pred cam} & \textbf{GT cam} & \textbf{SSIM} $\uparrow$ &  \textbf{L1} $\downarrow$ & \textbf{FID} $\downarrow$ \\
        \midrule
        aeroplane, car & \V & 2 & $\mathbf{0.550}$ & $\mathbf{0.639}$ & $\mathbf{0.700}$ & $\mathbf{0.732}$ & $\mathbf{0.066}$ & $\mathbf{353.61}$ \\
        aeroplane, car & & 2 & $0.541$ & $0.599$ & $0.675$ & $0.728$ & $0.069$ & $357.47$ \\
        \midrule
        bicycle, bus, car, motorbike & \V & 4 & $\mathbf{0.543}$ & $\mathbf{0.711}$ & $\mathbf{0.759}$ & $\mathbf{0.607}$ & $\mathbf{0.094}$ & $\mathbf{380.15}$ \\
        bicycle, bus, car, motorbike & & 4 & $0.534$ & $0.632$ & $0.727$ & $0.580$ & $0.111$ & $392.71$ \\
        \bottomrule
    \end{tabular}
    }
    \end{center}
    \vspace{-5pt}
    \caption{Evaluation on Pascal3D+~\cite{xiang2014beyond} using a ResNet-18 encoder with or without pre-trained weights on ImageNet~\cite{deng2009imagenet} (segmentation masks obtained with PointRend~\cite{kirillov2020pointrend}).}
    \label{tab:pascal_pretrain}
    \vspace{-0.5em}
\end{table*}

\begin{table}[t]
    \begin{center}
    \renewcommand{\arraystretch}{1.1}
    \resizebox{1\linewidth}{!}{
    \begin{tabular}{c|cc|ccc}
        \textbf{Imagenet} & \multicolumn{2}{c|}{\textbf{Mask IoU} $\uparrow$} & \multicolumn{3}{c}{\textbf{Texture metrics}} \\ 
        \textbf{pre-train} & \textbf{Pred cam} & \textbf{GT cam} & \textbf{SSIM} $\uparrow$ &  \textbf{L1} $\downarrow$ & \textbf{FID} $\downarrow$ \\
        \midrule
        \V & $\mathbf{0.642}$ & $\mathbf{0.723}$ & $\mathbf{0.715}$ & $\mathbf{0.065}$ & $\mathbf{231.95}$ \\
         & $0.563$ & $0.699$ & $0.693$ & $0.077$ & $259.36$ \\
        \bottomrule
    \end{tabular}
    }
    \end{center}
    \vspace{-5pt}
    \caption{Evaluation on CUB~\cite{WahCUB_200_2011} using a ResNet-18 encoder with or without pre-trained weights on ImagetNet~\cite{deng2009imagenet}.
    }
    \label{tab:cub_pretrain_single_col}
    \vspace{-1em}
\end{table}

\subsection{Impact of pre-training on shape selection}
\label{subsec:pre-training}
Since our model exploits a visual encoder pre-trained on ImageNet~\cite{deng2009imagenet}, we investigate the impact of using pre-trained weights or training the encoder from scratch, with a particular focus on the unsupervised shape selection module.
Indeed, we aim to verify that the proposed method is capable of learning meaningful meanshapes even without a pre-trained feature extractor.
Quantitative results and learned meanshapes are reported (i) in Table~\ref{tab:pascal_pretrain} and Figure~\ref{fig:meanshapes_4macroclasses_nopretrain} for Pascal3D+ and (ii) in Table~\ref{tab:cub_pretrain_single_col} and Figure~\ref{fig:meanshapes_cub_nopretrain} for the CUB dataset.
IoU and texture metrics show that the pre-trained version obtains better scores in every setting.
However, it is worth noting that the framework is capable of obtaining satisfactory results and learning meaningful meanshapes even without any pre-training of the encoder network, confirming the effectiveness of the proposed shape selection module.

\subsection{Impact of finer foreground masks}
In this section, we compare the scores obtained on Pascal3D+ using rough foreground masks, provided by Mask R-CNN~\cite{he2017mask}, or more precise masks, obtained with PointRend~\cite{kirillov2020pointrend}.
Results are reported in Table~\ref{tab:pascal_segmentation_masks}.
As expected, there is a clear advantage in using finer masks in the setting with $4$ automotive classes. Indeed, PointRend produces accurate masks, which present fine details and sharp edges, that are leveraged by the framework during the training process.
On the other hand, a relatively small improvement can be observed when training on just aeroplanes and cars. This may be due to a different quality of the aeroplane masks between Mask R-CNN and PointRend. 

\subsection{Meanshape learning during training}
In order to evaluate the unsupervised learning of multiple meanshapes during the training process, we report the learned shapes at different epochs in Figure~\ref{fig:meanshapes_4macroclasses_epochs} (Pascal3D+) and in Figure~\ref{fig:meanshapes_cub_epochs} (CUB).
These results show that the method distinguishes different object categories within the first few epochs and then progressively optimize each meanshape accordingly.
While the classes are clearly disentangled in just tens of epochs on Pascal3D+, the same process requires more epochs on CUB.
We believe that this difference is due to the class type: classes of different entities on Pascal3D+, different classes of the same entity ``bird'' on CUB.
Nevertheless, the method progressively learns meaningful meanshapes in both settings.

\subsection{Number of meanshapes on CUB}
The CUB dataset contains images of the same category ``bird''.
However, the dataset can be split in many sub-categories, for instance using the annotated bird type ($200$ different values) or one of the other annotated categorical attributes (\eg the ``has\_shape'' one provides $14$ different values, including \textit{duck-like}, \textit{gull-like}, \textit{hummingbird-like}, \textit{long-legged-like}).
Thus, in the paper we empirically set the number of meanshapes as the number of the ``has\_shape'' attribute values.
Here, we analyze the impact of using different numbers of meanshapes, testing the framework with $1$, $10$, $14$, and $18$ meanshapes and reporting the results in Table~\ref{tab:cub_meanshapes_single_col}.
Differently from the training on Pascal3D+, in this case there are no clear advantages, in terms of mask IoU and texture scores, in using a single or multiple meanshapes.
However, as clearly shown in the paper and in Figure~\ref{fig:meanshapes_cub_epochs}, the method can exploit the available meanshapes to learn meaningful base shapes in an unsupervised manner.
These base shapes can then be used as representative shapes for the whole dataset or as bird templates in other tasks.
In addition, we did not find an explicit pattern in using different numbers of meanshapes. This shows that the initialization of this hyper-parameter is not crucial for the learning process, in particular when the class division is not perfectly clear.

\subsection{Unsupervised shape classification}
In this section, we investigate the usage of the unsupervised shape selection module as classifier on the Pascal3D+ dataset. In particular, we evaluate whether the most weighted meanshape represents the object category.
In the $2$-class setting (aeroplane, car), the obtained classification accuracy is $98.82\%$; in the $4$-class setting (bicycle, bus, car, motorbike), the classification accuracy is $93.45\%$. In the latter case, the classes bicycle and motorbike are considered a single class, given that the method learned a single meanshape that represents both.

\begin{table}[t]
    \begin{center}
    \renewcommand{\arraystretch}{1.1}
    \resizebox{1\linewidth}{!}{
    \begin{tabular}{c|cc|ccc}
        \textbf{Number of} & \multicolumn{2}{c|}{\textbf{Mask IoU} $\uparrow$} & \multicolumn{3}{c}{\textbf{Texture metrics}} \\ 
        \textbf{meanshapes} & \textbf{Pred cam} & \textbf{GT cam} & \textbf{SSIM} $\uparrow$ &  \textbf{L1} $\downarrow$ & \textbf{FID} $\downarrow$ \\
        \midrule
        $1$ & $\mathbf{0.658}$ & $0.721$ & $0.717$ & $0.064$ & $\mathbf{227.24}$ \\
        $10$& $0.657$ & $0.721$ & $\mathbf{0.720}$ & $\mathbf{0.063}$ & $232.84$ \\
        $14$ & $0.642$ & $0.723$ & $0.715$ & $0.065$ & $231.95$ \\
        $18$ & $0.648$ & $\mathbf{0.724}$ & $0.715$ & $0.065$ & $228.24$ \\
        \bottomrule
    \end{tabular}
    }
    \end{center}
    \vspace{-5pt}
    \caption{Evaluation on CUB~\cite{WahCUB_200_2011} using different numbers of meanshapes (1, 10, 14, 18).
    }
    \label{tab:cub_meanshapes_single_col}
    \vspace{-1em}
\end{table}

\subsection{Average meanshape weights}
\begin{figure*}[t!]
    \centering
    \setlength{\tabcolsep}{1.5pt}
    \renewcommand{\arraystretch}{1}
    \vspace{16pt}
    \begin{tabular}{cccccccc}
        \includegraphics[width=0.07\linewidth,trim=10pt 60pt 10pt 40pt,clip]{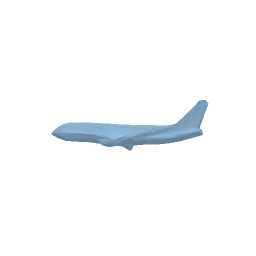} &
        \includegraphics[width=0.07\linewidth,trim=10pt 60pt 10pt 40pt,clip]{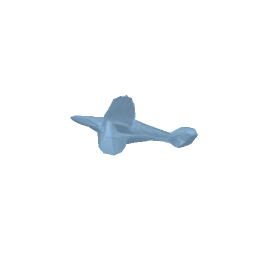} &
        \includegraphics[width=0.07\linewidth,trim=10pt 60pt 10pt 40pt,clip]{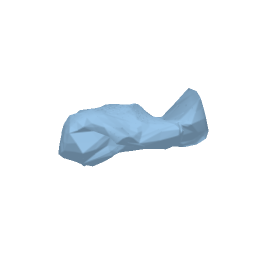} &
        \includegraphics[width=0.07\linewidth,trim=10pt 60pt 10pt 40pt,clip]{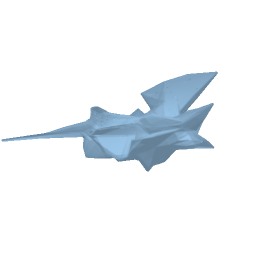} &
        \includegraphics[width=0.07\linewidth,trim=10pt 60pt 10pt 40pt,clip]{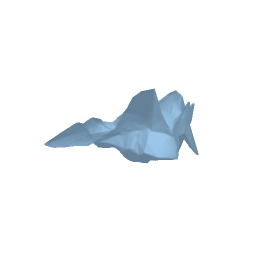} &
        \includegraphics[width=0.07\linewidth,trim=10pt 60pt 10pt 40pt,clip]{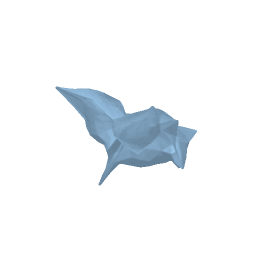} &
        \includegraphics[width=0.07\linewidth,trim=10pt 60pt 10pt 40pt,clip]{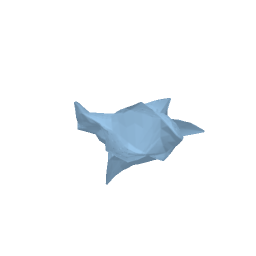} &
        \includegraphics[width=0.07\linewidth,trim=10pt 60pt 10pt 40pt,clip]{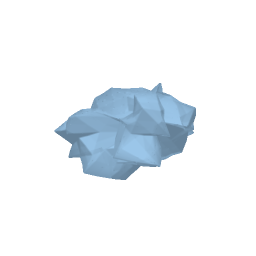} \\
        \footnotesize{$0.62$} & 
        \footnotesize{$0.15$} & 
        \footnotesize{$0.11$} & 
        \footnotesize{$0.03$} & 
        \footnotesize{$0.03$} & 
        \footnotesize{$0.03$} &
        \footnotesize{$0.02$} & 
        \footnotesize{$0.02$}
        \vspace{1em}
    \end{tabular}
    \begin{tabular}{cccccccccccccc}
        \includegraphics[width=0.055\linewidth,trim=10pt 40pt 10pt 40pt,clip]{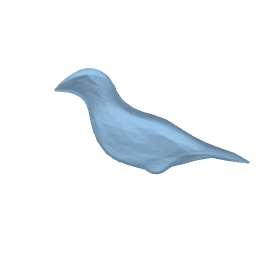} &
        \includegraphics[width=0.055\linewidth,trim=10pt 40pt 10pt 40pt,clip]{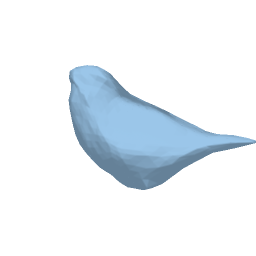} &
        \includegraphics[width=0.055\linewidth,trim=10pt 40pt 10pt 40pt,clip]{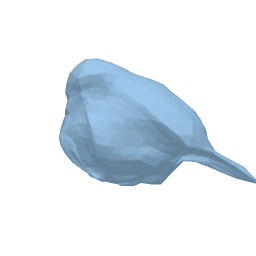} &
        \includegraphics[width=0.055\linewidth,trim=10pt 40pt 10pt 40pt,clip]{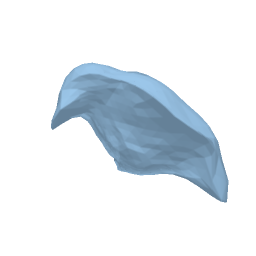} &
        \includegraphics[width=0.055\linewidth,trim=10pt 40pt 10pt 40pt,clip]{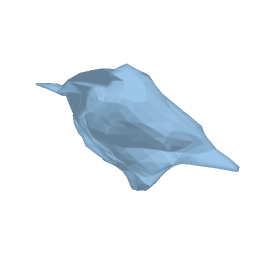} &
        \includegraphics[width=0.055\linewidth,trim=10pt 40pt 10pt 40pt,clip]{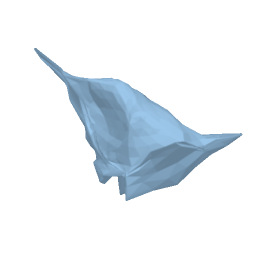} &
        \includegraphics[width=0.055\linewidth,trim=10pt 40pt 10pt 40pt,clip]{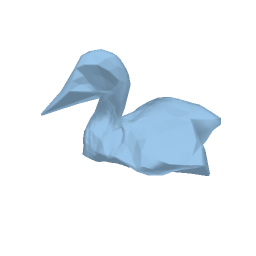} &
        \includegraphics[width=0.055\linewidth,trim=10pt 40pt 10pt 40pt,clip]{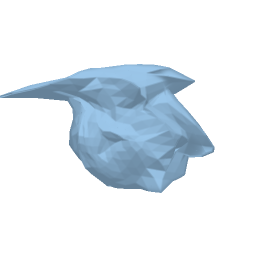} &
        \includegraphics[width=0.055\linewidth,trim=10pt 40pt 10pt 40pt,clip]{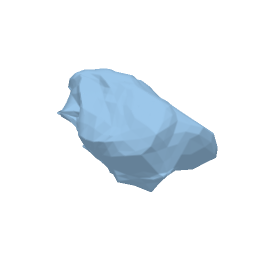} &
        \includegraphics[width=0.055\linewidth,trim=10pt 40pt 10pt 40pt,clip]{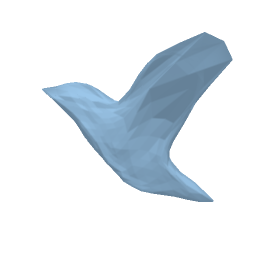} &
        \includegraphics[width=0.055\linewidth,trim=10pt 40pt 10pt 40pt,clip]{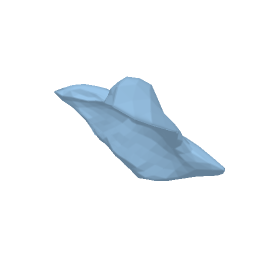} &
        \includegraphics[width=0.055\linewidth,trim=10pt 40pt 10pt 40pt,clip]{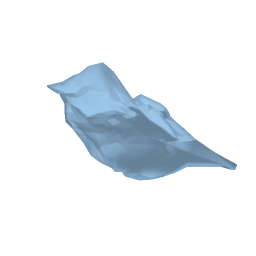} &
        \includegraphics[width=0.055\linewidth,trim=10pt 40pt 10pt 40pt,clip]{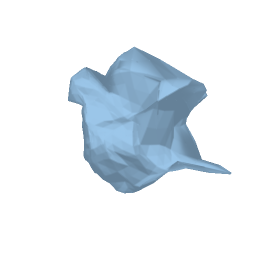} &
        \includegraphics[width=0.055\linewidth,trim=10pt 40pt 10pt 40pt,clip]{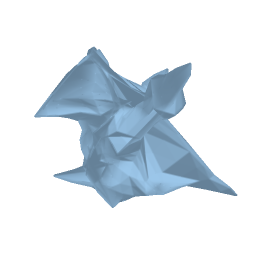} \\
\end{tabular}
    \vspace{-5pt}
    \captionof{figure}{Average meanshape weights for Pascal3D+ (aeroplane class, $8$ meanshapes) and CUB ($14$ meanshapes) ordered from the most weighted to the least one.}
    \label{fig:meanshapes}
    \vspace{-1em}
\end{figure*}

To evaluate the importance of each meanshape on the predicted shape, we compute the average meanshape weight predicted by the unsupervised shape selection module. Results are reported in Figure~\ref{fig:meanshapes} for all the meanshapes of the experiments with aeroplanes (Pascal3D+) and birds (CUB).
While we acknowledge that there are few learned meanshapes that do not correspond to a clear object category, these meanshapes have a marginal impact on the weighted meanshape. On the contrary, the most representative meanshapes have, on average, a major contribution on the weighted one.

\section{Additional qualitative results}\label{sec:qualitative}
\begin{figure*}[t]
    \centering
    \setlength{\tabcolsep}{2pt}
    \renewcommand{\arraystretch}{0.8}
    \begin{tabular}{ccccccc}
    \small{{Input}} & \small{{weighted}} & \small{{Predicted}} & \multicolumn{4}{c}{\small{{Predicted shape} ${\hat{M}}$}} \\
    \small{{image} ${I}$} & \small{{meanshape} ${M}$} & \small{{shape} ${\hat{M}}$} & \multicolumn{4}{c}{\small{{with texture} ${\hat{I}_{tex}}$}} \\ \\
    \includegraphics[width=0.09\linewidth]{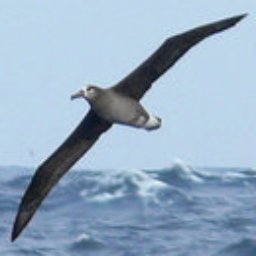} &
    \includegraphics[width=0.14\linewidth,trim=64px 0 64px 0,clip]{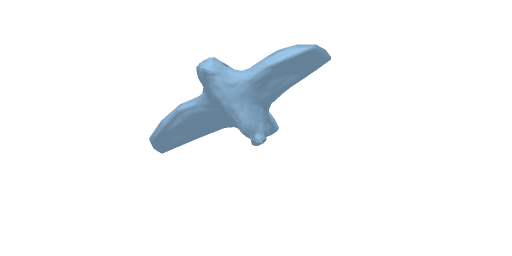} &
    \includegraphics[width=0.14\linewidth,trim=64px 0 64px 0,clip]{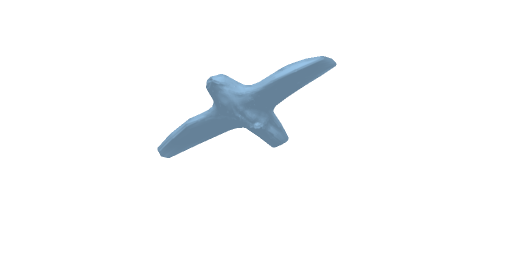} &
    \includegraphics[width=0.14\linewidth,trim=64px 0 64px 0,clip]{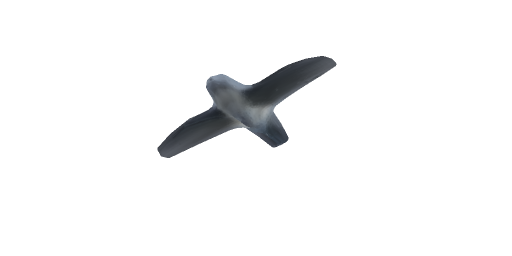} &
    \includegraphics[width=0.14\linewidth,trim=64px 0 64px 0,clip]{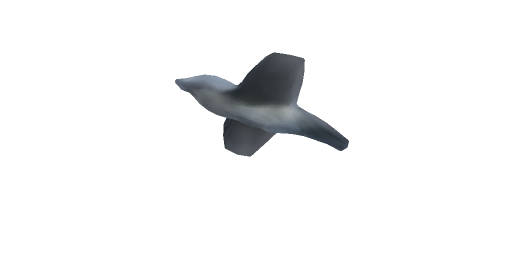} &
    \includegraphics[width=0.14\linewidth,trim=64px 0 64px 0,clip]{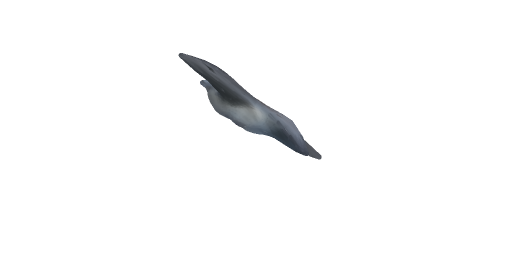} &
    \includegraphics[width=0.14\linewidth,trim=64px 0 64px 0,clip]{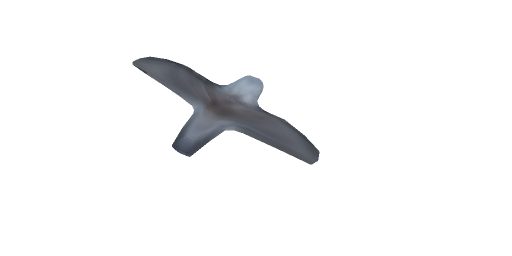} \\
    \includegraphics[width=0.09\linewidth]{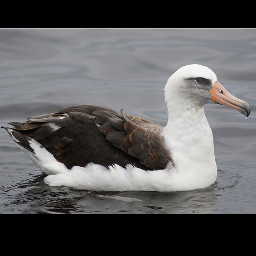} &
    \includegraphics[width=0.14\linewidth,trim=96px 0 32px 0,clip]{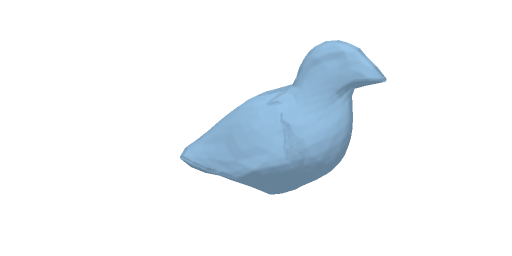} &
    \includegraphics[width=0.14\linewidth,trim=64px 0 64px 0,clip]{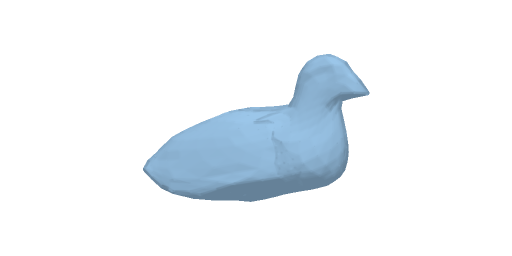} &
    \includegraphics[width=0.14\linewidth,trim=96px 0 32px 0,clip]{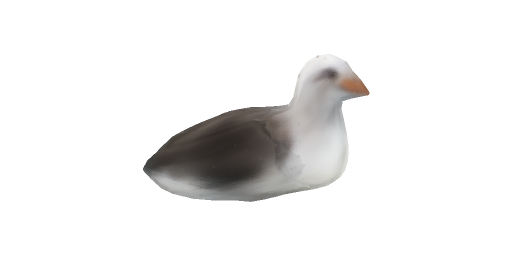} &
    \includegraphics[width=0.14\linewidth,trim=128px 0 0 0,clip]{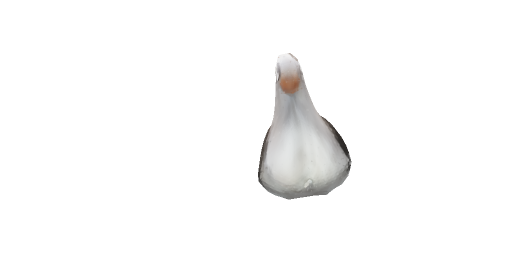} &
    \includegraphics[width=0.14\linewidth,trim=160px 0 -32px 0,clip]{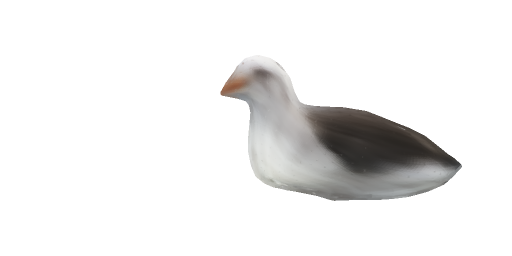} &
    \includegraphics[width=0.14\linewidth,trim=192px 0 -64px 0,clip]{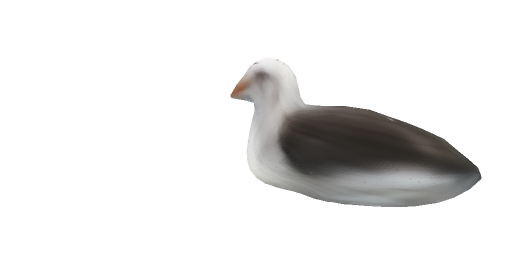} \\
\end{tabular}
    \vspace{-5pt}
    \caption{Additional qualitative results of our method trained on CUB~\cite{WahCUB_200_2011}, using $14$ meanshapes.}
    \label{fig:qualitative_results_cub}
\end{figure*}

\begin{figure*}[t]
    \centering
    \setlength{\tabcolsep}{1.6pt}
    \renewcommand{\arraystretch}{0.6}
    \begin{tabular}{cccccccc}
    & \small{{Input}} & \small{{weighted}} & \small{{Predicted}} & \multicolumn{4}{c}{\small{{Predicted shape} ${\hat{M}}$}} \\
    & \small{{image} ${I}$} & \small{{meanshape} ${M}$} & \small{{shape} ${\hat{M}}$} & \multicolumn{4}{c}{\small{{with texture} ${\hat{I}_{tex}}$}} \\ \\
    \rotatebox{90}{\parbox[t]{0.09\linewidth}{\hspace*{\fill}{\footnotesize{aeroplane}}\hspace*{\fill}}} &
    \includegraphics[width=0.09\linewidth]{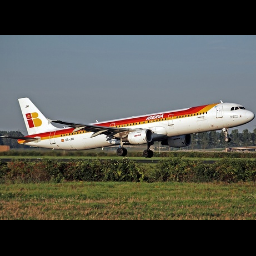} &
    \includegraphics[width=0.14\linewidth,trim=64px 0 64px 0,clip]{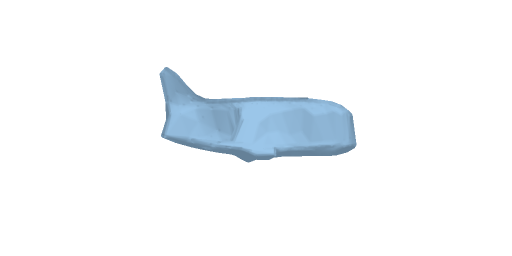} &
    \includegraphics[width=0.14\linewidth,trim=64px 0 64px 0,clip]{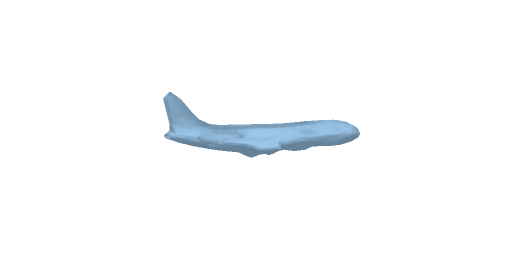} &
    \includegraphics[width=0.14\linewidth,trim=64px 0 64px 0,clip]{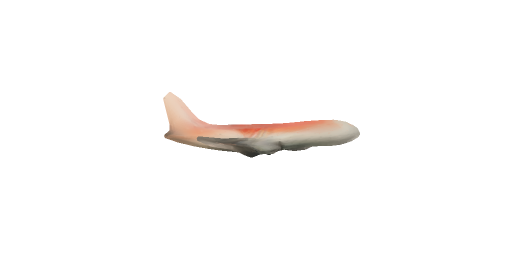} &
    \includegraphics[width=0.14\linewidth,trim=64px 0 64px 0,clip]{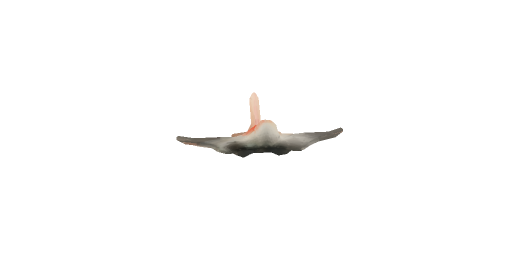} &
    \includegraphics[width=0.14\linewidth,trim=64px 0 64px 0,clip]{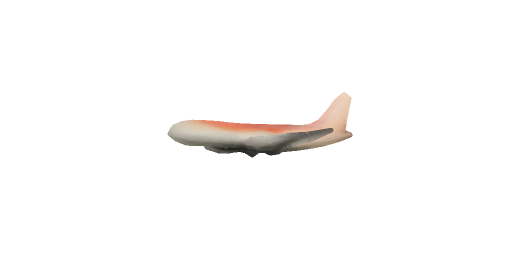} &
    \includegraphics[width=0.14\linewidth,trim=64px 0 64px 0,clip]{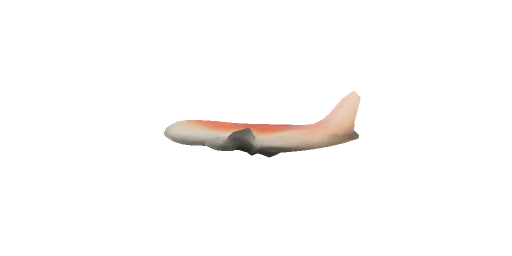} \\
\end{tabular}
    \vspace{-5pt}
    \caption{Additional qualitative results of our method trained jointly on all $12$ classes of Pascal3D+~\cite{xiang2014beyond}.}
    \label{fig:qualitative_results_pascal_all}
\end{figure*}

\begin{figure*}[t]
    \centering
    \setlength{\tabcolsep}{2pt}
    \renewcommand{\arraystretch}{0.6}
    \begin{tabular}{ccccccc}
    \small{{Input}} & \small{{weighted}} & \small{{Predicted}} & \multicolumn{4}{c}{\small{{Predicted shape} ${\hat{M}}$}} \\
    \small{{image} ${I}$} & \small{{meanshape} ${M}$} & \small{{shape} ${\hat{M}}$} & \multicolumn{4}{c}{\small{{with texture} ${\hat{I}_{tex}}$}} \\ \\
    \includegraphics[width=0.09\linewidth]{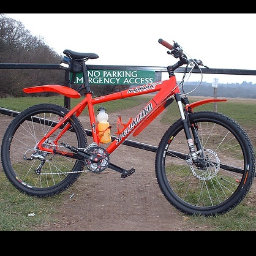} &
    \includegraphics[width=0.14\linewidth,trim=64px 0 64px 0,clip]{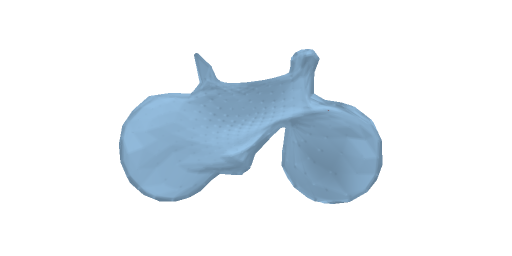} &
    \includegraphics[width=0.14\linewidth,trim=64px 0 64px 0,clip]{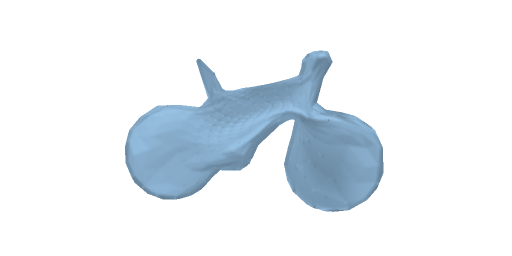} &
    \includegraphics[width=0.14\linewidth,trim=64px 0 64px 0,clip]{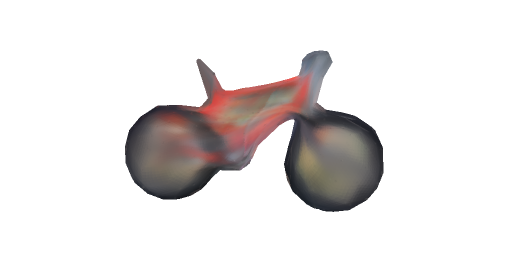} &
    \includegraphics[width=0.14\linewidth,trim=64px 0 64px 0,clip]{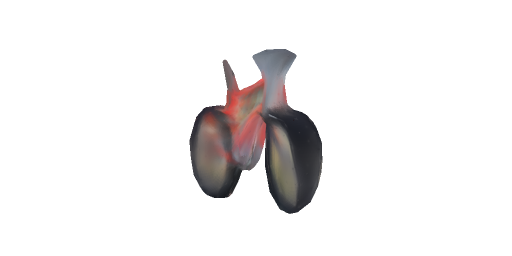} &
    \includegraphics[width=0.14\linewidth,trim=64px 0 64px 0,clip]{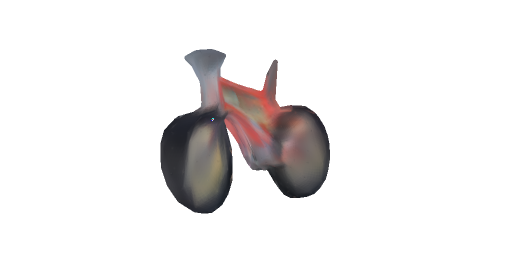} &
    \includegraphics[width=0.14\linewidth,trim=64px 0 64px 0,clip]{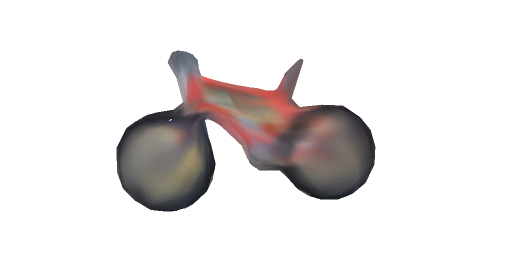} \\
    \includegraphics[width=0.09\linewidth]{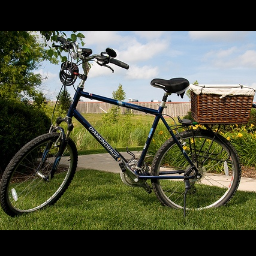} &
    \includegraphics[width=0.14\linewidth,trim=64px 0 64px 0,clip]{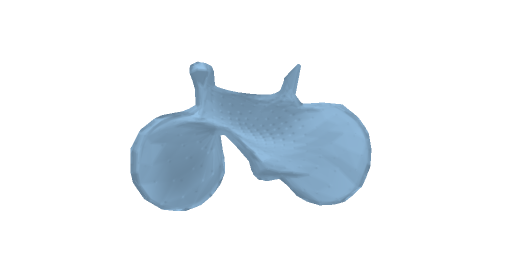} &
    \includegraphics[width=0.14\linewidth,trim=64px 0 64px 0,clip]{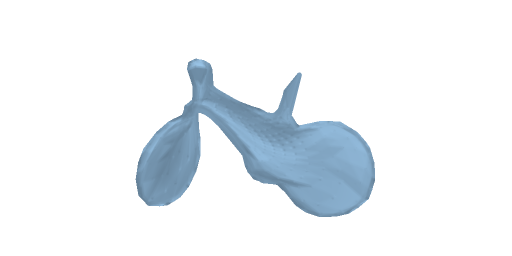} &
    \includegraphics[width=0.14\linewidth,trim=64px 0 64px 0,clip]{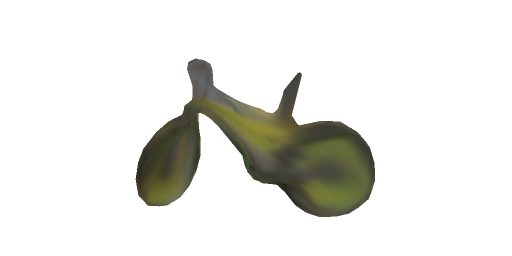} &
    \includegraphics[width=0.14\linewidth,trim=64px 0 64px 0,clip]{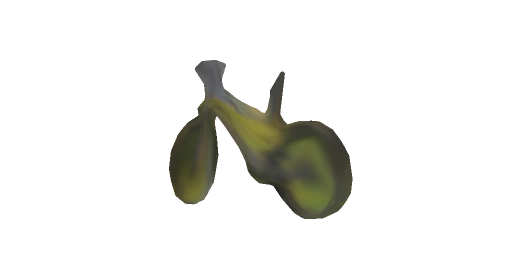} &
    \includegraphics[width=0.14\linewidth,trim=64px 0 64px 0,clip]{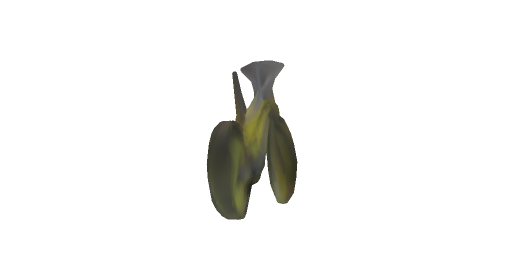} &
    \includegraphics[width=0.14\linewidth,trim=64px 0 64px 0,clip]{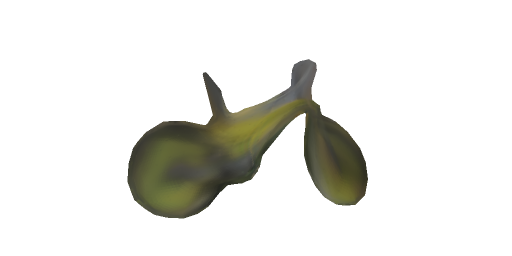} \\
\end{tabular}
    \vspace{-5pt}
    \caption{Additional qualitative results of our method trained jointly on $4$ automotive classes (bicycle, bus, car, motorbike) of Pascal3D+~\cite{xiang2014beyond}.}
    \label{fig:qualitative_results_pascal}
\end{figure*}

\begin{figure*}[t]
    \centering
    \setlength{\tabcolsep}{2pt}
    \renewcommand{\arraystretch}{0.8}
    \begin{tabular}{ccccccc}
    \small{{Input}} & \small{{weighted}} & \small{{Predicted}} & \multicolumn{4}{c}{\small{{Predicted shape} ${\hat{M}}$}} \\
    \small{{image} ${I}$} & \small{{meanshape} ${M}$} & \small{{shape} ${\hat{M}}$} & \multicolumn{4}{c}{\small{{with texture} ${\hat{I}_{tex}}$}} \\ \\
    \includegraphics[width=0.09\linewidth]{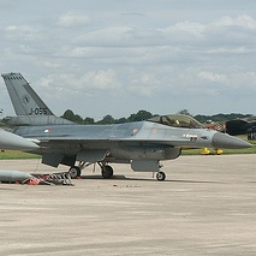} &
    \includegraphics[width=0.14\linewidth,trim=64px 0 64px 0,clip]{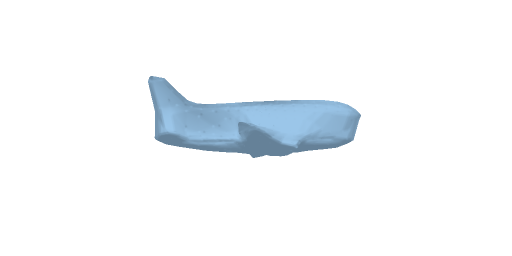} &
    \includegraphics[width=0.14\linewidth,trim=64px 0 64px 0,clip]{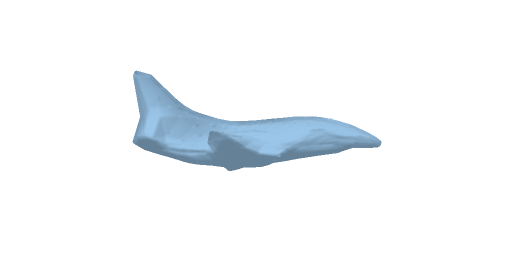} &
    \includegraphics[width=0.14\linewidth,trim=64px 0 64px 0,clip]{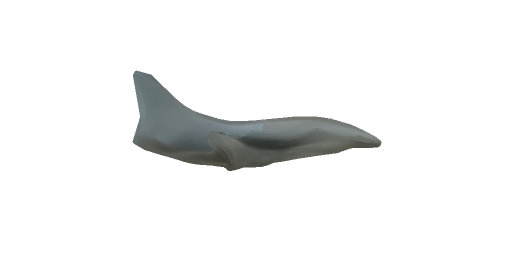} &
    \includegraphics[width=0.14\linewidth,trim=64px 0 64px 0,clip]{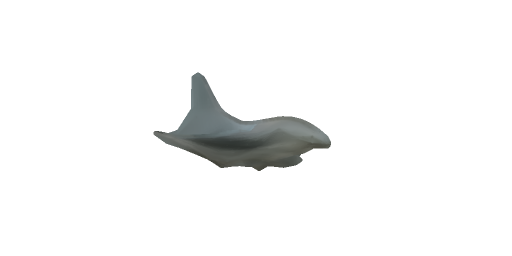} &
    \includegraphics[width=0.14\linewidth,trim=64px 0 64px 0,clip]{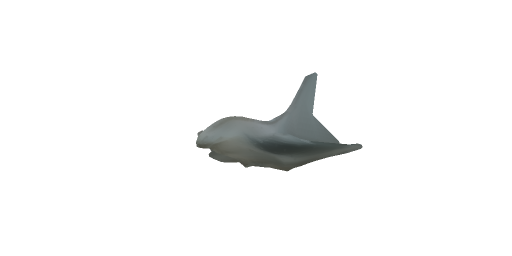} &
    \includegraphics[width=0.14\linewidth,trim=64px 0 64px 0,clip]{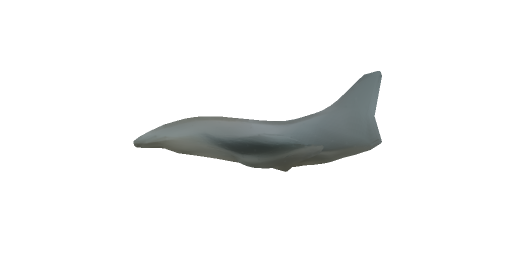} \\
    \includegraphics[width=0.09\linewidth]{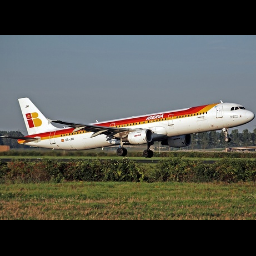} &
    \includegraphics[width=0.14\linewidth,trim=64px 0 64px 0,clip]{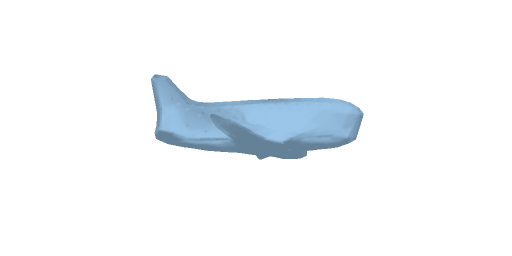} &
    \includegraphics[width=0.14\linewidth,trim=64px 0 64px 0,clip]{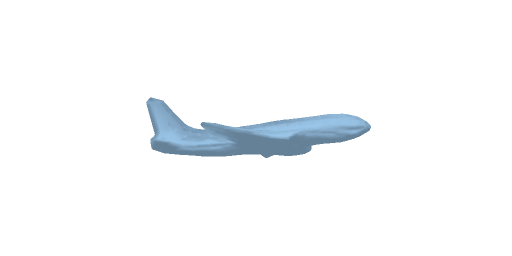} &
    \includegraphics[width=0.14\linewidth,trim=64px 0 64px 0,clip]{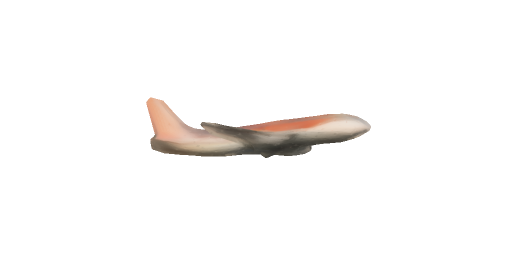} &
    \includegraphics[width=0.14\linewidth,trim=64px 0 64px 0,clip]{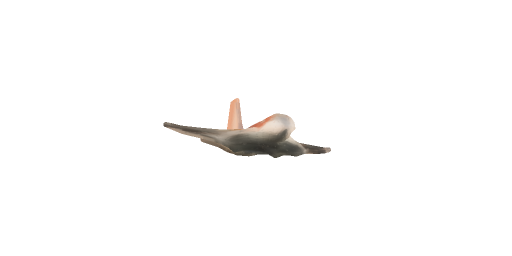} &
    \includegraphics[width=0.14\linewidth,trim=64px 0 64px 0,clip]{samples/pascal_2_suppl/texture/0148_0120.png} &
    \includegraphics[width=0.14\linewidth,trim=64px 0 64px 0,clip]{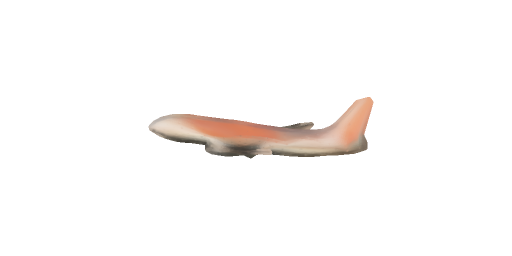} \\
\end{tabular}
    \vspace{-5pt}
    \caption{Additional qualitative results of our method trained jointly on aeroplanes and cars of Pascal3D+~\cite{xiang2014beyond}.}
    \label{fig:qualitative_results_pascal_aerocar}
\end{figure*}

We report additional qualitative results for the CUB dataset in Figure~\ref{fig:qualitative_results_cub} and for experiments on Pascal3D+ in Figure~\ref{fig:qualitative_results_pascal_all} (all $12$ classes), Figure~\ref{fig:qualitative_results_pascal} ($4$ automotive classes) and Figure~\ref{fig:qualitative_results_pascal_aerocar} (aeroplane, car).

\subsection{Failure cases}
\begin{figure*}[t]
    \centering
    \setlength{\tabcolsep}{2pt}
    \renewcommand{\arraystretch}{0.8}
    \begin{tabular}{ccccccc}
    \small{{Input}} & \small{{weighted}} & \small{{Predicted}} & \multicolumn{4}{c}{\small{{Predicted shape} ${\hat{M}}$}} \\
    \small{{image} ${I}$} & \small{{meanshape} ${M}$} & \small{{shape} ${\hat{M}}$} & \multicolumn{4}{c}{\small{{with texture} ${\hat{I}_{tex}}$}} \\ \\
    \includegraphics[width=0.09\linewidth]{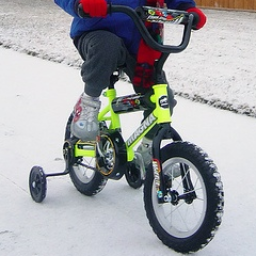} &
    \includegraphics[width=0.14\linewidth,trim=64px 0 64px 0,clip]{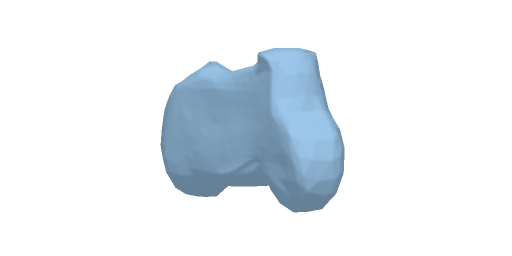} &
    \includegraphics[width=0.14\linewidth,trim=64px 0 64px 0,clip]{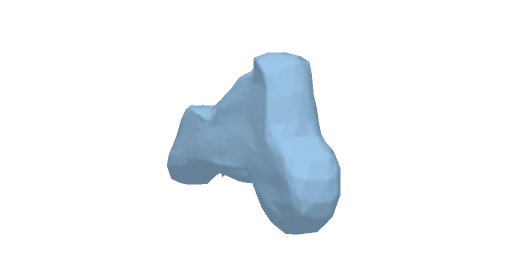} &
    \includegraphics[width=0.14\linewidth,trim=64px 0 64px 0,clip]{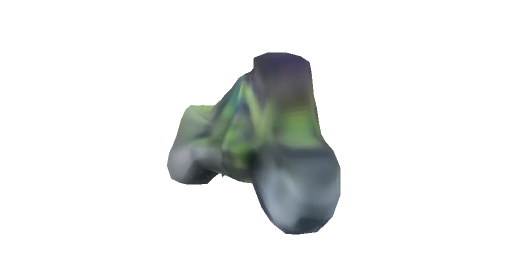} &
    \includegraphics[width=0.14\linewidth,trim=64px 0 64px 0,clip]{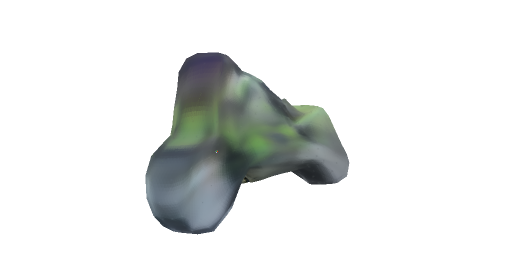} &
    \includegraphics[width=0.14\linewidth,trim=36px 0 64px 0,clip]{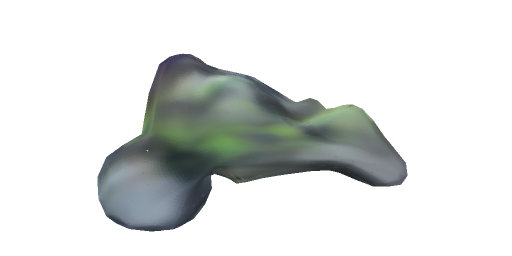} &
    \includegraphics[width=0.14\linewidth,trim=64px 0 64px 0,clip]{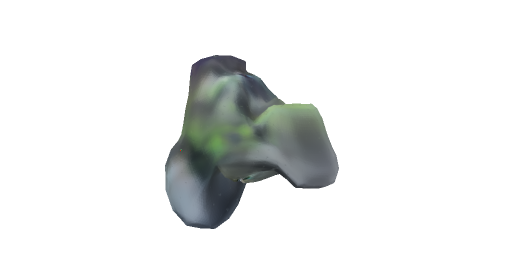} \\
    \includegraphics[width=0.09\linewidth]{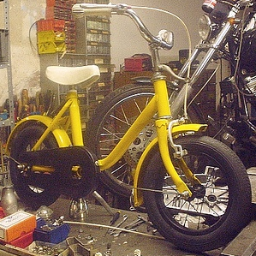} &
    \includegraphics[width=0.14\linewidth,trim=64px 0 64px 0,clip]{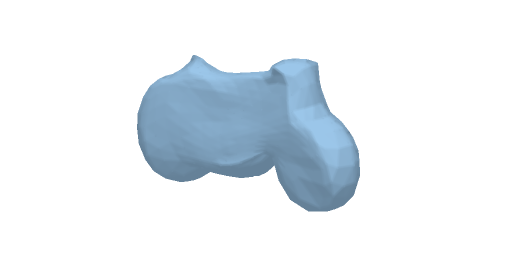} &
    \includegraphics[width=0.14\linewidth,trim=64px 0 64px 0,clip]{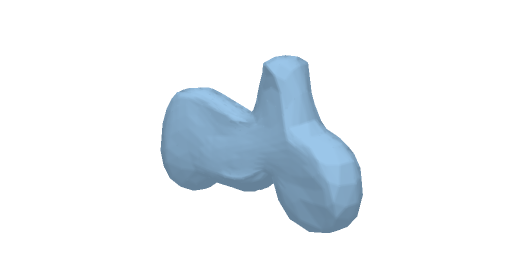} &
    \includegraphics[width=0.14\linewidth,trim=64px 0 64px 0,clip]{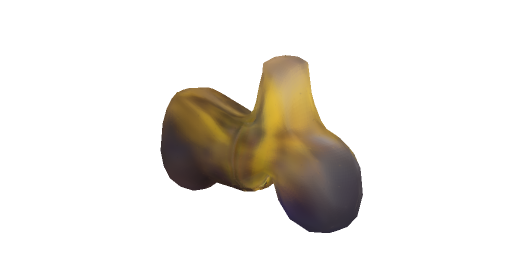} &
    \includegraphics[width=0.14\linewidth,trim=64px 0 64px 0,clip]{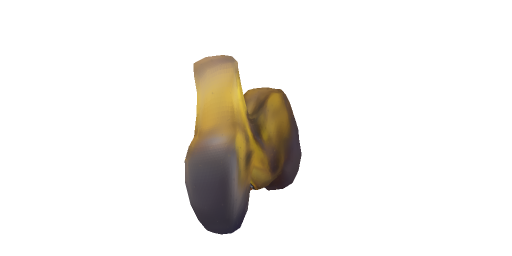} &
    \includegraphics[width=0.14\linewidth,trim=36px 0 64px 0,clip]{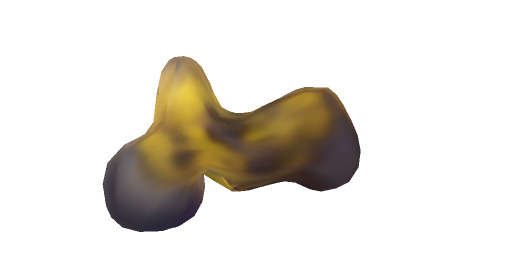} &
    \includegraphics[width=0.14\linewidth,trim=64px 0 64px 0,clip]{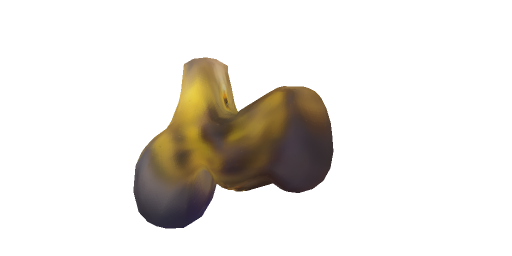} \\
\end{tabular}
    \vspace{-5pt}
    \caption{Some failure cases of our method trained jointly on $4$ automotive classes (bicycle, bus, car, motorbike) of Pascal3D+~\cite{xiang2014beyond}.}
    \label{fig:failure_cases_pascal}
\end{figure*}

In Figure~\ref{fig:failure_cases_pascal}, we report some failure cases of our method trained on $4$ automotive classes of Pascal3D+.
First of all, we identified some rare cases in which the predicted meanshape is incorrect.
For instance, bicycles with large wheels are sometimes mistaken for motorbikes while cars with roofboxes are confused with buses (Fig.~\ref{fig:failure_cases_pascal}, rows 1-3).
Moreover, we detected that the method sometimes outputs wrong deformations, causing the objects to be skewed, when the viewpoint is very close to the object (Fig.~\ref{fig:failure_cases_pascal}, rows 4-5).
Finally, in some cases the method can not predict correct deformations of articulated parts (Fig.~\ref{fig:failure_cases_pascal}, rows 6).

\begin{figure*}[t!]
    \centering
    \setlength{\tabcolsep}{1.5pt}
    \renewcommand{\arraystretch}{1}
\begin{tabular}{ccccccccccc}
    \footnotesize{$0^{th}$} & \footnotesize{$10^{th}$} & \footnotesize{$20^{th}$} & \footnotesize{$30^{th}$} & \footnotesize{$40^{th}$} & \footnotesize{$50^{th}$} & \footnotesize{$100^{th}$} & \footnotesize{$200^{th}$} & \footnotesize{$300^{th}$} & \footnotesize{$400^{th}$} & \footnotesize{$500^{th}$} \\
    \includegraphics[width=0.084\linewidth,trim=10pt 40pt 10pt 40pt,clip]{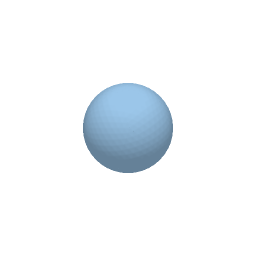} &
    \includegraphics[width=0.084\linewidth,trim=10pt 60pt 10pt 40pt,clip]{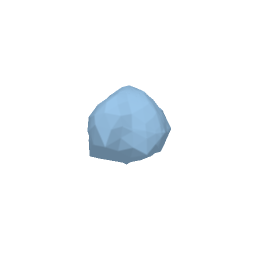} &
    \includegraphics[width=0.084\linewidth,trim=10pt 60pt 10pt 40pt,clip]{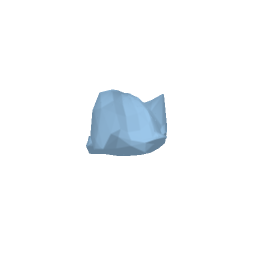} &
    \includegraphics[width=0.084\linewidth,trim=10pt 60pt 10pt 40pt,clip]{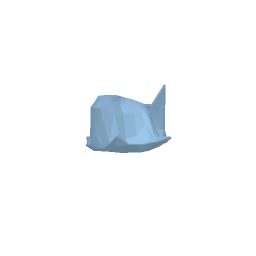} &
    \includegraphics[width=0.084\linewidth,trim=10pt 60pt 10pt 40pt,clip]{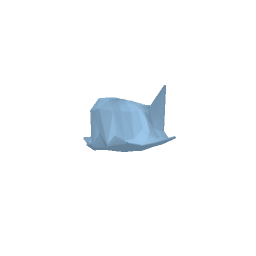} &
    \includegraphics[width=0.084\linewidth,trim=10pt 60pt 10pt 40pt,clip]{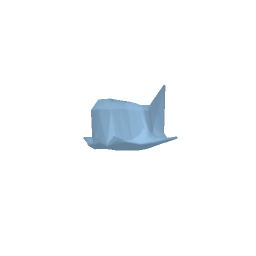} &
    \includegraphics[width=0.084\linewidth,trim=10pt 60pt 10pt 40pt,clip]{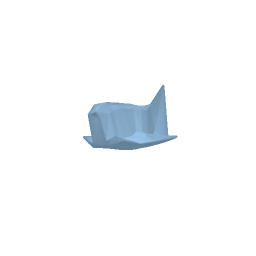} &
    \includegraphics[width=0.084\linewidth,trim=10pt 60pt 10pt 40pt,clip]{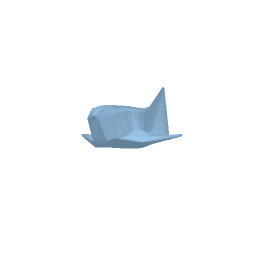} &
    \includegraphics[width=0.084\linewidth,trim=10pt 60pt 10pt 40pt,clip]{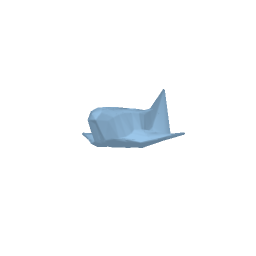} &
    \includegraphics[width=0.084\linewidth,trim=10pt 60pt 10pt 40pt,clip]{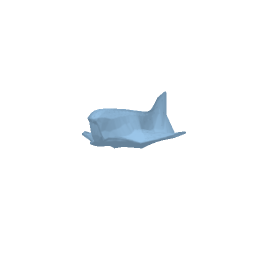} &
    \includegraphics[width=0.084\linewidth,trim=10pt 60pt 10pt 40pt,clip]{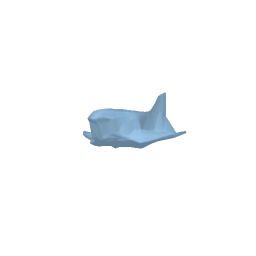} \\
\end{tabular}
    \vspace{-5pt}
    \caption{Meanshapes learned during different epochs of the training procedure on aeroplanes and cars (rows 1-2), and on 4 automotive classes (rows 3-6) of Pascal3D+~\cite{xiang2014beyond}.}
    \label{fig:meanshapes_4macroclasses_epochs}
\end{figure*}

\begin{figure*}[t!]
    \centering
    \setlength{\tabcolsep}{1.5pt}
    \renewcommand{\arraystretch}{1}
    \vspace{16pt}
    \begin{tabular}{ccccccccccc}
    \footnotesize{$0^{th}$} & \footnotesize{$10^{th}$} & \footnotesize{$20^{th}$} & \footnotesize{$30^{th}$} & \footnotesize{$40^{th}$} & \footnotesize{$50^{th}$} & \footnotesize{$100^{th}$} & \footnotesize{$200^{th}$} & \footnotesize{$300^{th}$} & \footnotesize{$400^{th}$} & \footnotesize{$500^{th}$} \\
    \includegraphics[width=0.084\linewidth,trim=10pt 60pt 10pt 40pt,clip]{samples/meanshapes_4macroclasses/sphere.png} &
    \includegraphics[width=0.084\linewidth,trim=10pt 60pt 10pt 40pt,clip]{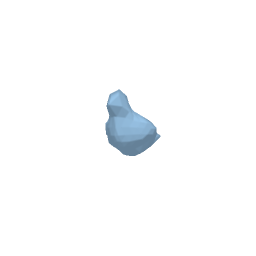} &
    \includegraphics[width=0.084\linewidth,trim=10pt 60pt 10pt 40pt,clip]{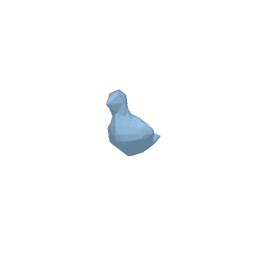} &
    \includegraphics[width=0.084\linewidth,trim=10pt 60pt 10pt 40pt,clip]{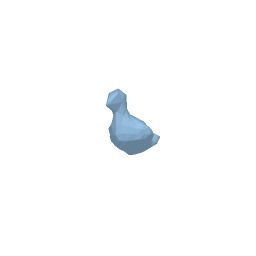} &
    \includegraphics[width=0.084\linewidth,trim=10pt 60pt 10pt 40pt,clip]{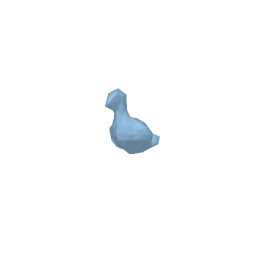} &
    \includegraphics[width=0.084\linewidth,trim=10pt 60pt 10pt 40pt,clip]{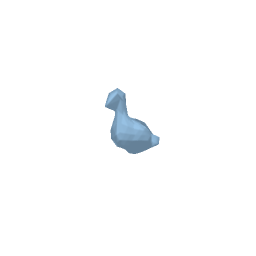} &
    \includegraphics[width=0.084\linewidth,trim=10pt 60pt 10pt 40pt,clip]{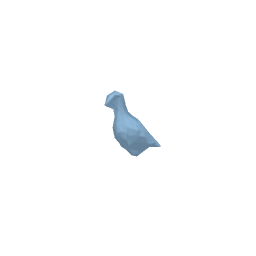} &
    \includegraphics[width=0.084\linewidth,trim=10pt 60pt 10pt 40pt,clip]{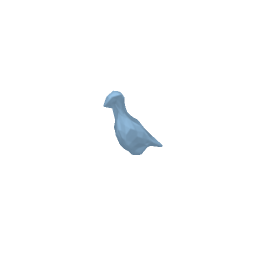} &
    \includegraphics[width=0.084\linewidth,trim=10pt 60pt 10pt 40pt,clip]{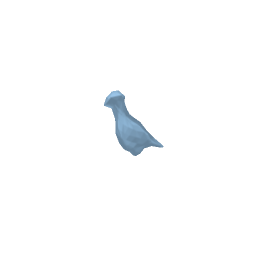} &
    \includegraphics[width=0.084\linewidth,trim=10pt 60pt 10pt 40pt,clip]{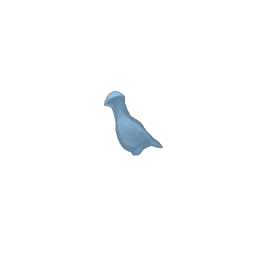} &
    \includegraphics[width=0.084\linewidth,trim=10pt 60pt 10pt 40pt,clip]{samples/meanshapes_cub/ms_000_300.png} \\
\end{tabular}
    \vspace{-5pt}
    \caption{Some of the meanshapes learned during different epochs of the training procedure on CUB~\cite{WahCUB_200_2011}, using $14$ meanshapes.}
    \label{fig:meanshapes_cub_epochs}
    \vspace{-1em}
\end{figure*}

\end{document}